\documentclass[10pt,twocolumn,letterpaper]{article}

\usepackage[pagenumbers]{cvpr} % To force page numbers, e.g. for an arXiv version

\usepackage[dvipsnames]{xcolor}

\definecolor{cvprblue}{rgb}{0.21,0.49,0.74}
\usepackage[pagebackref,breaklinks,colorlinks,citecolor=cvprblue]{hyperref}
\usepackage[accsupp]{axessibility}
\usepackage{wrapfig}
\usepackage{multirow}
\usepackage{graphicx} 
\usepackage{cancel}
\usepackage{float}
\usepackage{subcaption}
\usepackage{caption}
\usepackage{microtype}
\usepackage{booktabs}
\usepackage{array}
\usepackage{soul}

\usepackage{color, colortbl}
\def\paperID{10727} % *** Enter the Paper ID here
\def\confName{CVPR}
\def\confYear{2024}

\newcommand{\suppmat}{\emph{Supp.\,Mat.\,}}

\newcommand\blfootnote[1]{%
  \begingroup
  \renewcommand\thefootnote{}\footnote{#1}%
  \addtocounter{footnote}{-1}%
  \endgroup
}

\newcommand{\highlight}[2]{\colorbox{#1!17}{$\displaystyle #2$}}

\title{PAIR Diffusion: A Comprehensive Multimodal Object-Level Image Editor}

\author{Vidit Goel\textsuperscript{\textcolor{cvprblue}{1,2},*} \quad Elia Peruzzo\textsuperscript{\textcolor{cvprblue}{3},*,\textdagger} \quad Yifan Jiang \textsuperscript{\textcolor{cvprblue}{4}} \quad Dejia Xu\textsuperscript{\textcolor{cvprblue}{4}} \quad Xingqian Xu\textsuperscript{\textcolor{cvprblue}{2,1}} \quad  \\ Nicu Sebe\textsuperscript{\textcolor{cvprblue}{3}} \quad Trevor Darrell\textsuperscript{\textcolor{cvprblue}{5}} \quad Zhangyang Wang\textsuperscript{\textcolor{cvprblue}{1,4}} \quad Humphrey Shi\textsuperscript{\textcolor{cvprblue}{1,2}} \\
\normalsize{\textsuperscript{\textcolor{cvprblue}{1}}Picsart AI Research (PAIR)  \quad \textsuperscript{\textcolor{cvprblue}{2}}SHI Labs @ Georgia Tech \& UIUC\quad \textsuperscript{\textcolor{cvprblue}{3}}University of Trento \quad \textsuperscript{\textcolor{cvprblue}{4}}UT Austin \quad \textsuperscript{\textcolor{cvprblue}{5}}UC Berkeley}
\vspace{2mm}\\\href{https://github.com/Picsart-AI-Research/PAIR-Diffusion}{\texttt{https://github.com/Picsart-AI-Research/PAIR-Diffusion}}\vspace{-5mm}%}
}

\begin{document}
\twocolumn[{
\vspace{-10mm}
\maketitle
\begin{center}
        \includegraphics[width=\textwidth]{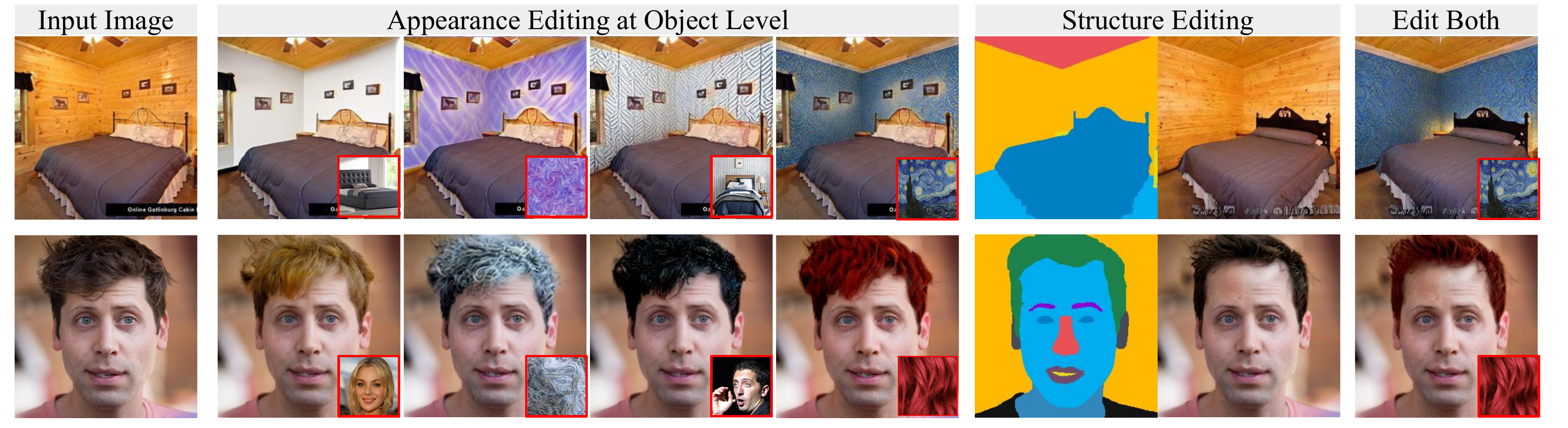}
        (a)
        \label{fig:teaser1}
        \includegraphics[width=\textwidth]{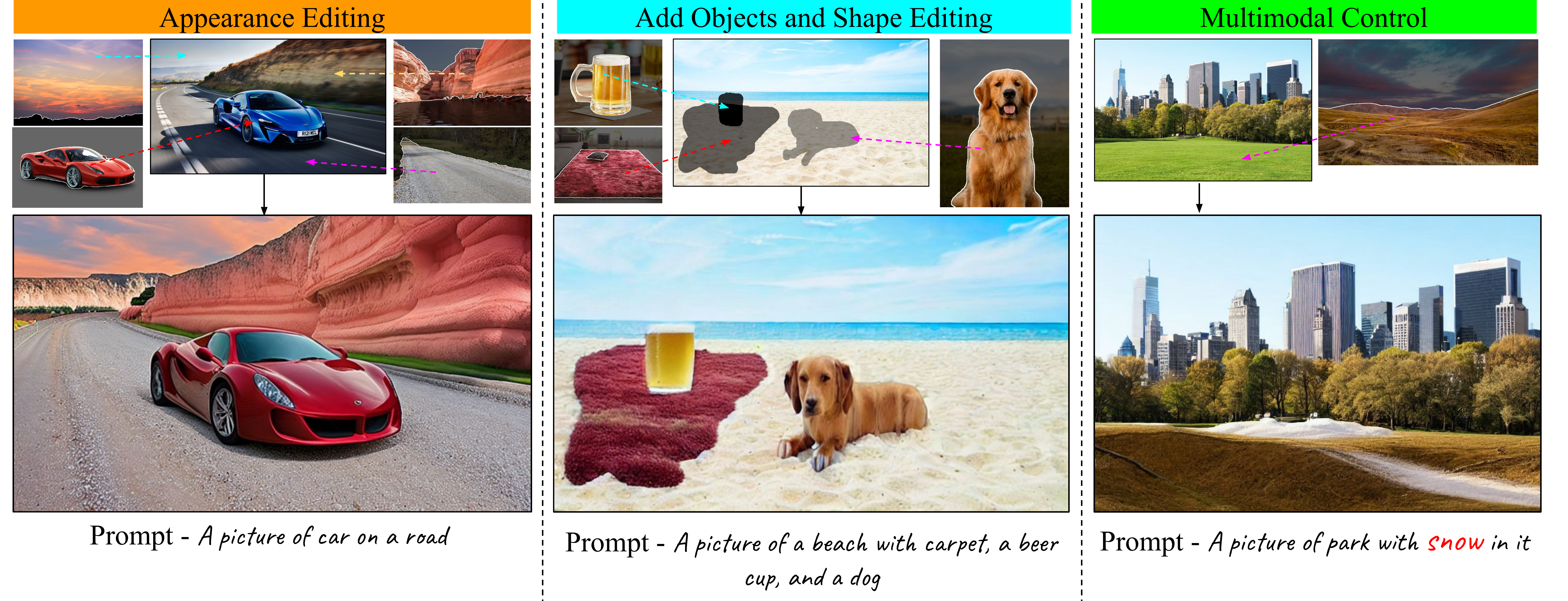}
        (b)
        \label{fig:teaser2}
        % \end{subfigure}   
    \vspace{-2mm}
    \captionof{figure}{
     PAIR Diffusion framework allows the appearance and structure editing of an image at the object level. Our framework is general and can enable object-level editing capabilities in both (a) unconditional diffusion models and (b) foundational diffusion models. Using our framework with a foundational diffusion model allows for comprehensive in-the-wild object-level editing capabilities.
    }
    %\vspace{-2mm}
    \label{fig:front}
\end{center}
}]
\blfootnote{\hspace{-2em}\textsuperscript{*}Denotes equal contribution.\\\textdagger Work performed while interning at Picsart AI Research.%\vspace{2mm} Code and models available at the \href{https://vidit98.github.io/publication/conference-paper/pair_diff.html}{Project Page}
}

\begin{abstract}
Generative image editing has recently witnessed extremely fast-paced growth.
Some works use high-level conditioning such as text, while others use low-level conditioning. Nevertheless, most of them lack fine-grained control over the properties of the different objects present in the image, \ie object-level image editing. In this work, we tackle the task by perceiving the images as an amalgamation of various objects and aim to control the properties of each object in a fine-grained manner. Out of these properties, we identify structure and appearance as the most intuitive to understand and useful for editing purposes.  
 We propose \textbf{PAIR Diffusion}, a generic framework that enables a diffusion model to control the structure and appearance properties of each object in the image.
We show that having control over the properties of each object in an image leads to comprehensive editing capabilities.
Our framework allows for various object-level editing operations on real images such as reference image-based appearance editing, free-form shape editing, adding objects, and variations. 
Thanks to our design, we do not require any inversion step.
Additionally, we propose multimodal classifier-free guidance which enables editing images using both reference images and text when using our approach with foundational diffusion models.  We validate the above claims by extensively evaluating our framework on both unconditional and foundational diffusion models.
\end{abstract}    
\section{Introduction}\label{sec:intro}

Diffusion-based generative models have shown promising results in synthesizing and manipulating images with great fidelity, among which text-to-image models and their follow-up works have great influence in both academia and industry.
When editing a real image a user generally desires to have intuitive and precise control over different elements (\ie the objects) composing the image, and to manipulate them independently.
We can categorize existing image editing methods based on the level of control they have over individual objects in an image. One line of work involves the use of text prompts to manipulate images~\cite{hertz2022prompt, liu_compositional_2022, brooks2022instructpix2pix, liew_magicmix_2022}. These methods have limited capability for fine-grained control at the object level, owing to the difficulty of describing the shape and appearance of multiple objects simultaneously with text. In the meantime, prompt engineering makes the manipulation task tedious and time-consuming. Another line of work uses low-level conditioning signals 
such as masks~\citet{hu_unified_2022, zeng2022scenecomposer, patashnik2023localizing}, sketches~\cite{voynov_sketch-guided_2022}, images~\cite{song_objectstitch_2022, cao2023masactrl, yang2023paint} to edit the images. However, most of these works either fall into the prompt engineering pitfall or fail to independently manipulate multiple objects.
Different from previous works, we aim to independently control the properties of multiple objects composing an image \emph{i.e.}\,object-level editing. 
We show that we can formulate various image editing tasks under the object-level editing framework leading to comprehensive editing capabilities.

To tackle the aforementioned task, we propose a novel framework, dubbed Structure-and-Appearance \textbf{Pair}ed \textbf{Diffusion} Models (\textbf{PAIR Diffusion}).
Specifically, we perceive an image as an amalgamation of diverse objects, each described by various factors such as shape, category, texture, illumination, and depth. Then we further identified two crucial macro properties of an object: \ul{structure} and \ul{appearance}. Structure oversees an object's shape and category,
while appearance contains details like texture, color, and illumination.
% \new{while appearance captures visual aspects of an object without strict requirement to capture identity}
To accomplish this goal, PAIR Diffusion adopts an off-the-shelf network to estimate panoptic segmentation maps as the structure, and then extract appearance representation using pre-trained image encoders. We use the extracted per-object appearance and structure information to condition a diffusion model and train it to generate images.
In contrast to previous text-guided image editing works~\cite{avrahami_spatext_2022, brooks2022instructpix2pix, couairon_diffedit_2022, ruiz_dreambooth_2022}, we consider an additional reference image to control the appearance. Compared to text prompts that, although conveniently, can only vaguely describe the appearance, images can precisely define the expected texture and make fine-grained image editing easier. 
Having the ability to control the structure and appearance of an image at an object level gives us comprehensive editing capabilities. Using our framework we can achieve, 
localized free-form shape editing, appearance editing, editing shape and appearance simultaneously, adding objects in a controlled manner, and object-level image variation (Fig.~\ref{fig:front}). 
Moreover, thanks to our design we do not require any inversion step for editing real images. 

The novelty of our work lies in the way we formulate the image editing tasks that lead to a general approach to enable comprehensive editing capabilities in various models. We show the efficacy of our framework on unconditional diffusion models and foundational text-to-image diffusion models. Lastly, we propose multimodal classifier-free guidance to reap the full benefits of the text-to-image diffusion models. It enables PAIR Diffusion to control the final output using both reference images and text in a controlled manner hence getting the best of both worlds.
Thanks to our easy-to-extract representations we do not require specialized datasets for training and we show results on LSUN and Celeb-HQ datasets for unconditional models, and use the COCO dataset for foundational diffusion models. To summarize our contributions are as follows:

\begin{itemize}
    \item We propose PAIR Diffusion, a general framework to enable object-level editing in diffusion models. It allows editing the structure and appearance of each object in the image independently.
    \item The proposed design inherently supports various editing tasks using a single model:  localized free-form shape editing, appearance editing, editing shape and appearance simultaneously, adding objects in a controlled manner, and object-level image variation.
    \item Additionally, we propose a multimodal classifier-free guidance, enabling PAIR Diffusion to edit images using both reference images and text in a controlled manner when using the approach with foundational diffusion models.
\end{itemize}
\section{Related Works}\label{sec:related}
\textbf{Diffusion Models.}\,Diffusion probabilistic models~\cite{sohl2015deep} are a class of deep generative models that synthesize data through an iterative denoising process. Diffusion models utilize a forward process that applies
noise into data distribution and then reverses the forward process to reconstruct the data itself. Recently, they have gained popularity for the task of image generation~\cite{ho2020denoising, song2019generative}. Dhariwal \emph{et\,al.\,} ~\cite{dhariwal2021diffusion} introduced various techniques such as architectural improvements and classifier guidance, that helped diffusion models beat GANs in image generation tasks for the first time. Followed by this, many works started working on scaling the models ~\cite{nichol2021glide, ramesh2022hierarchical, rombach2022high, saharia2022photorealistic} to billions of parameters, improving the inference speed~\cite{salimans2022progressive} and memory cost~\cite{rombach2022high, vahdat2021score}. LDM~\cite{rombach2022high} is one the most popular models which reduced the compute cost by applying the diffusion process to the low-resolution latent space and scaled their model successfully for text-to-image generation trained on webscale data.
Other than image generation, they have been applied to various fields such as multi-modal generation~\cite{xu22versatile}, text-to-3D~\cite{poole2022dreamfusion,singer2023text}, language generation~\cite{li2022diffusion}, 3D reconstruction~\cite{gu2023nerfdiff}, novel-view synthesis~\cite{xu2022neurallift}, music generation~\cite{mittal2021symbolic}, object detection~\cite{chen2022diffusiondet}, etc.

\noindent \textbf{Generative Image Editing.} Image generation models have been widely used in image editing tasks since the inception of GANs~\cite{karras2019style, jiang2021transgan, gong2019autogan, epstein2022blobgan, ling2021editgan}, however, they were limited to edit a restricted set of images. Recent developments in the diffusion model has enabled image editing in the wild. Earlier works~\cite{rombach2022high, nichol2021glide, ramesh2022hierarchical} started using text prompts to control the generated image. This led to various text-based image editing works such as~\cite{feng2022training, mokady2023null, liu_compositional_2022}. To make localized edits works such as~\cite{hertz2022prompt, parmar2023zero, tumanyan_plug-and-play_2022} use cross-attention feature maps between text and image. InstructPix2Pix~\cite{brooks2022instructpix2pix} further enabled instruction-based image editing. However, using only text can only provide coarse edits. Works such as~\cite{avrahami_spatext_2022, zeng2022scenecomposer} explored explicit spatial conditioning to control the structure of generated images and used text to define the appearance of local regions. Works such as~\cite{couairon_diffedit_2022, liew_magicmix_2022} rely on input images and text descriptions to get the region of interest for editing. 
% Most of the mentioned works rely on text for describing the appearance of the image and some either lack localized control or do not have comprehensive editing capabilities.
% ~\cite{mou2023t2i, huang2023composer} add various conditioning signals to control images.
However, most of the mentioned works lack object-level editing capabilities and some still rely only on text for describing the appearance.
 % however their design lacks control over the individual objects.
Recent works such as ~\cite{mou2023dragondiffusion, epstein2023diffusion} have object-level editing capabilities, however, they are based on the classifier guidance technique at inference time which leads to limited precision. Further, they show results only on stable diffusion and require inversion to edit real images. Our framework is general and can be applied to any diffusion model. We also enable multimodal control of the appearances of objects in the image when using our framework with stable diffusion.
\section{PAIR Diffusion}\label{sec:method}
\begin{figure*}
  \centering
  \includegraphics[width=1\linewidth]{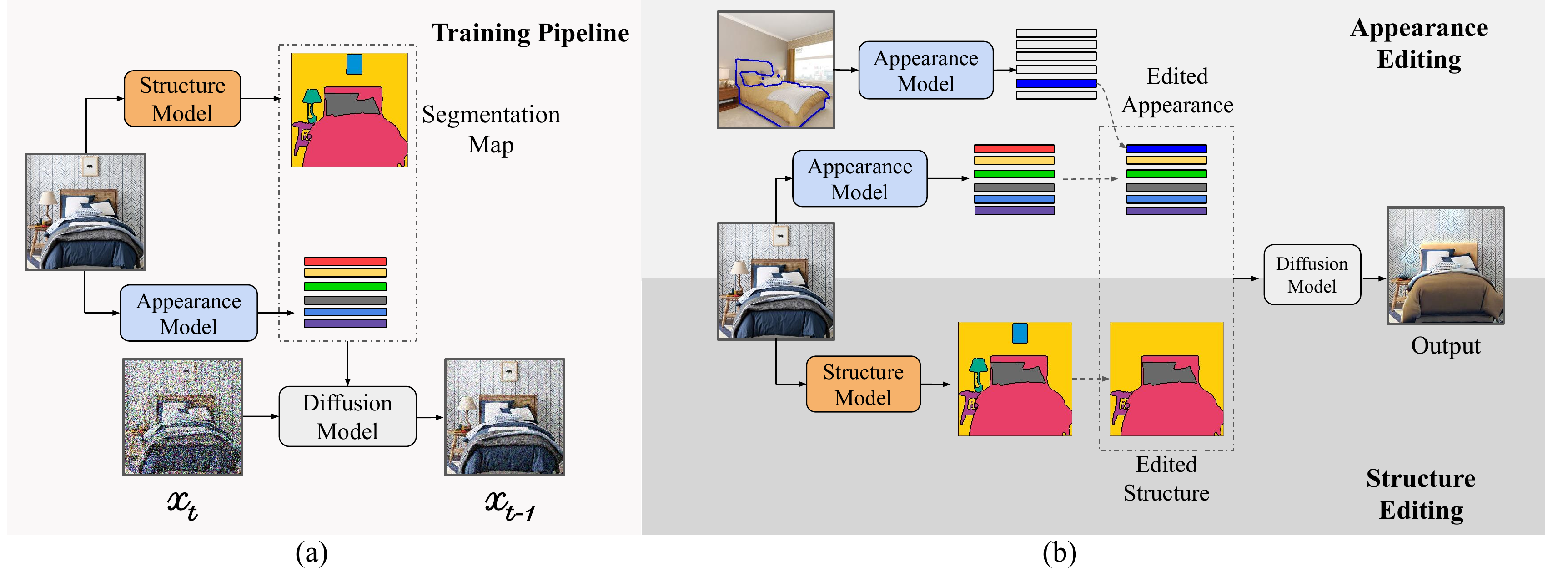}
  \caption{Overview of PAIR Diffusion. An image is seen as a composition of objects each defined by different properties like structure (shape and category), appearance, depth, etc. We focus on controlling structure and appearance. 
   (a) During training, we extract structure and appearance information and train a diffusion model in a conditional manner. (b) At inference, the framework supports multiple editing operations by independently controlling the structure and appearance of any real image at the object level.}
\label{fig:main} 
\end{figure*}

In this work, we aim to develop an image-editing framework that allows the editing of the properties of individual objects in the image. We perceive an image $x \in R^{3 \times H \times W}$ as composition of objects $O = \{o_1, o_2, \ldots, o_n\}$ where $o_i$ represents the properties of $i^{\text{th}}$ object in the image.
 As discussed in \cref{sec:intro}, we focus on enabling control over the structure and the appearance of each object. Thus, let %using reference images
 $o_i = (s_i, f_i)$ where $s_i$ represents the structure, $f_i$ represents the appearance.  The distribution that we aim to model can be written as:

\begin{equation}
\label{eq:basic}
p(x |  O, y) =  p(x | \{(s_1\,f_1),\dots,  (s_n,\,f_n)\}, y)
\end{equation}

We use $y$ to represent any form of conditioning signal already present in the generative model, e.g. text, and develop our framework to enable new object-level editing capabilities while preserving the original conditioning.
%\change{Here $y$ represents other conditions which were already present in the generative model such as text.}
The rest of the method section is organized as follows.
In \cref{sec:sar}, we describe the method to obtain $s_i$ and $f_i$ for every object in a given image. Next, in \cref{sec:formulate}, we show that various image editing tasks can be defined in the scope of the proposed object-level formulation of images.
Finally, in \cref{sec:pairdiff}, we describe the usage of the representations to augment the generative models and inference techniques to achieve object-level editing in practice. 
 
\subsection{Structure and Appearance Representation}\label{sec:sar}
Given an image $x \in R^{3 \times H \times W}$ we want to extract the structure and appearance of each object present in the image. 

\noindent\textbf{Structure.}~The structure oversees the object’s shape and category and is represented as  $s_i = (c_i, m_i)$ where $c_i$ represents the category and $m_i \in \{0,1\}^{H \times W}$ represents the shape. We extract the structure information using a panoptic segmentation map, as it readily provides each object's category and shape information and is easy to compute. Given an off-the-shelf segmentation network $E_S(\cdot)$, we obtain 
$S = E_S(x)$, with $S \in \mathbb{N}^{H \times W}$ which gives direct access to $c_i, m_i$. 

\noindent\textbf{Appearance.}~The appearance representation is designed to capture the visual aspects of the object. To represent the object faithfully, it needs to capture both the low-level features like color, texture, etc., as well as the high-level features in the case of complex objects.
% like houses or cars. 
To capture such a wide range of information, we choose a combination of convolution and transformer-based image encoders~\cite{raghu2021vision}, namely VGG~\cite{simonyan2014very} and DINOv2~\cite{oquab2023dinov2}.% to get appearance information. 
 We use initial layers of VGG to capture low-level characteristics such as color, texture etc.~\cite{yosinski2015understanding, zeiler2014visualizing}.
Conversely, DINOv2 has well-learned representations and has shown promising results for various downstream computer vision tasks. 
Hence, we use the middle layers of DINOv2 to capture the high-level characteristics of the object.

To compute per-object appearance representations, we first extract the feature maps from $l^{\text{th}}$ block of an encoder $E_G(\cdot)$, \ie $\tilde{G} = E^l_G(x)$, $\tilde{G} \in \mathbb{R}^{C \times h \times w}$,  with $ h \times w$ the spatial size and $C$ the number of channels. Then, we parse object-level features, relying on $m_i$  to pool over the spatial dimension and obtain the appearance vector $g^l_i\in  \mathbb{R}^{C}$:

\begin{equation}
\label{eq:style}
    g^l_i = \dfrac{\sum_{j, k} E^l_G(x) \odot m_i}{\sum_{j, k} {m_i}}
\end{equation}

In our framework, $E_G(\cdot)$ could be either DINOv2 or VGG. We use $g^{Vl}_i$ and $g^{Dl}_i$ to respectively denote the appearance vectors obtained using the features of VGG and DINOv2 extracted at the $l^{\text{th}}$-block. The appearance information of $i^{\text{th}}$ object is then given by a tuple $f_i = (g^{Vl_1}_i, g^{Dl_2}_i, g^{Dl_3}_i)$ where $l_2 < l_3$. As a convention, we arrange the features in $f_i$ in ascending order of abstraction, from low-level to high-level representations.
%The abstraction level of features in $f_i$ increases from $g^{Vl_1}_i$ to $g^{Dl_3}_i$.

\subsection{Image Editing Formulation}
\label{sec:formulate}
%We can define various image editing tasks using the proposed object-level design.
The proposed object-level design allows the definition of various image editing tasks within a single framework.
Consider an image $x$ with $n$ objects $O = \{o_1, o_2, \ldots, o_n\}$, with each object $o_i$ described by the structure $s_i$ and the appearance $f_i$ (see Sec.~\ref{sec:sar}).
Below we present the fundamental image editing operations that can be obtained with our framework. Importantly, they can be composed and applied to multiple objects, enabling \emph{comprehensive} editing capabilities.
%We present fundamental image editing operations below. The editing operations can be mixed with each other enabling a wide range of editing capabilities.

\noindent\textbf{Appearance Editing $(s_i, f_i) \rightarrow (s_i, f'_i)$}. It is achieved by swapping appearance vector $f_i$ with an edited appearance vector $f'_i$. Formally,  $f'_i = a_0f_i + a_1f^{R}_j$ with $f^{R}_j$ the appearance vector of the $j^{\text{th}}$ object in the reference image. 

%We use the method described in  Sec.~\ref{sec:sar} to extract the appearance vectors from reference images and compute a convex combination of them to get $f'_i$. Formally, $f'_i = a_0f_i + a_1f^{R}_i$ where $f^{R}_i$ represents the appearance vectors of $i^{th}$ object in the reference image.

\noindent\textbf{Shape Editing $(s_i, f_i) \rightarrow (s'_i, f_i)$}. It is obtained by modifying the structure $(c_i, m_i)$ to $(c_i, m'_i)$ i.e. the shape
can be explicitly changed by the user while maintaining the appearance.

\noindent\textbf{Object Addition $O \rightarrow O \cup \{o_{n+1}\}$}. We can incorporate an object into an image by specifying both its structure and appearance. These attributes can be derived either entirely from a reference image or the user can provide a sketch of the structure alone, with the appearance being inferred from a reference image.
%These properties can be obtained entirely from a reference image, or the user may sketch the structure only and the appearance is computed from a reference image.

\noindent\textbf{Object Appearance Variation}.  We can also get object-level appearance variations due to information loss in the pooling operation to calculate appearance vectors and the stochastic nature of the diffusion process. %generative models.

Once we get object with edited properties $O'$ and conditioning $y$ we can sample a new image from the learned distribution $p(x |  O', y)$. Our object-level design can easily incorporate various editing abilities and help us achieve a comprehensive image editor.
In the next section, we will describe a way to implement  $p(x | O, y)$ in practice, and present inference methods to sample and control the edited image.

\subsection{Architecture Design and Inference}\label{sec:pairdiff}
In practice, \cref{eq:basic} represents a conditional generative model; building upon the recent success of diffusion models, we leverage them to implement it. Next, we describe a method to use the object-level representations outlined in \cref{sec:sar} both in unconditional diffusion models and foundational text-to-image (T2I) diffusion models. In this way, we can transform any diffusion model into an object-level editor. 
%Our extracted representations in Sec.~\ref{sec:sar}  can be used to enable object-level editing in any diffusion model. Here we briefly describe a method to use our represents on the unconditional diffusion models and foundational text-to-image (T2I) diffusion model.

We start by representing structure and appearance in a spatial format to conveniently use them for conditioning. We represent the structure conditioning as $S \in \mathbb{N}^{2 \times H \times W}$ where the first channel contains the category, while the second channel contains the shape information of each object. For appearance conditioning, we first $L2$-normalize each vector along channel dimension, splat them spatially using $m_i$, and combine them in a single tensor represented as 
 $G \in \mathbb{R}^{C \times H \times W}$. The process is repeated for the features extracted through different encoders and at different layers, leading to the tuple $F = (G^{Vl_1},
 G^{Dl_2}, G^{Dl_3})$. Lastly, we channel-wise concatenate $S$ to every element of $F$ which results in our final conditioning signals $F_s = (G^{Vl_1}_s, G^{Dl_2}_s, G^{Dl_3}_s)$. 

In the case of the foundational T2I diffusion model, we choose Stable Diffusion (SD)~\cite{rombach2022high} as our base model. To condition it, we adopt ControlNet \cite{zhang2023adding} because of its training and data efficiency in conditioning SD model. The control module consists of encoder blocks and middle blocks that are replicated from SD UNet architecture.  
Various works show the tendency of the SD inner layers to focus more on high-level features, whereas the outer layers to focus more on low-level features~\cite{cao2023masactrl, tumanyan_plug-and-play_2022, liew_magicmix_2022}. Exploiting this finding, we use $G^{Vl_1}_s$ as input to the control module and add $G^{Dl_2}_s$, $G^{Dl_3}_s$ to the features after cross-attention in the first and second encoder blocks of the control module respectively. For the unconditional diffusion model, we use the unconditional latent diffusion model (LDM)~\cite{rombach2022high} as our base model. Pertaining to the simplicity of the architecture and training of these models we simply concatenate the features in $F_s$ to the input of LDM. The architecture is accordingly modified to incorporate the increased number of input channels.  For further details please refer to \suppmat\!.

For training both the models we follow standard practice~\cite{rombach2022high} and use the simplified training objective $\mathcal{L} =  ||\epsilon -\epsilon_\theta(z_t, S, F, y, t)||_2^2$, where $z_t$ represents the noisy version of $x$ in latent space at timestep $t$, $\epsilon$ is the noise used to get $z_t$, and $\epsilon_\theta$ is the trainable model. In the case of Stable Diffusion, $y$ represents the text prompts, while it is not present in the case of the unconditional diffusion model.%whereas $y$ can be ignored in the case of the unconditional diffusion model.

\noindent\textbf{Multimodal Inference.}~Once we have a trained model, we require an inference method that allows us to adjust the strengths of different conditioning signals and control the edited image accordingly. We consider the scenario where $y$ represents text, with the unconditional diffusion models being a special case where $y$ is null. Specifically, the structure $S$ and appearance $F$ come from a reference image and the information in $y$ could be disjoint from $F$, we need a way to capture both in the final image.
A well-trained diffusion model estimates the score function of the underlying data distribution~\cite{song2020score}, \ie $ \nabla_{z_t} p(z_t |  O, y) = \nabla_{z_t} p(z_t |  S, F, y)$, which in our case can be expanded as
\begin{equation}
\begin{aligned}
\label{eq:log_prop}
     \nabla_{z_t} \log p(z_t |  S, F, y) = & \nabla_{z_t} \log p(z_t |  S, F) \\ 
     + & \nabla_{z_t} \log p(z_t | y)  \\
     - & \nabla_{z_t} \log p(z_t) 
\end{aligned}
\end{equation}
 We use the concept of classifier-free guidance (CFG)~\cite{ho_classifier-free_2022} to represent all score functions in the above equation using a single model by dropping the conditioning with some probability during training. Using the CFG formulation we get the following update rule expanding \cref{eq:log_prop}:
\begin{equation}
\begin{split}
\label{eq:cfg}
    \tilde{\epsilon_\theta}(z_t, S, F, y) & = \epsilon_\theta(z_t, \phi, \phi, \phi) \\
    & + s_{S} \cdot(\epsilon_\theta\left(z_t, S, \phi, \phi\right)) -  \epsilon_\theta\left(z_t, \phi, \phi, \phi\right))  \\
    & + s_{F} \cdot\left(\epsilon_\theta\left(z_t, S, F, \phi\right)-\epsilon_\theta\left(z_t, S, \phi, \phi\right)\right) \\
    & + s_{y} \cdot\left(\epsilon_\theta\left(z_t, \phi, \phi, y\right) - \epsilon_\theta(z_t, \phi, \phi, \phi)\right)
\end{split}
\end{equation}

For brevity, we did not include $t$ in the equation above. A formal proof of the above equations is provided in  \suppmat\!.
Intuitively, $F$ is more information-rich compared to $y$. For this reason, during training the network learns to give negligible importance to $y$ in the presence of $F$, and we need to use $y$ independently of $F$ during inference to see its effect on the final image.
In \cref{eq:cfg} $s_S, s_F, s_y$ are guidance strengths for each conditioning signal.
It provides PAIR Diffusion with an intuitive way to control and edit images using various conditions.
For example, if a user wants to give more importance to a text prompt compared to the appearance from the reference image, it can set $s_y > s_F$ and vice-versa. For the unconditional diffusion models, we simply ignore the term corresponding to $s_y$ in \cref{eq:cfg}.

\section{Experiments}\label{sec:experiments}

In this section, we present qualitative and quantitative analysis that show the advantages of the PAIR diffusion framework introduced in \cref{sec:method}. 
We refer to UC-PAIR Diffusion to denote our framework applied to unconditional diffusion models and reserve the name PAIR Diffusion when applying the framework to Stable Diffusion.
Evaluating image editing models is hard, moreover, few works have comprehensive editing capabilities at the object level making a fair comparison even more challenging. For these reasons, we perform two main sets of experiments. Firstly, we train UC-PAIR Diffusion on widely used image-generation datasets such as the bedroom and church partitions of the LSUN Dataset \citep{yu2015lsun}, and the CelebA-HQ Dataset \citep{karras2017progressive}. We conduct quantitative experiments on these datasets as they represent a well-study benchmark, with a clear distinction between training and testing sets, making it easier and fairer to perform evaluations. 
Secondly, we fine-tune PAIR Diffusion on the COCO~\citep{lin2014microsoft} dataset. We use this model to perform in-the-wild editing and provide examples for the use cases described in \cref{sec:formulate}, showing the comprehensive editing capabilities of our method. 
We refer the reader to the \suppmat for the details regarding model training and implementations, along with additional results.

\subsection{Editing Applications}\label{sec:edit}

\begin{figure*}
  \centering
  \includegraphics[width=1\linewidth]{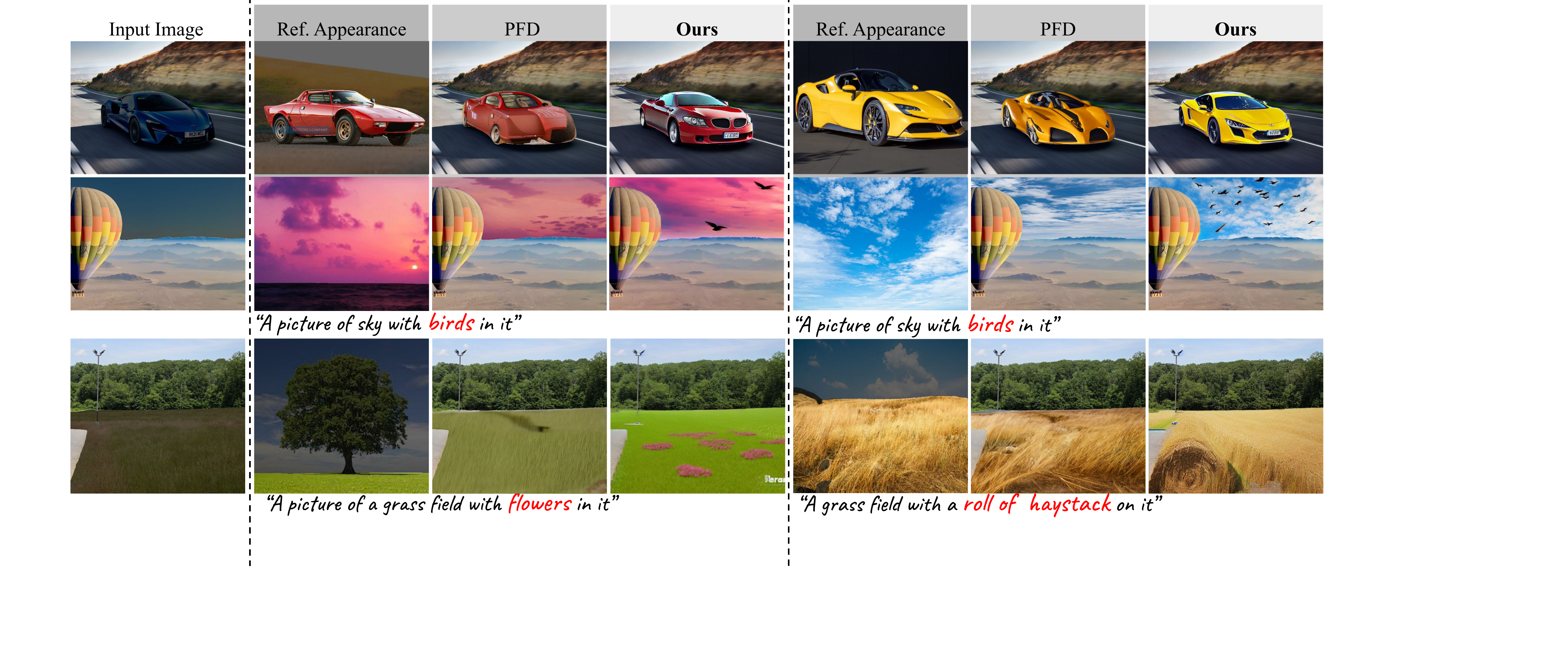}
  \vspace{-0.7cm}
  \caption{Qualitative results for appearance editing. We can drive the edit with reference images as well as with text prompts.}
  \vspace{-0.2cm}
\label{fig:app_edit} 
\end{figure*}

In this section, we qualitatively validate that our model can achieve comprehensive object-level editing capabilities in practice.
We primarily show results using PAIR Diffusion and refer to the \suppmat for results on smaller datasets. 
We use different baselines according to the editing task. We adapt Prompt-Free-Diffusion (PFD) ~\citet{xu2023prompt} as a baseline for localized appearance editing, %adapting it for localized appearance editing 
by introducing masking and using the cropped reference image as input. Moreover, we adopt Paint-By-Example (PBE)~\citet{yang2023paint} as a baseline for adding objects and shape editing.
For further details regarding implementation please refer to \suppmat\!. 
When we want the final output to be influenced by the text prompt as well we set $s_y > s_F$ else we set $s_y < s_F$.
For the figures where there is no prompt provided below the image assume that prompt was auto-generated using the template: \emph{`A picture of \{category\}'}, with the category inferred form the edited object. When editing a local region we used a masked sampling technique to only affect the selected region~\citep{rombach2022high}. 

\begin{figure*}
  \centering
  \includegraphics[width=0.95\linewidth]{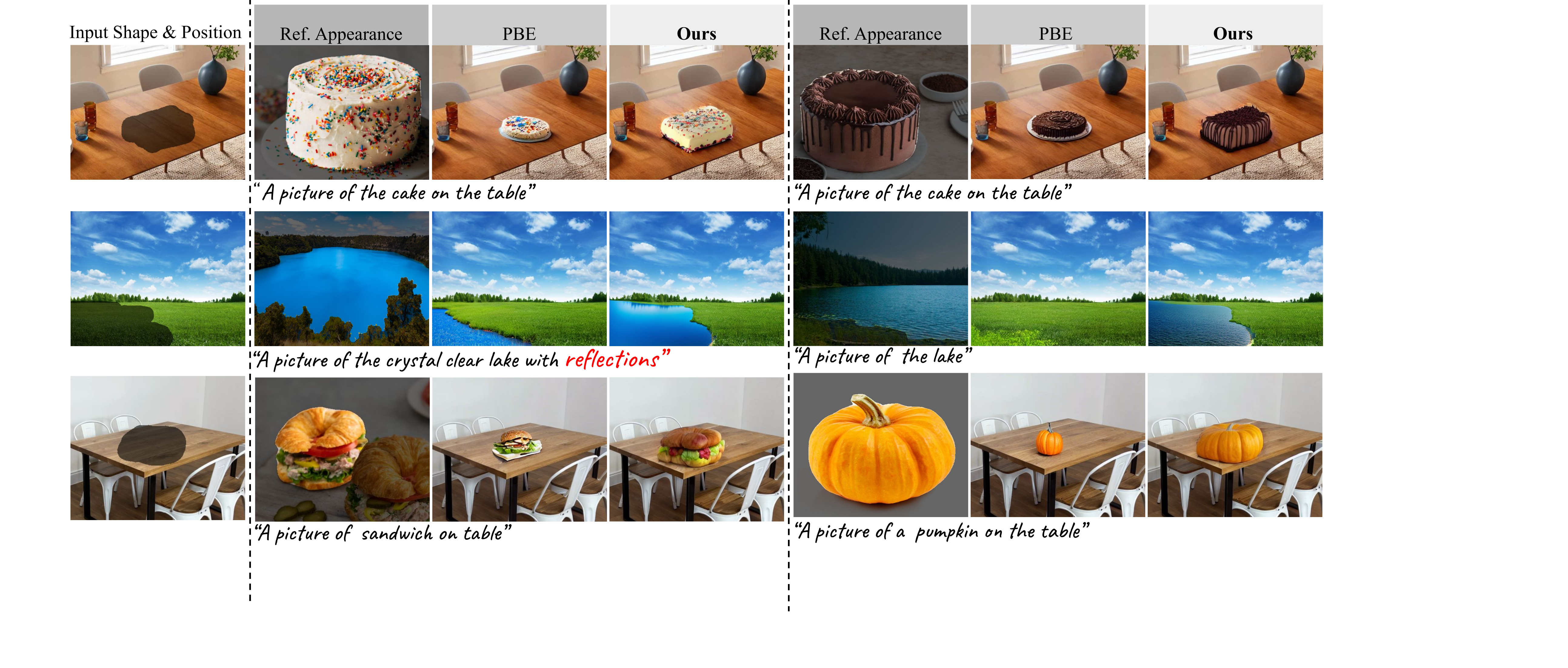}
  \vspace{-0.2cm}
  \caption{Qualitative results for adding objects and shape editing.}
  \vspace{-0.2cm}
\label{fig:struct_edit} 
\end{figure*}

\noindent \textbf{Appearance Editing.} In \cref{fig:app_edit}, we report qualitative results for appearance editing driven by reference images and text. 
We observe that our multilevel appearance representation and object-level design help us edit the appearance of both simple objects such as the sky as well as complex objects like cars.
On the other hand, PFD~\citep{xu2023prompt}  gives poor results when editing the appearance of complex objects due to the missing object-level design. 
Furthermore, using our multimodal classifier free guidance, our model can seamlessly blend the information from the text and the reference images to get the final edited output whereas PFD~\citep{xu2023prompt}  lacks this ability.

\noindent \textbf{Add objects and Shape editing.}~We show the object addition and shape editing operations result together in \cref{fig:struct_edit}. 
With PAIR Diffusion we can add complex objects with many details like a cake, as well as simpler objects like a lake. 
When changing the structure of the cake from a circle to a square, the model captures the sprinkles and dripping chocolate on the cake while rendering it in the new shape. 
In all the examples, we can see that the edges of the newly added object blend smoothly with the underlying image. On the other hand, PBE~\citep{yang2023paint} completely fails to follow the desired shape and faces issues with large objects like lakes.

\noindent \textbf{Object Variations.}~We can also achieve image variations at an object level as shown in \cref{fig:sd_variations} in \suppmat\!. We note that our model can capture various details of the original object and still produce variations. 

\begin{table}[t]
\centering
\def\arraystretch{0.0}
\footnotesize
\resizebox{\columnwidth}{!}{
\setlength\tabcolsep{0.5pt}
\begin{tabular}{ccccc}
        %\rotatebox[origin=c]{45}{Input Image} & \rotatebox[origin=c]{45}{Reference Image}&  \rotatebox[origin=c]{45}{CP+DDIM+LDM}& E2EVE & SAP \\
        Input & Reference & CP+Den. & E2EVE %\cite{Brown2022E2EVE}
        & \textbf{PAIR Diff.} \\
        \includegraphics[trim=0 0 0 0,clip,width=0.2\columnwidth]{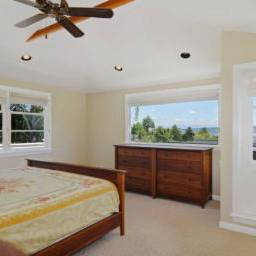} &
        \includegraphics[trim=0 0 0 0,clip,width=0.2\columnwidth]{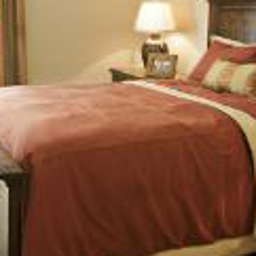} &
        \includegraphics[trim=0 0 0 0,clip,width=0.2\columnwidth]{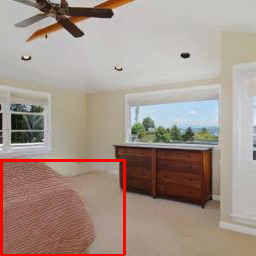} &
        \includegraphics[trim=0 0 0 0,clip,width=0.2\columnwidth]{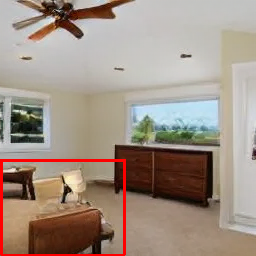} &
        \includegraphics[trim=0 0 0 0,clip,width=0.2\columnwidth]{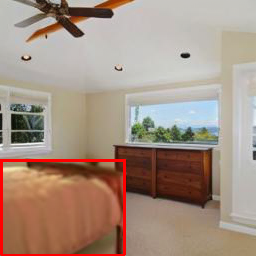} \\

        \includegraphics[trim=0 0 0 0,clip,width=0.2\columnwidth]{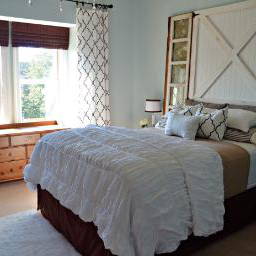} &
        \includegraphics[trim=0 0 0 0,clip,width=0.2\columnwidth]{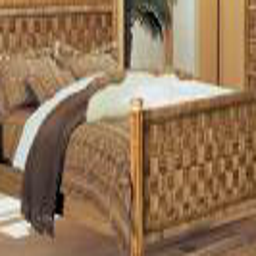} &
        \includegraphics[trim=0 0 0 0,clip,width=0.2\columnwidth]{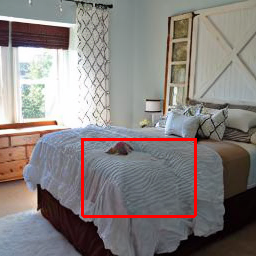} &
        \includegraphics[trim=0 0 0 0,clip,width=0.2\columnwidth]{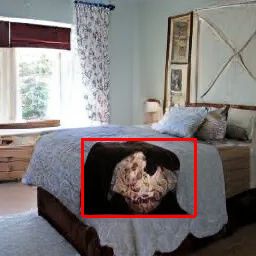} &
        \includegraphics[trim=0 0 0 0,clip,width=0.2\columnwidth]{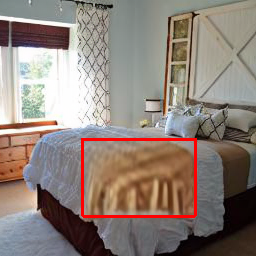} \\

        \includegraphics[trim=0 0 0 0,clip,width=0.2\columnwidth]{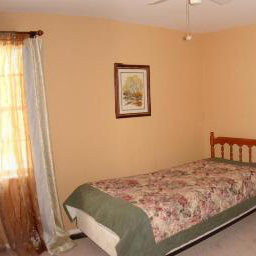} &
        \includegraphics[trim=0 0 0 0,clip,width=0.2\columnwidth]{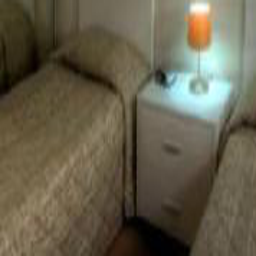} &
        \includegraphics[trim=0 0 0 0,clip,width=0.2\columnwidth]{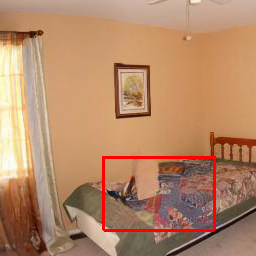} &
        \includegraphics[trim=0 0 0 0,clip,width=0.2\columnwidth]{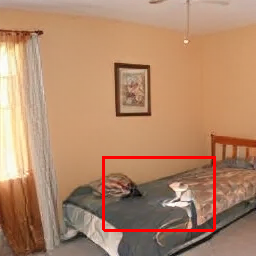} &
        \includegraphics[trim=0 0 0 0,clip,width=0.2\columnwidth]{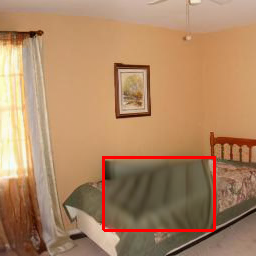} \\

        \includegraphics[trim=0 0 0 0,clip,width=0.2\columnwidth]{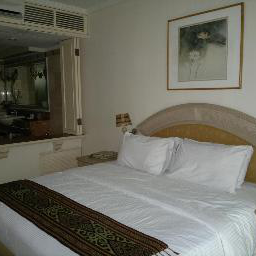} &
        \includegraphics[trim=0 0 0 0,clip,width=0.2\columnwidth]{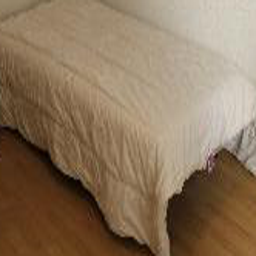} &
        \includegraphics[trim=0 0 0 0,clip,width=0.2\columnwidth]{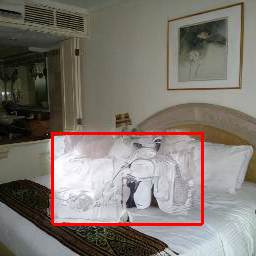} &
        \includegraphics[trim=0 0 0 0,clip,width=0.2\columnwidth]{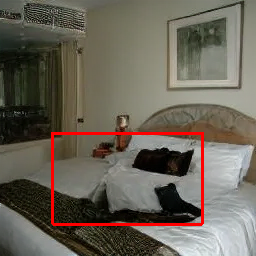} &
        \includegraphics[trim=0 0 0 0,clip,width=0.2\columnwidth]{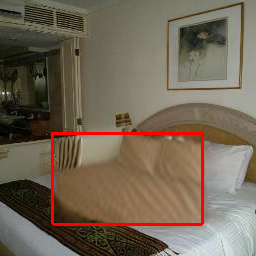} \\
           
\end{tabular}
}
\captionof{figure}{Visual results for appearance control on LSUN bedroom. We show the results obtained with relevant baselines for editing the red area in the input image using the reference as a driver.}
\vspace{-3mm}
\label{fig:local_edits_baselines}
\end{table}

\begin{table}[t]
\centering
\caption{Quantitative results for appearance control on the LSUN Bedroom validation set.}
\begin{tabular}{lccc}
\toprule
\textbf{Model} & \textbf{FID} ($\downarrow$) & \textbf{L1} ($\downarrow$) & \textbf{SSIM} ($\uparrow$) \\
\midrule
Copy-Paste (CP) & 21.37 & 0.0 & 0.87 \\
Inpainting \cite{rombach2022high} & 8.25 & 0.02 & 0.17 \\
CP + Denoise & 9.15 & 0.02 & 0.32 \\ 
E2EVE \cite{Brown2022E2EVE} & 13.59 & 0.05 & 0.34 \\
\rowcolor{gray!20}\textbf{PAIR Diffusion} & 12.81 & 0.02 & 0.51 \\
\bottomrule
\end{tabular}
\label{tab:baselines}
\vspace{-0.2em}
\end{table}

\begin{table}[ht]
    \small
    \centering
    \begin{minipage}[t]{0.49\linewidth}
    \caption{Quantitative results for structure control on CelebA-HQ validation dataset.}
    %\vspace{0.35cm}
    \resizebox{\linewidth}{!}{
    \begin{tabular}{lcc}
        \toprule
        \textbf{Model} & \textbf{mIoU} & \textbf{SSIM} \\ 
        \midrule
         SEAN \cite{zhu2020sean} & 0.64  &  0.32 \\
         \rowcolor{gray!20}\textbf{PAIR Diff.}  & \textbf{0.67}  & \textbf{0.52}\\
        \bottomrule
    \end{tabular}}
    \label{tab:exp_iou}
  \end{minipage}
  \hfill
  \begin{minipage}[t]{0.49\linewidth}
    \caption{Quantitative results of ablation study on appearance representation}\label{wrap-tab:1}
    \resizebox{\columnwidth}{!}{
    \begin{tabular}{lcc}\toprule  
        \textbf{Model} & \textbf{L1}& \textbf{LPIPS} \\ \midrule
         $M_{\text{VGG}}$ &   0.1893 & 0.555\\ 
        $M_{\text{DINO}}$ & 0.1953 & 0.549\\ 
        \rowcolor{gray!20}\textbf{Full} & \textbf{0.1891} &  \textbf{0.545} \\
        \bottomrule
    \end{tabular}}
    \label{tab:}
    \end{minipage}
\vspace{-2em}
\end{table}

\subsection{Quantitative Results}\label{sec:quant}
As described in \cref{sec:method}, the backbone of our design is the ability to control two major properties of the objects, the appearance and the structure.
The aim of the quantitative evaluation is to verify that we can control the mentioned properties and not to push the state-of-the-art results.
We start by evaluating our model on appearance control: the task consists of modifying a specific region of the input image using a reference image to drive the edit. 
We compare our method with the recent work of \cite{Brown2022E2EVE} (E2EVE), and follow their evaluation procedure. In particular, different models are compared based on: (i) Naturalness: we expect the edited image to look realistic and rely on FID between input and edited images to assess it, (ii) Locality: we expect the edit to be limited to the specific region where the edit is performed and use L1 distance to measure it, (iii) Faithfulness: we expect the edited region and the target image to be similar and we use SSIM to evaluate it. As discussed in E2EVE, all the above-mentioned criteria should hold at the same time, and the best-performing method is the one giving good results in the three metrics at the same time.
We compare our method with four baselines: (1) Copy-Paste: the driver image is simply copied in the edit region of the input image, (2) Inpainting: we use LDM \citet{rombach2022high} to inpaint the target edit region, (3) Copy-Paste + Denoise: starting form copy-paste edit, we invert the image with DDIM, and denoise it with LDM, (4) E2EVE. In \cref{tab:baselines} we report the quantitative results on the validation set of LSUN Bedroom \citep{yu2015lsun} and visual comparisons are shown in \cref{fig:local_edits_baselines}. The copy-paste baseline provides an upper bound to the faithfulness and locality but produces images that are unrealistic (high FID score). Vice-versa, Inpainting and CP+Denoise produce natural results (low FID score) but are not faithful to the driver image (low SSIM score). Only our method performs well w.r.t. all the aspects and outperforms E2EVE in all metrics showing that we can control the appearance of a region. We refer the reader to \suppmat for a detailed description of the evaluation procedure and baseline implementation.

Secondly, we evaluate the structure-controlling ability of our method. We adopt the validation set of CelebA-HQ (5000 samples) and compare with SEAN~\citep{zhu2020sean}. We generate images conditioning the model on the ground truth structure maps from the validation set and then segment the generated images with a pre-trained model~\cite{faceparse}. We report the mIoU score, calculated using the ground truth segmentation map as the reference, as well as the SSIM score in \cref{tab:exp_iou}. The proposed method outperforms \cite{zhu2020sean}  in terms of both mIoU and SSIM, demonstrating that our method can precisely follow the guidance of structure and retain the appearance.

\begin{figure*}[!ht]
  \centering
  \includegraphics[width=1\linewidth]{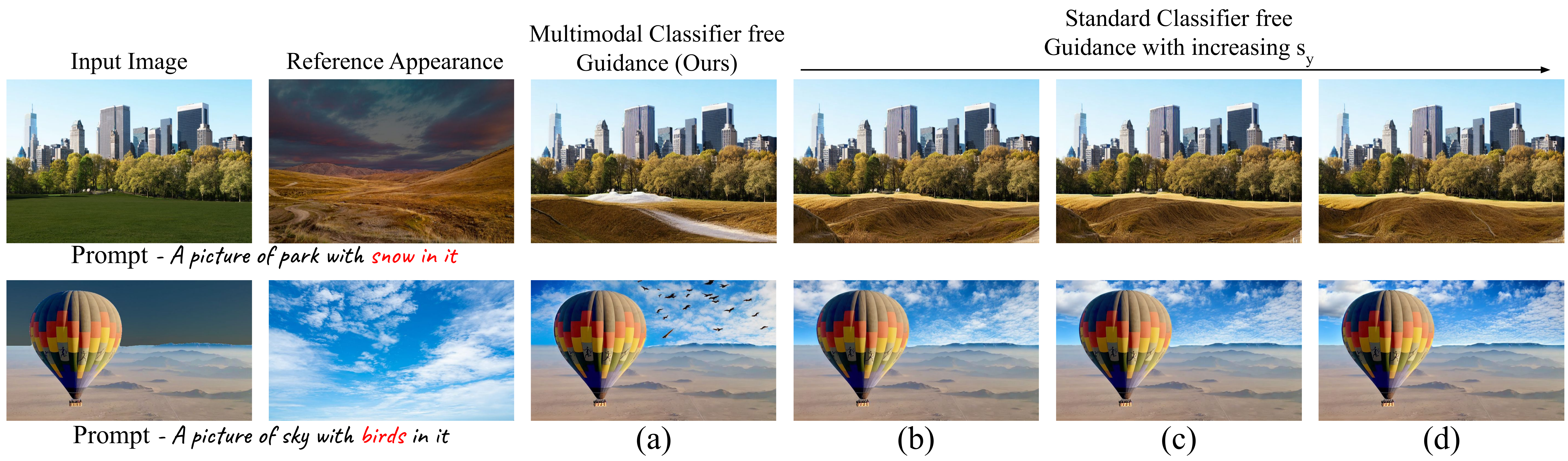}
  \caption{Ablation study for multimodal classifier-free guidance. We can see that if we use standard classifier-free guidance \cref{eq:cfg_abla} the model completely ignores the text when sampling the image. }
  \vspace{-1em}
\label{fig:abla_cfg} 
\end{figure*}

\subsection{Ablation Study}
\label{sec:abla}
\textbf{Multimodal Classifier Free Guidance.} We validate the effectiveness of the proposed multimodal classifier-free guidance. Instead of factorizing, which results in \cref{eq:log_prop}, we directly expand the conditional score function $\nabla_{z_t} \log p(z_t | S, F, y)$ and apply classifier free guidance formulation on it and get the following equation:
\begin{equation}
\begin{aligned}
\label{eq:cfg_abla}
    \tilde{\epsilon_\theta}(z_t, S, F, y) & = \epsilon_\theta(z_t, \phi, \phi, \phi)  \\
     & +s_{S} \cdot(\epsilon_\theta\left(z_t, S, \phi, \phi\right) -\epsilon_\theta\left(z_t, \phi, \phi, \phi\right)) \\
     & + s_{F} \cdot\left(\epsilon_\theta\left(z_t, S, F, \phi\right) - \epsilon_\theta\left(z_t, S, \phi, \phi\right)\right) \\
     & + \highlight{blue}{s_{y} \cdot\left(\epsilon_\theta\left(z_t, S, F, y\right)-\epsilon_\theta(z_t, S, F, \phi)\right)}
% \end{multline*}
\end{aligned}
\end{equation}
We highlight the difference between \cref{eq:cfg} and \cref{eq:cfg_abla} using the \colorbox{blue!17}{blue color}. In \cref{fig:abla_cfg}, we compare the results sampled from \cref{eq:cfg} (column (a)) and \cref{eq:cfg_abla} (columns (b)-(d)). 
%Respectively, column (a) shows results from \cref{eq:cfg} whereas column (b)-(d) shows results from \cref{eq:cfg_abla} with increasing $s_y$. 
We use the same seed to generate all the images, further the values of $s_S, s_F, s_y$ are the same in columns (a) and (b). For the first row we set $s_S\!=\!8, s_F\!=\!3, s_y\!=\!8$ and for second row it is $s_S=6, s_F=4, s_y=8$. The values of $s_y$ for (b)-(d) are $8, 15, 20$ respectively.
We can clearly see that sampling results using \cref{eq:cfg_abla} completely fail to consider text prompt even after increasing the value of $s_y$. This shows the effectiveness of the proposed classifier-free guidance \cref{eq:cfg} for controlling the image in a multimodal manner. Lastly, we conduct an ablation study on control parameters of CFG, namely $s_S, s_F, s_y$,  to better understand the relationship between them. Please refer to \suppmat for discussion and the visual results in \cref{fig:cfg_sf}-\cref{fig:cfg_ft}.

\begin{figure*}[!t]
\centering
\includegraphics[width=1\textwidth]{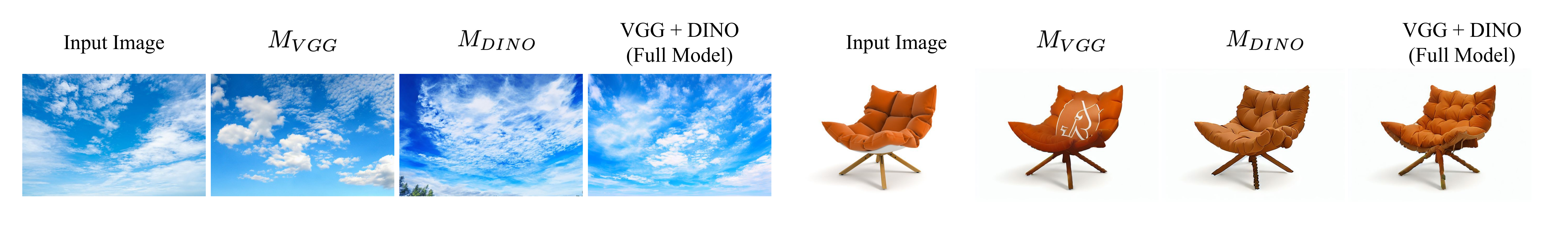}
\caption{\label{fig:app_abla}Object level appearance variations are shown for models trained with different appearance representations. It can be observed that $M_{\text{DINO}}$ faces difficulty in preserving the color when generating variations. The sky is a bit darker than the input image whereas, in the case of the couch, the shade of orange color does not match the input image. We can see that $M_{\text{VGG}}$ is able to preserve the color however it can have artifacts even for slightly complex objects. In the case of the couch, we can see some white artifacts that might have come from the white bottom of the couch in the input image. $M_{\text{VGG}}$ have a poor understanding of objects compared to  $M_{\text{DINO}}$. 
Only the full model which uses both captures the visual aspects of the object faithfully.}
\end{figure*}

\noindent \textbf{Appearance representation.}~We ablate the importance of using VGG and DINOv2 for representing the appearance of an object. We train two models, one using only VGG features ($M_{\text{VGG}}$) and the second using only DINOv2 features ($M_{\text{DINO}}$) to capture the appearance of an object. We train both models using identical hyperparameters to our original model.
We assess the performance of each model using pairwise image similarity metrics on the COCO \cite{caesar2018coco} validation set.
We use L1 as our low-level metric and LPIPS~\citep{zhang2018unreasonable} as our high-level metric and report the results \cref{wrap-tab:1}. While $M_{\text{VGG}}$ has a better L1 score compared to $M_{\text{DINO}}$, the LPIPS score indicates that $M_{\text{DINO}}$ outperforms $M_{\text{VGG}}$. This experiment confirms the intuition that VGG features are good at capturing low-level details, while DINO features excel at capturing high-level details in our framework. In our final design, we found that combining both VGG and DINOv2 features for appearance vectors yielded the best L1 and LPIPS scores, leveraging the strengths of both representations. Supporting visuals illustrating this can be found in Fig.~\ref{fig:app_abla}.

%As discussed in \cref{sec:intro}, our definition of appearance does not aim to maintain the exact identity of the object in the reference image. 
We define appearance as the visual characteristics of an object (see \cref{sec:intro}), and do not aim to maintain the exact identity of the object in the reference image.
This is in contrast with recent research on personalization \cite{ruiz_dreambooth_2022}. However, this formulation allows us to employ reference images in a versatile manner, contributing to our comprehensive editing capabilities. We use reference images that depict texture-only images (\cref{fig:front}(a) second row), perform image variations (\cref{fig:sd_variations}), realistic edits (\cref{fig:app_edit}), style transfer (\cref{fig:local_od_apperance_supp}) as well as semantically complex edits (\cref{fig:sd_app_edits}).

% Importantly, we define the appearance as the visual characteristics of an object, such as texture, color, and pattern \cite{gong2019autogan, zeiler2014visualizing}, and do not aim to maintain the exact identity of the object in the reference image. This is in contrast with recent research on personalization \cite{ruiz_dreambooth_2022}, but allows us to employ reference images in a versatile manner, contributing to the desired comprehensive editing capabilities. Moreover, by incorporating the ability to control the structure along with the appearance, we can achieve a wide range of results including
%Only when both features are used together we get the best L1 and LPIPS scores, getting the best of both of representations. Hence, in our final design, we used both VGG and DINOv2 features for appearance vectors. Supporting visuals can be found in Fig~\ref{fig:app_abla}.

% \change{We also conduct ablation study on control parameters of CFG, namely $s_S, s_F ,s_y$,  to better understand the relationship between them. Please refer to \suppmat for discussion. Visual results can be found in Fig~\ref{fig:cfg_ft} and ~\ref{fig:cfg_sf}.}

\section{Conclusion}
In this paper, we show that we can build a comprehensive image editor by interpreting images as the amalgamations of various objects. 
We propose a generic framework, dubbed PAIR Diffusion, that enables structure and appearance editing at the object-level in any diffusion model. Our framework enables various object-level editing operations on real images without the need for inversion including appearance editing, structure editing, adding objects, and object variations. All the operations can be obtained with a single model trained once. Furthermore, we propose multimodal classifier-free guidance which enables precise control in the editing operations. We demonstrate its effectiveness with extensive editing examples on real image across different domains.

%We validate the framework's efficacy by showing extensive editing results using diverse domains of real images. 

\noindent \textbf{Limitations and future work.} Currently, the architecture modifications present a simple formulation of the appearance vectors and the structure conditioning. While offering advantages by seamlessly integrating into existing Diffusion Models with minimal modification, in the future we plan to explore more sophisticated designs while maintaining the core object-level formulation. We plan to extend the explicit control over other aspects of the objects, such as the illumination, pose, etc., and improve the identity preservation of the edited object. The proposed object-level formulation can also help in devising standardizing metrics for image editing tasks in a unified manner which is lacking in the field. 

\noindent \textbf{Acknowledgement.}~This work was partly supported by the MUR PNRR project FAIR (PE00000013) funded by the NextGenerationEU and by the EU Horizon project AI4Trust (No. 101070190), NSF CAREER Award \#2239840, and the National AI Institute for Exceptional Education (Award \#2229873) by National Science Foundation and the Institute of Education Sciences, U.S. Department of Education.
{
    \small
    \bibliographystyle{ieeenat_fullname}
    \bibliography{main}
}

\clearpage
\setcounter{page}{1}
\maketitlesupplementary
%\section*{Appendix}
This appendix contains three sections organized as follows. In \cref{app:mcfg} we discuss our proposed multimodal classifier free guidance, proving \cref{eq:cfg} of the main paper and providing qualitative results for its usage. In \cref{app:implementation}, we report the implementation details of our unconditional model along with details about the baselines adopted in the paper. Lastly, in \cref{app:qualitatives} we show additional qualitative results and applications of our framework.

\section{Multimodal Classifier Free Guidance}
\label{app:mcfg}

\subsection{Proof}
Let us represent $p(z|c)$ as the distribution we want to learn. In our case, the conditioning $c = (S, F, y)$, while $z$ represents the input image $x$ in latent space. A well-trained diffusion model learns to estimate $ \nabla_{z_t}  \log p(z_t|c)$. %Let us focus on $  \log p(z_t|c)$. 
Using Bayes' rule and taking the logarithm, we obtain:

\begin{equation}
    \label{eq:proof}
     \log p(z_t|c) =   \log p(c|z_t) +  \log p(z_t) -  \log p(c)
\end{equation}

In our setting of multimodal conditioning, we want to control the final image using the appearance from the reference image $F$ and the text prompt $y$ \emph{independently}. Multimodal inference would be particularly useful when the information in $y$ is disjoint from $(S, F)$, and we need to capture it in the final image. Following this assumption, we obtain:

\begin{equation}
    \label{eq:proof_assum}
    \begin{split}
     \log p(c) & = \log p(S, F, y) \\
               & = \log p(S, F) +  \log p(y)
    \end{split}
\end{equation}

Similarly, expanding $ \log p(c|z_t)$ and using \cref{eq:proof_assum}:

\begin{equation}
    \label{eq:proof_1}
    \begin{split}
     \log p(c|z_t) &=  \log p(S, F, y|z_t) \\
                   & =  \log p(S, F|z_t) +  \log p(y|z_t)
    \end{split}
\end{equation}
% In the above equation we assume that $ \log p(S, F|z_t, y) \approx  \log p(S, F|z_t)$
Plugging the \cref{eq:proof_assum}-\cref{eq:proof_1} in \cref{eq:proof}, and applying Bayes' rule: 

% \begin{equation}
%     \label{eq:proof_2}
%     \begin{aligned}
%      \log p(z_t|S, F, y) =   \log p(S, F|z_t) +  \log p(y|z_t) +  
%  \\ \log p(z_t) -  \log p(S, F, y) \\
%      \log p(z_t|S, F, y) =   \log p(S, F|z_t) +  \log p(y|z_t)  +  \\ \log p(z_t) - \log p(S, F) -  \log p(y)\\
%      \log p(z_t|S, F, y) =   \log p(z_t|S, F) +  \log p(z_t|y) -  \log p(z_t) 
%     \end{aligned}
% \end{equation}
\begin{equation}
    \label{eq:proof_2}
    \begin{aligned}
     \log p(z_t|S, F, y) & =   \log p(S, F|z_t) +  \log p(y|z_t) \\ 
     & \quad\quad + \log p(z_t) -  \log p(S, F, y) \\
     & =   \log p(S, F|z_t) +  \log p(y|z_t)  \\ 
     & \quad\quad +\log p(z_t) - \log p(S, F) -  \log p(y)\\
     & =   \log p(z_t|S, F) +  \log p(z_t|y) -  \log p(z_t) 
    \end{aligned}
\end{equation}

Taking gradient w.r.t. $ \nabla_{z_t}$ we can get 

\begin{equation}
\label{eq:basic_deri}
\begin{aligned}
 \nabla_{z_t}  \log p(z_t|S, F, y) & = \highlight{orange}{\nabla_{z_t}  \log p(z_t|S, F)} \\ 
                                   & + \highlight{green}{\nabla_{z_t}  \log p(z_t|y)} \\ 
                                   & -  \nabla_{z_t}  \log p(z_t) 
 \end{aligned}
\end{equation}

% Next, we apply Bayes rule to $\nabla_{z_t}\log p(z_t|S, F)$ and $ \nabla_{z_t}  \log p(z_t|y)$
Next, we apply Bayes' rule to the first term:

\begin{equation}
    \label{eq:proof_3}
    \begin{aligned}
        \highlight{orange}{\nabla_{z_t}\log p(z_t|S, F)} & =  \nabla_{z_t}  \log p(F|S, z_t) \\ 
                                       & +  \nabla_{z_t}  \log p(S| z_t)  \\
                                       & +  \nabla_{z_t}  \log p(z_t) \\
                                       & - \cancelto{0}{ \nabla_{z_t}  \log p(S, F)}
    \end{aligned}
\end{equation}

and the second term:

\begin{equation}
    \label{eq:proof_4}
    \begin{aligned}
    \highlight{green}{\nabla_{z_t}  \log p(z_t|y)} & =  \nabla_{z_t}  \log p(y| z_t) \\
                                & + \nabla_{z_t}  \log p(z_t) \\ 
                                & -  \cancelto{0}{ \nabla_{z_t}  \log p(y)}
    \end{aligned}
\end{equation}

Pluggin \cref{eq:proof_3}-\cref{eq:proof_4} in \cref{eq:basic_deri} we obtain:

\begin{equation}
    \label{eq:proof_5}
    \begin{aligned}
    \nabla_{z_t}  \log p(z_t|S, F, y) & =\highlight{orange}{\nabla_{z_t}  \log p(F|S, z_t)}  \\ 
                                      & + \highlight{orange}{\nabla_{z_t}  \log p(S| z_t)} \\
                                      & + \highlight{green}{\nabla_{z_t}  \log p(y| z_t)} \\
                                      & +  \highlight{green}{\nabla_{z_t}\log p(z_t)}
    \end{aligned}
\end{equation}

 We can use the concept of classifier-free-guidance~\cite{ho_classifier-free_2022} to approximate the above equation using a single model which has been trained by dropping the conditions during training and get the final sampling equation \cref{eq:cfg}:

\begin{equation}
\begin{split}
\label{eq:cfg_deriv}
    \tilde{\epsilon_\theta}(z_t, S, F, y) & = \epsilon_\theta(z_t, \phi, \phi, \phi)  \\
                                          & + s_{S} \cdot(\epsilon_\theta\left(z_t, S, \phi, \phi\right)) - \epsilon_\theta\left(z_t, \phi, \phi, \phi\right)) \\
                                          & + s_{F} \cdot\left(\epsilon_\theta\left(z_t, S, F, \phi\right) - \epsilon_\theta\left(z_t, S, \phi, \phi\right)\right) \\
                                          & + s_{y} \cdot\left(\epsilon_\theta\left(z_t, \phi, \phi, y\right) - \epsilon_\theta(z_t, \phi, \phi, \phi)\right)
\end{split}
\end{equation}

\subsection{Ablation on CFG control parameters}
The multimodal classifier free guidance has three control parameters namely $s_S, s_F, s_y$. In practice, the values $s_S, s_F, s_y$ can be understood as the guidance strengths to control how the final image is affected by the structure, reference image, and the given text prompt. When using it to edit real-world images it is crucial to understand how they change the output image when varied together. In this section, we study the effect of (a) varying structure guidance $s_S$ and reference image guidance $s_F$  (b) varying text prompt guidance $s_y$ and reference image guidance $s_F$.

\begin{figure*}[!ht]
  \centering
  \includegraphics[width=\linewidth]{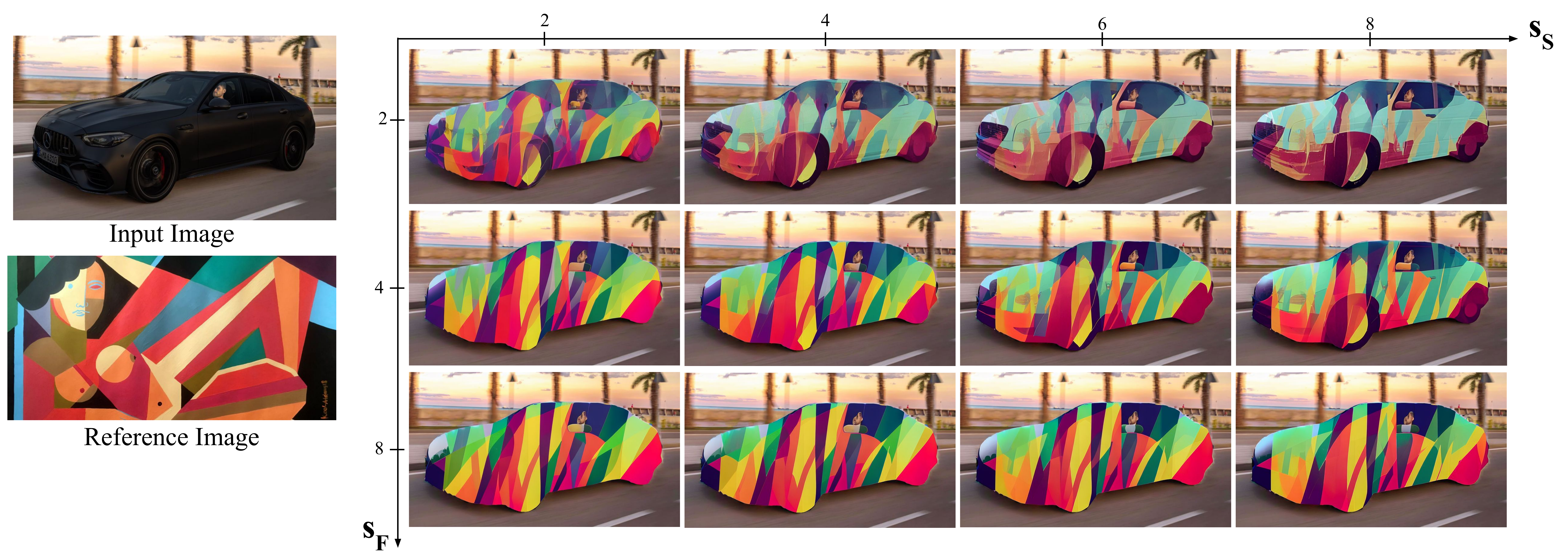}
  \caption{The figure shows the affect of $s_F$ and $s_S$ parameter in multimodal classifier free guidance Eq.~\ref{eq:cfg} when they are varied. We can see that as we increase $s_F$ the output forcefully imposes the appearance from the reference image and the car starts losing the structure details that were helping to make it look like a car. In order to maintain the structural integrity of the car we need to increase  $s_S$.
  }
\label{fig:cfg_sf} 
\end{figure*}

\noindent \textbf{Structure ($s_s$) and Appearance ($s_F$).}~We explain the importance of parameters by choosing a reference image that is completely different from the underlying structure in the input image. The results are shown in Fig.~\ref{fig:cfg_sf}. We can observe that as we increase $s_F$ the model forcefully imposes the reference image appearance on the object being edited and the object starts losing its structural integrity. We can get back the structural integrity when increasing $s_S$ as well. Notice, in the second row, when $s_F = 4$ and $s_S = 2$ we cannot see any parts of the car such as the wheel, headlight, windshield, \etc. When we start increasing $s_S$ the subpart of the car starts appearing the edited region starts looking more like a car. However, when we increase $s_F$ too much as in the last row, even after increasing $s_S$ does not help much. In general, it is good practice to keep $s_S > s_F$ when editing real images. Further, we can adjust $s_F$ how closely we want the edited output to follow the reference appearance.

\begin{figure*}[!ht]
  \centering
  \includegraphics[width=\linewidth]{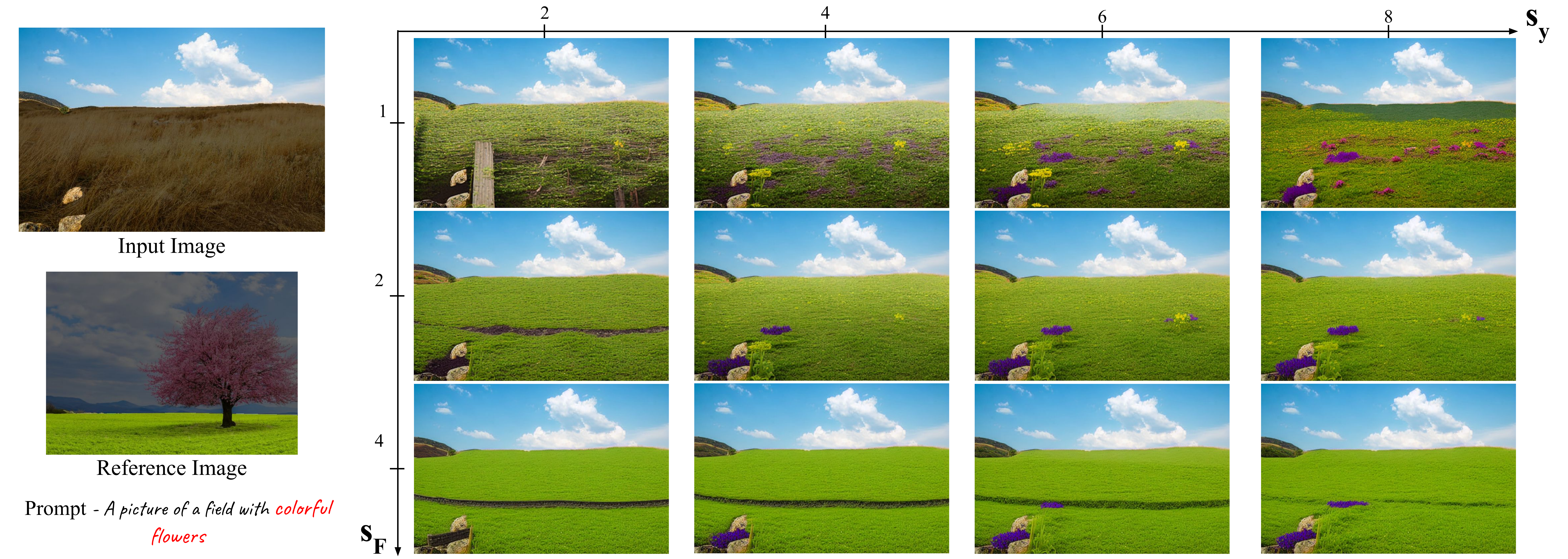}
  \caption{The figure shows the effect of $s_F$ and $s_T$ parameter in multimodal classifier free guidance Eq.~\ref{eq:cfg} when they are varied. $s_F$ controls the effect of the reference image and $s_y$ control the effect of the prompt.
  We can see that the output is highly sensitive to $s_F$. Even if we increase $s_F$ slightly the effect of the prompt starts disappearing. When we want to use both reference images and prompts to affect the final image it is best to keep $s_y > s_F$ and $s_F$ should have a low value in an absolute sense. }
\label{fig:cfg_ft} 
\end{figure*}

\noindent  \textbf{Text Prompt ($s_y$) and Appearance ($s_F$).}~We analyze the effect of two crucial guidance parameters that help us in controlling and editing images in a multimodal manner.  The results are shown in \cref{fig:cfg_ft}. We can see that it is crucial to keep $s_y > s_F$ to see the effects of the text prompt on the output image. Further, we can see that the model is more sensitive to the $s_F$ parameter compared to $s_y$. Even if we increase $s_F$ slightly we can see diminishing effects of the prompt on the edited image. In general, when editing images in a multimodal manner it is a good practice to keep $s_y > s_F$, $s_F$ should have a low value in an absolute sense. Keeping these constraints we can vary the parameters to adjust the final output as needed.

\section{Implementation Details}\label{app:implementation}
\subsection{PAIR Diffusion}
\textbf{Unconditional Diffusion Model.} In this section, we provide additional details for implementing PAIR Diffusion framework on unconditional diffusion models. We used LDM~\cite{rombach2022high} as our base architecture and trained on LSUN Church, Bedroom, and CelebA-HQ  datasets. To extract the structure information, we apply SeMask-L~\cite{jain2021semask} with Mask2former~\cite{cheng2022masked} trained on ADE20K~\cite{zhou2017scene}, and compute the segmentation mask for LSUN Church
and Bedroom datasets~\cite{yu2015lsun}. In CelebA-HQ, ground truth segmentation masks are available. 
Given the simplicity of the architecture and the training datasets, we found that simply using the features extracted by the VGG network ($G^{Vl_1}$, see \cref{sec:method}) to be sufficient in this case to achieve various editing capabilities. To condition the model on this information, we simply concatenate $G^{Vl_1}$ to the noisy latent $z_t$ along the channels dimension. We increase the number of channels of the first convolutional layer of the UNet from $C_{in}$ to $C_{in}+C+1$, with $C$ the number of channels in $G^{Vl_1}$, and keep the rest of the architecture as in \cite{rombach2022high}. % where $l_1 = 1$. 
For all three datasets namely, LSUN Church, Bedroom, and CelebA-HQ we start with the pre-trained weights provided by LDM~\cite{rombach2022high} and finetune with the same hyperparameters mentioned in the paper~\cite{rombach2022high}. The number of steps, learning rate, and batch size are reported in \cref{tab:hyper}. We train our models using  A100 GPUs. During training, we randomly dropped structure and appearance conditioning with a probability of $10 \%$.
At inference time, we adapt \cref{eq:cfg} for sampling from the model by setting $s_y = 0$ and use classifier-free guidance style sampling using the DDIM algorithm \cite{song2019generative} with 250 steps.

\begin{table*}
\centering
\begin{tabular}{llcccc}
\toprule
\textbf{Base Model} & \textbf{Dataset} & \textbf{LR} & \textbf{Batch Size} & \textbf{Iterations} & \textbf{GPU Days} \\
\midrule
LDM & Bedroom & 9.6e-5 & 48 & 750k & 24 \\
LDM  & Churches & 1.0e-5 & 96 & 350k  & 12 \\
LDM  & CelebA-HQ & 9.6e-5 & 24 & 120k &  1\\ 
SD +  ControlNet  & COCO & 1.5e-5 & 128 & 86k  & 96\\ 
\bottomrule
\end{tabular}
\caption{Hyperparameters for PAIR diffusion when used with LDM~\cite{rombach2022high} and Stable Diffusion (SD) with ControlNet~\cite{zhang2023adding}.}
\label{tab:hyper}
\end{table*}

\noindent  \textbf{Foundational Diffusion Model.}~We use Stable Diffusion (SD)~\cite{rombach2022high} as our base architecture and ControlNet~\cite{zhang2023adding} to efficiently condition SD. We use COCO~\cite{caesar2018coco} dataset for the experiments, which contains panoptic segmentation masks and image captions. In vanilla ControlNet, the conditioning signal is passed through a zero convolution network and added to the input noise after being passed through the first encoder block of the control module. Similarly, we utilize $G^{Vl_1}$ in the same fashion, employing it as input to the control module as in the standard ControlNet approach. Furthermore, we modulate the features in the first encoder block of the control module using $G^{Dl_2}$ and in the second encoder block using $G^{Dl_3}$ by adding them to the respective features after cross-attention blocks. Before adding $G^{Dl_2}, G^{Dl_3}$ we pass them through two linear layers to match the dimension of the features of the network. We train the network as described in ControlNet~\cite{zhang2023adding}, training the control module and the linear layers while keeping all the other parameters frozen. We perform a grid search to identify the layers of the VGG and DINOv2 that achieve the best results. We found $l_1 = 1, l_2 = 6, l_3 = 18$ to yield the best results. During training, we randomly dropped structure, appearance, and text conditioning with a probability of $10 \%$.
Our model is trained across 4 A100 machines with 8 GPUs requiring 3 days to train the model. The number of steps, learning rate, and batch size are reported in \cref{tab:hyper}.
At inference time, we apply \cref{eq:cfg} for sampling and use classifier-free guidance style sampling using the DDIM algorithm \cite{song2019generative} with 20 steps.

 % % TODO: move it to the appropriate placve

\subsection{Baselines}

\noindent \textbf{Quantitative Experiments.} We provide additional details about implementation and evaluation procedures for the results shown in \cref{sec:quant} of the main paper.
We evaluate the models on the task of in-domain appearance manipulation. We first describe the data-collection procedure. We use 5000 images from the validation set of LSUN Bedroom, and choose the bed as the object to edit. For each image, we randomly select a patch within the bed, and use a patch extracted in the same way from another image as the driver for the edit. Next, we describe the baselines.
\textbf{Copy-Paste}: the target patch is copied and pasted in the target region of the input image, resizing the patch to fit the target region. \textbf{Inpainting}, we use the model pretrained on LSUN Bedroom Dataset by \cite{rombach2022high}, and use it to inpaint the edit region. To do that, we use a masked sampling technique, as done in the inpainting task in ~\citet{rombach2022high}. \textbf{CP+Denoise}, we start from the results of copy-paste and apply DDIM Inversion to map the image to the diffusion noise space \cite{song2019generative}. Subsequently, we apply LDM to denoise the image to the final result. Lastly, we compare our method with \textbf{E2EVE}~\cite{Brown2022E2EVE}. We use the original pre-trained weights shared by the authors and use their model to perform the edit. 
Next, we detail the metrics calculation pipeline. We compute \emph{naturalness} by measuring the FID between the edited images and the original images from the whole dataset. We estimate the \emph{locality}, by measuring the L1 loss between the original image and the edited image outside the edited region (\ie the region that should not change). Finally, the \emph{faithfulness} is measured by the SSIM between the driver image and the edited region in the edited image.
% We provide an additional qualitative comparison between the baselines and our method in Fig.\ref{fig:local_edits_baselines_supp}.

\noindent \textbf{Qualitative Experiments.}~We run Prompt-Free-Diffusion~\cite{xu2023prompt} following the description in Sec.~4.5 of the paper. For implementation, we follow the author's instructions at the following \href{https://github.com/SHI-Labs/Prompt-Free-Diffusion/issues/3#issuecomment-1573091747}{GitHub issue}. Specifically, we crop the input image around the object being edited and the reference image around the selected object. We feed the two crops to the SeeCoder and use the segmentation ControlNet with the segmentation map of the cropped input image as the conditioning signal. We then get the edited image by cutting the region of interest in the output image and pasting it in the input image. Regarding Paint-By-Example \cite{yang2023paint}, we crop the reference image around the selected object and follow the original inference procedure afterward.

\section{Qualitative Results}\label{app:qualitatives}

\subsection{Stable Diffusion Results}
In this section, we show additional results for PAIR Diffusion when coupled with a foundational diffusion model like Stable Diffusion \cite{rombach2022high}. In \cref{fig:sd_app_edits}, we perform appearance manipulation in the wild, showing realistic edits in different scenarios.  In \cref{fig:sd_add_objects}, we provide additional results for the task of adding a new object to a given scene. Lastly, we showcase another capability of our model, \ie producing variations of a given object. Specifically, we sample different initial latent codes $z_i \sim \mathcal{N}(0, 1)$, while fixing the structure and appearance representation. We report the results in \cref{fig:sd_variations}.

\subsection{Unconditional PAIR Diffusion}
We start by providing additional results for the task of appearance control. 
In \cref{fig:local_edits_indomain} (a), we can notice that our method can easily transfer the appearance of a church from a completely different structure in the reference image to the structure of the church in the input image. At the same time, we can copy relatively homogeneous regions like the sky, transferring the color accurately, as well as more textured objects such as the trees. In \cref{fig:local_edits_indomain} (b), it is interesting to note that, when we change the style of the floor, the model can appropriately place the reflections hence realistically harmonizing the edit with the rest of the scene. 
Similar observations can be made when we edit the wall. Lastly, in \cref{fig:local_edits_indomain} (c) we show results on faces. We can observe that our method accurately transfers the appearance from the reference image, modifying the skin, hair, and eyebrows of the input. We notice that all our edits do not alter the identity of the person in the input image, which is a desirable property when editing faces.

\begin{table*}[!tb]
    \centering
    \def\arraystretch{0.0}
    \resizebox{\linewidth}{!}{
    \setlength\tabcolsep{0.0pt}
    \footnotesize
    \renewcommand{\arraystretch}{0.0}
    \begin{tabular}{c@{\hskip 2.0pt}c@{\hskip 0.5pt}c@{\hskip 2.0pt}c@{\hskip 0.5pt}c@{\hskip 2.0pt}c@{\hskip 0.5pt}c}

            Input &  \multicolumn{2}{c}{Building} &  \multicolumn{2}{c}{Sky} &  \multicolumn{2}{c}{Tree} \\
            \includegraphics[trim=2 0 0 2,clip,width=0.18\columnwidth]{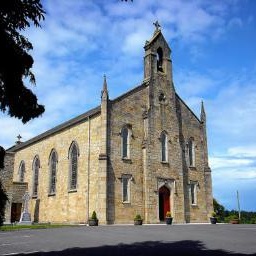} &
            \includegraphics[trim=2 0 0 2,clip,width=0.18\columnwidth]{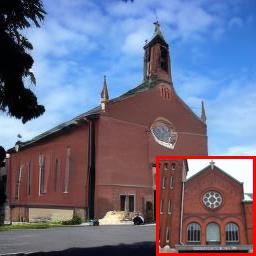} &
            \includegraphics[trim=2 0 0 2,clip,width=0.18\columnwidth]{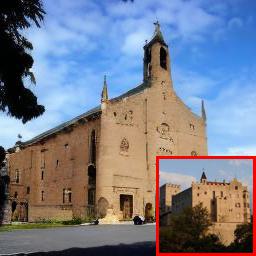} &
            \includegraphics[trim=2 0 0 2,clip,width=0.18\columnwidth]{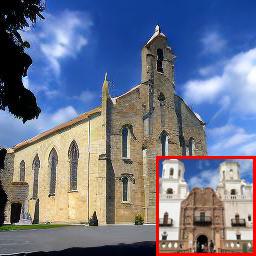} &
            \includegraphics[trim=2 0 0 2,clip,width=0.18\columnwidth]{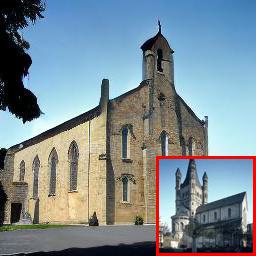} &
            \includegraphics[trim=2 0 0 2,clip,width=0.18\columnwidth]{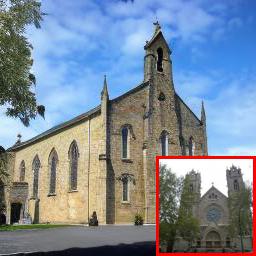} &
            \includegraphics[trim=2 0 0 2,clip,width=0.18\columnwidth]{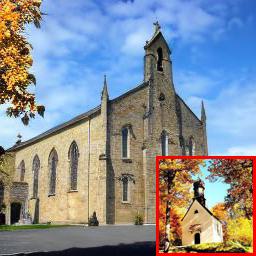} \\
            \multicolumn{7}{c}{(a)} \\

            Input &  \multicolumn{2}{c}{Bed} &  \multicolumn{2}{c}{Wall} &  \multicolumn{2}{c}{Floor} \\
            \includegraphics[trim=2 0 0 2,clip,width=0.18\columnwidth]{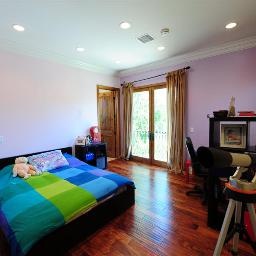} &
            \includegraphics[trim=2 0 0 2,clip,width=0.18\columnwidth]{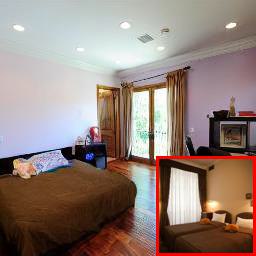} &
            \includegraphics[trim=2 0 0 2,clip,width=0.18\columnwidth]{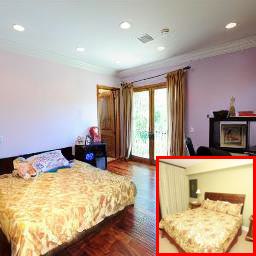} &
            \includegraphics[trim=2 0 0 2,clip,width=0.18\columnwidth]{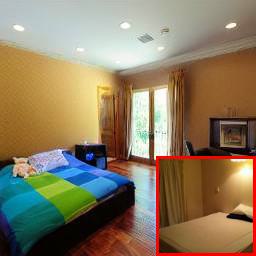} &
            \includegraphics[trim=2 0 0 2,clip,width=0.18\columnwidth]{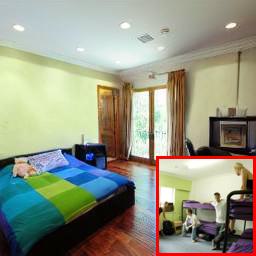} &
            \includegraphics[trim=2 0 0 2,clip,width=0.18\columnwidth]{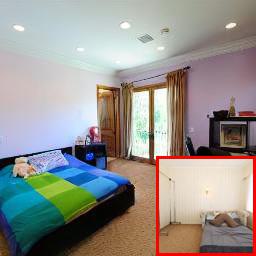} &
            \includegraphics[trim=2 0 0 2,clip,width=0.18\columnwidth]{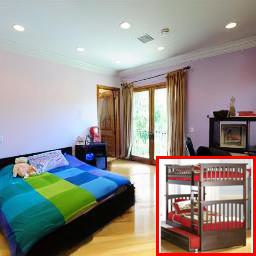} \\
            \multicolumn{7}{c}{(b)} \\
    
            Input &  \multicolumn{2}{c}{Skin} &  \multicolumn{2}{c}{Hair} &  \multicolumn{2}{c}{Eyebrows} \\
            \includegraphics[trim=2 0 0 2,clip,width=0.18\columnwidth]{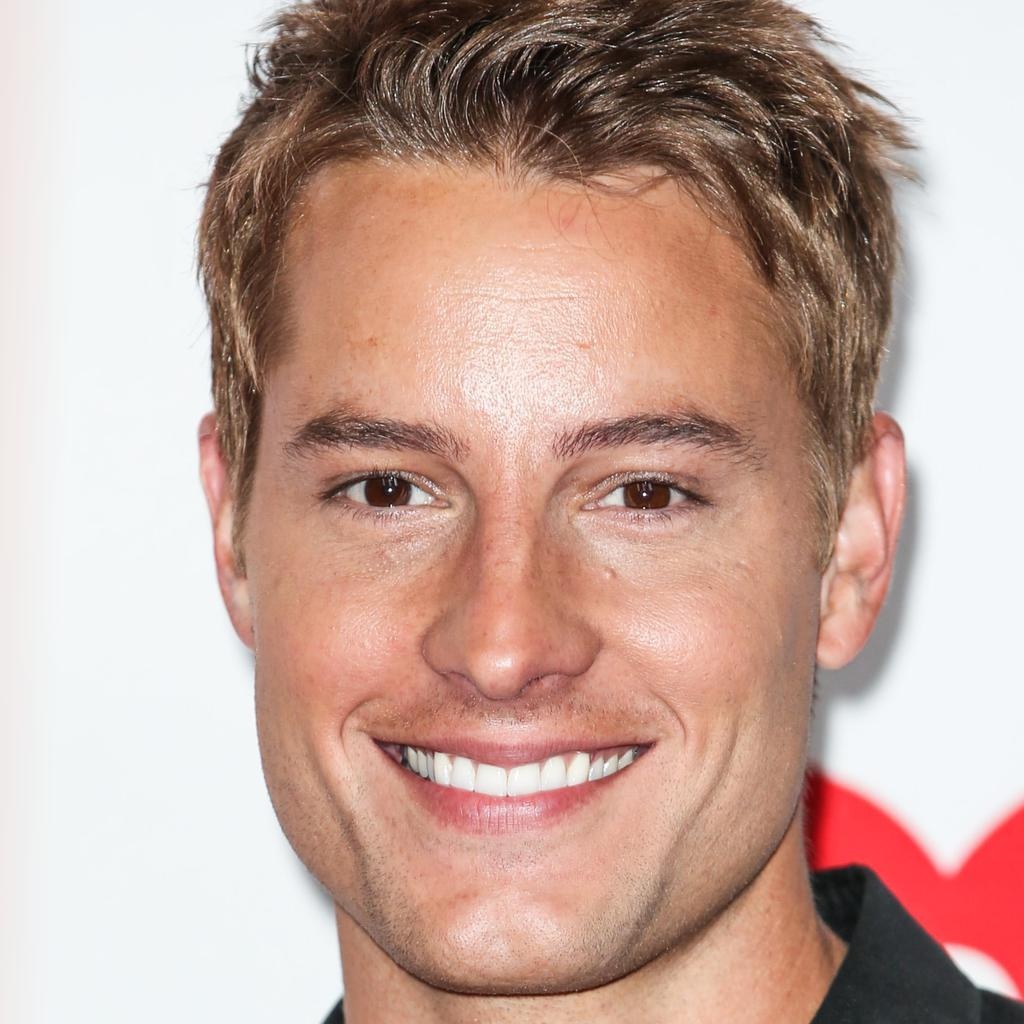} &
            \includegraphics[trim=2 0 0 2,clip,width=0.18\columnwidth]{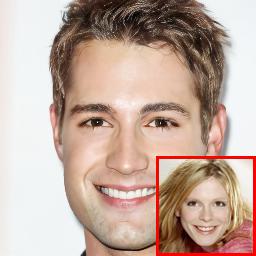} &
            \includegraphics[trim=2 0 0 2,clip,width=0.18\columnwidth]{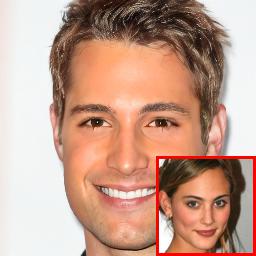} &
            \includegraphics[trim=2 0 0 2,clip,width=0.18\columnwidth]{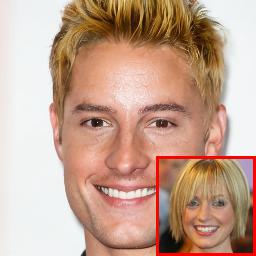} &
            \includegraphics[trim=2 0 0 2,clip,width=0.18\columnwidth]{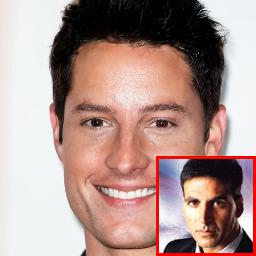} &
            \includegraphics[trim=2 0 0 2,clip,width=0.18\columnwidth]{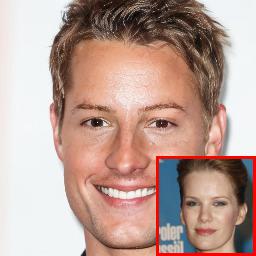} &
            \includegraphics[trim=2 0 0 2,clip,width=0.18\columnwidth]{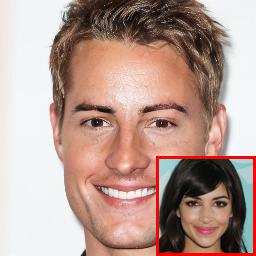} \\
            \multicolumn{7}{c}{(c)} \\
    \end{tabular}
    }
    \vspace{-3mm}
    \captionof{figure}{Qualitative results of Appearance Manipulation using UC-PAIR diffusion model. Reference images are shown in the bottom right, while the text on top indicates the object targeted by the editing operation.}
    \label{fig:local_edits_indomain}
\end{table*}

\begin{figure*}[b]
  \centering
  \includegraphics[width=0.9\linewidth]{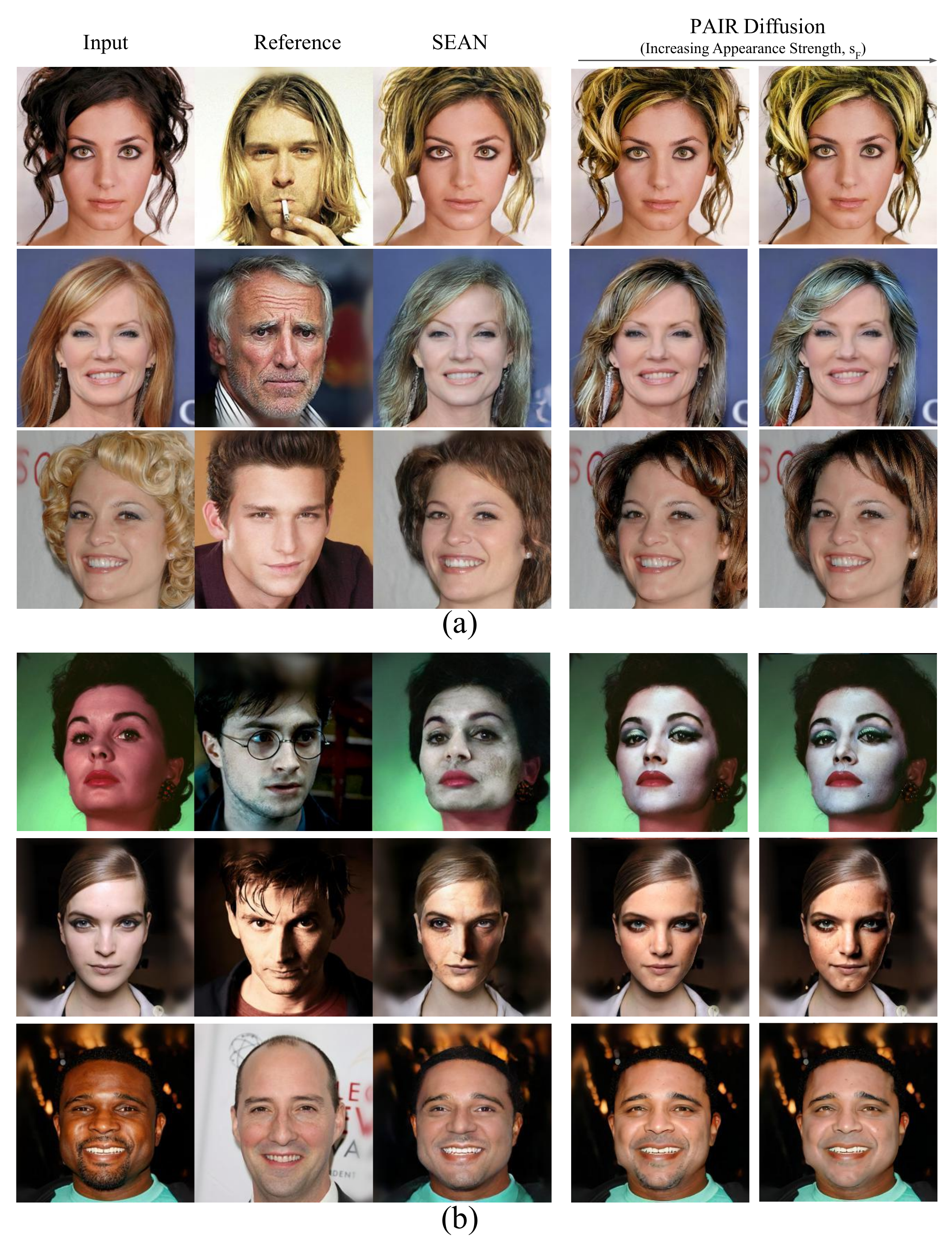}
  \caption{ Qualitative comparison against SEAN \cite{zhu2020sean} for editing the appearance of (a) Hair (b) Skin. Unlike SEAN~\cite{zhu2020sean} we can also control the edit strength using our proposed classifier free guidance \cref{eq:cfg}}
\label{fig:sean_comparison} 
\end{figure*}

Next, we provide an additional qualitative comparison for the task of appearance control with the baselines detailed in \cref{app:implementation}.
% Next, we provide additional qualitative results for the task of appearance control as described in Sec.~\ref{sec:quant} of the main paper, showing the performance of the different baselines detailed in Sec.\ref{app:implementation}.
In \cref{fig:local_edits_baselines_supp}, we can observe that our method seamlessly transfers the appearance from the reference image to the input image, while maintaining the edit to the targeted region. 
Moreover, we show the qualitative comparison with SEAN~\cite{zhu2020sean} in \cref{fig:sean_comparison}. We can see that our method gives better editing results and we also allow to control the strength of the edit.  In \cref{fig:inter_supp}, we showcase more nuanced appearance editing results instead of simply swapping the appearance of input and reference images (\ie $f'_i = f_i^{R}$) by linearly combining the two. We exploit the flexibility of our formulation by setting $a_0=\lambda, a_1=1-\lambda$, with $\lambda \in [0, 1]$, \ie interpolating input and reference images. We can notice how the appearance of the edited region smoothly transitions from the original appearance to the reference, providing an additional level of control for the end user. 

Lastly, we present a more challenging editing scenario using reference images that contain no semantics (\eg abstract paintings) and use it to perform both localized and global editing in \cref{fig:local_od_apperance_supp}.

\begin{table*}[t]
    \centering
    \def\arraystretch{0.5}
    \resizebox{0.95\linewidth}{!}{
    \setlength\tabcolsep{0.5pt}
    \footnotesize
    \begin{tabular}{ccc}
            \scalebox{.5}{Input} &  \scalebox{.5}{Reference Appearance} &  \scalebox{.5}{\textbf{PAIR Diffusion}} \\
            \includegraphics[trim=2 0 0 2,clip,width=0.25\columnwidth]{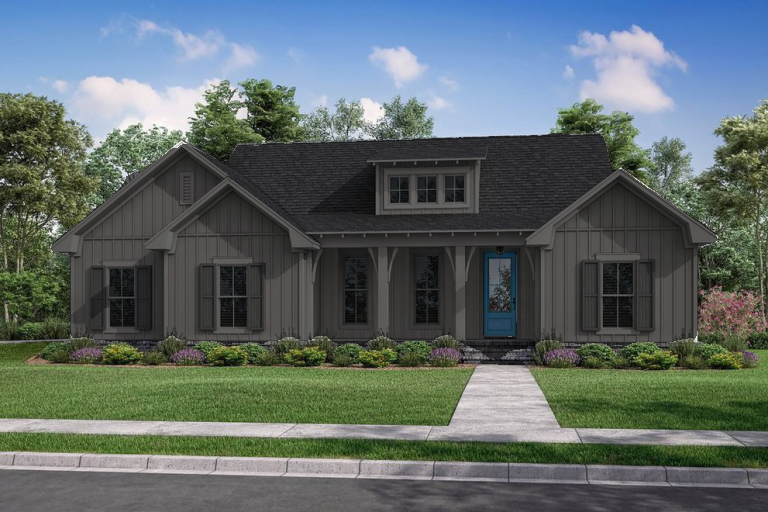} &
            \includegraphics[trim=2 0 0 2,clip,width=0.25\columnwidth]{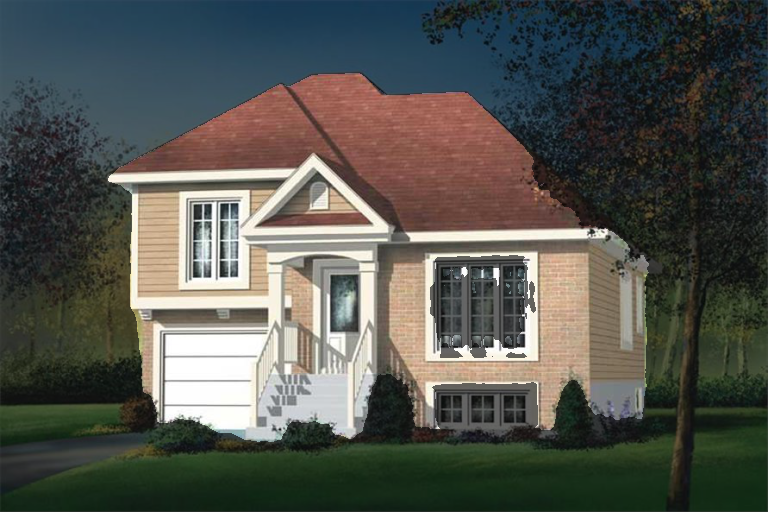} &
            \includegraphics[trim=2 0 0 2,clip,width=0.25\columnwidth]{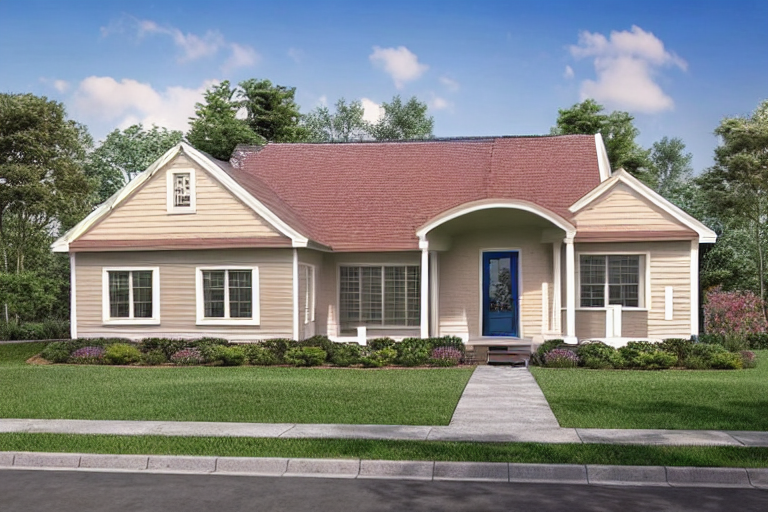} \\

            \includegraphics[trim=2 0 0 2,clip,width=0.25\columnwidth]{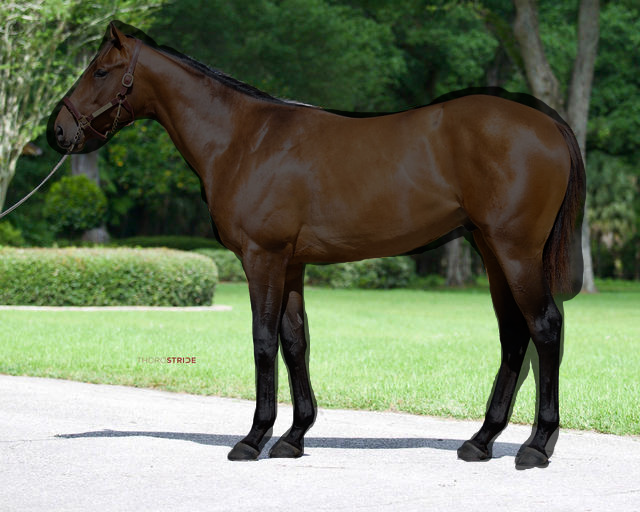} &
            \includegraphics[trim=2 0 0 2,clip,width=0.25\columnwidth]{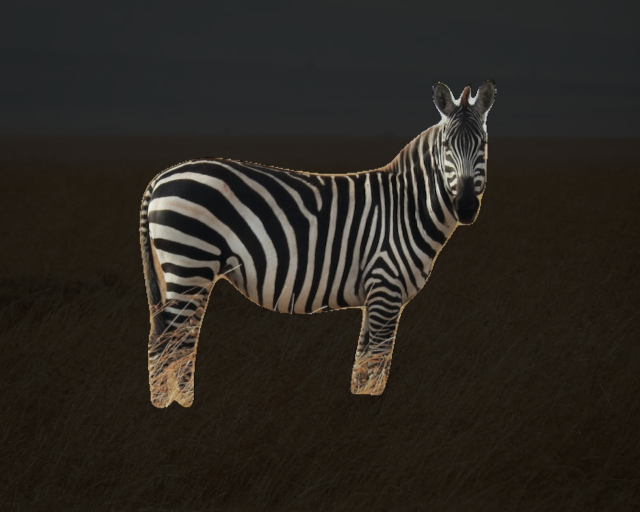} &
            \includegraphics[trim=2 0 0 2,clip,width=0.25\columnwidth]{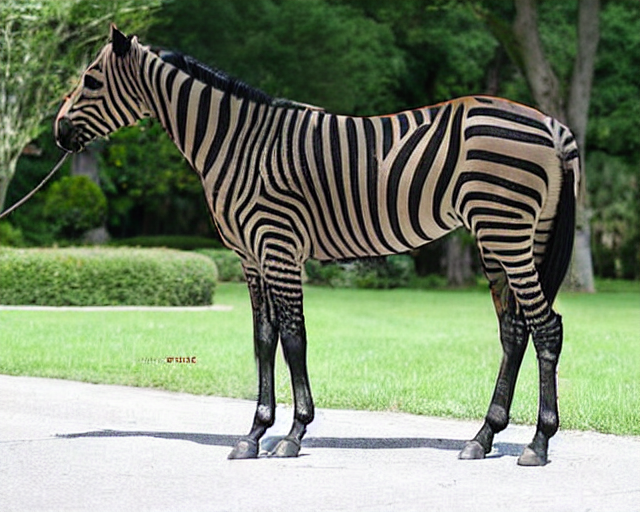} \\

            \includegraphics[trim=2 0 0 2,clip,width=0.25\columnwidth]{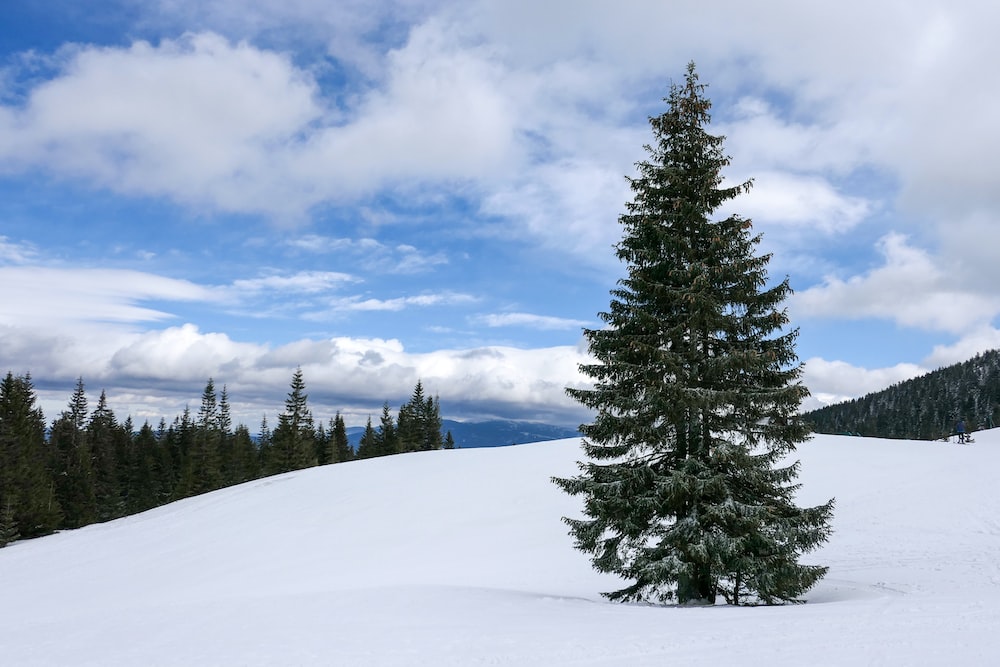} &
            \includegraphics[trim=2 0 0 2,clip,width=0.25\columnwidth]{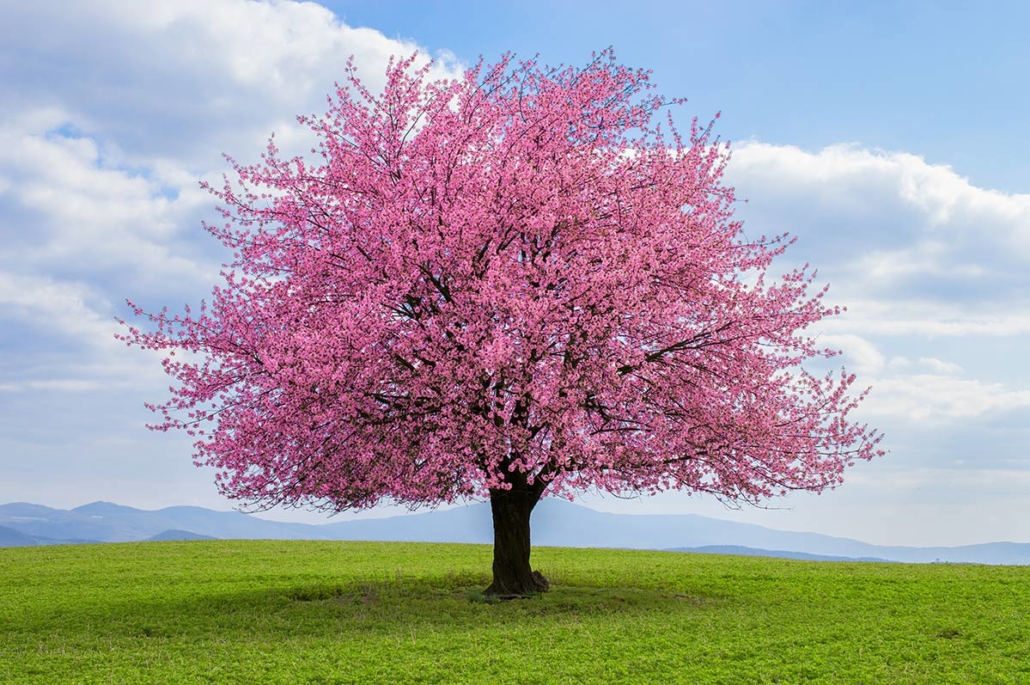} &
            \includegraphics[trim=2 0 0 2,clip,width=0.25\columnwidth]{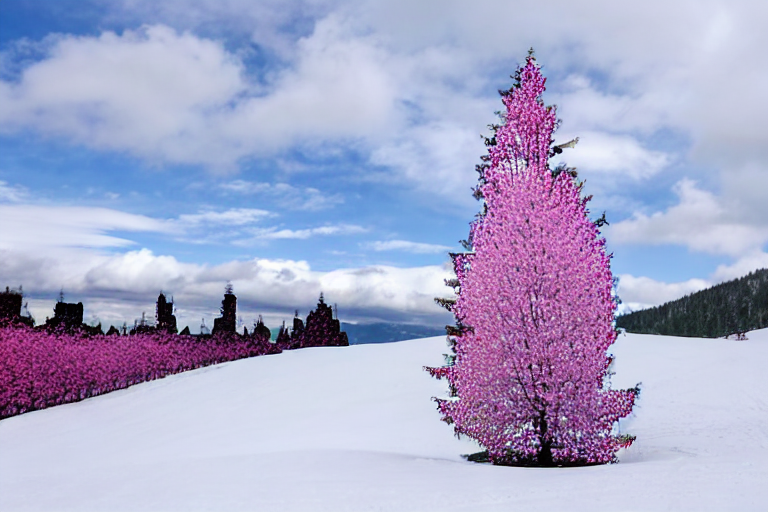} \\

            \includegraphics[trim=2 0 0 2,clip,width=0.25\columnwidth]{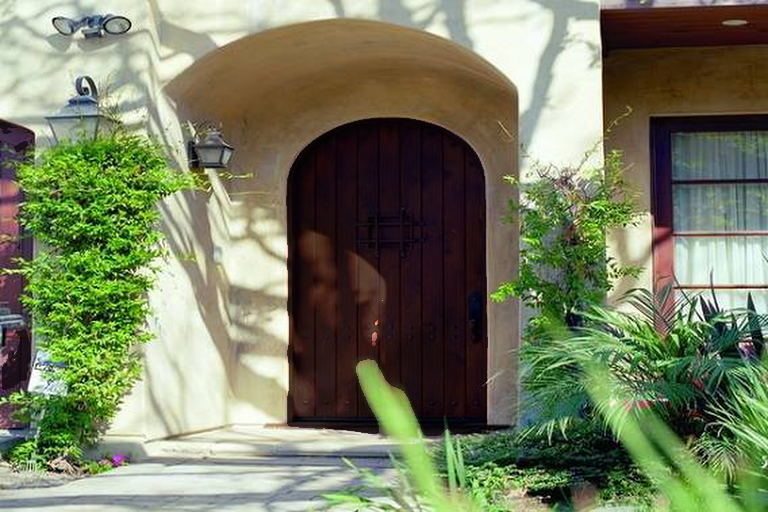} &
            \includegraphics[trim=2 0 0 2,clip,width=0.25\columnwidth]{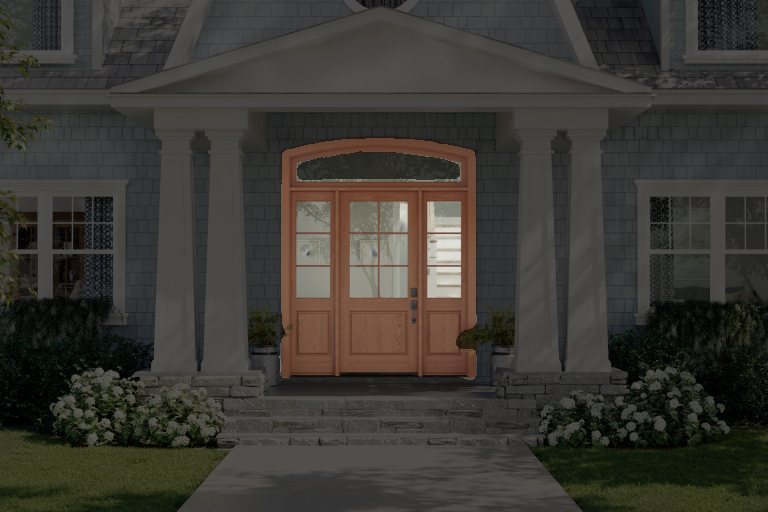} &
            \includegraphics[trim=2 0 0 2,clip,width=0.25\columnwidth]{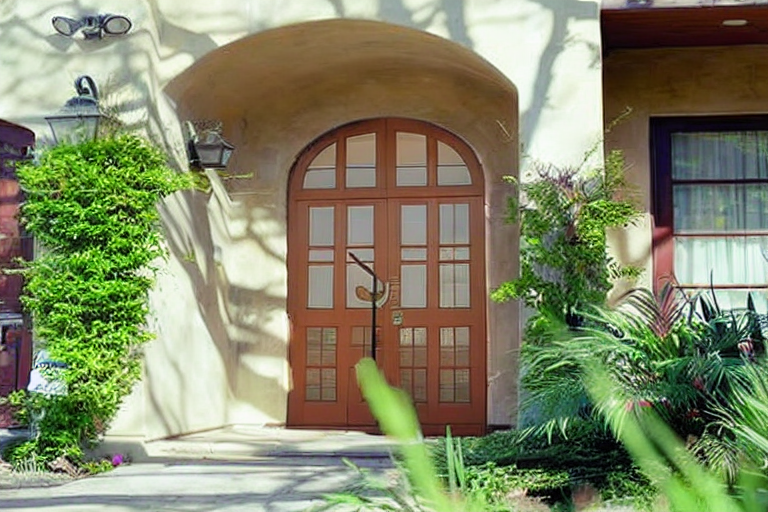} \\

            \includegraphics[trim=2 0 0 2,clip,width=0.25\columnwidth]{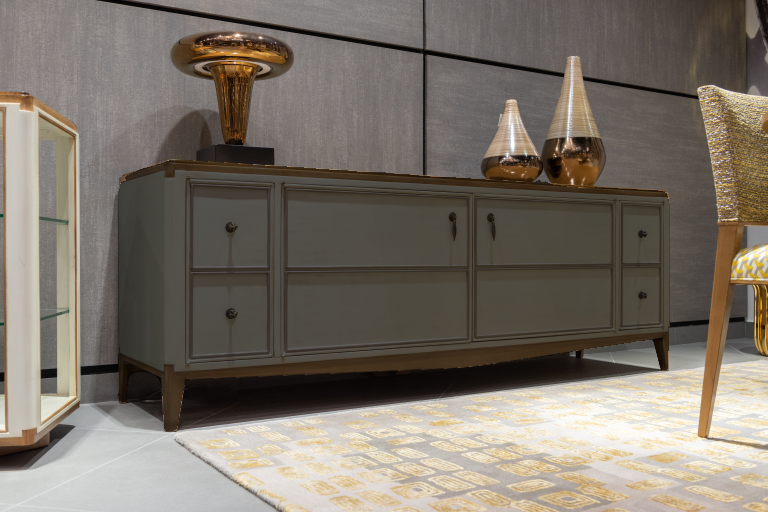} &
            \includegraphics[trim=2 0 0 2,clip,width=0.25\columnwidth]{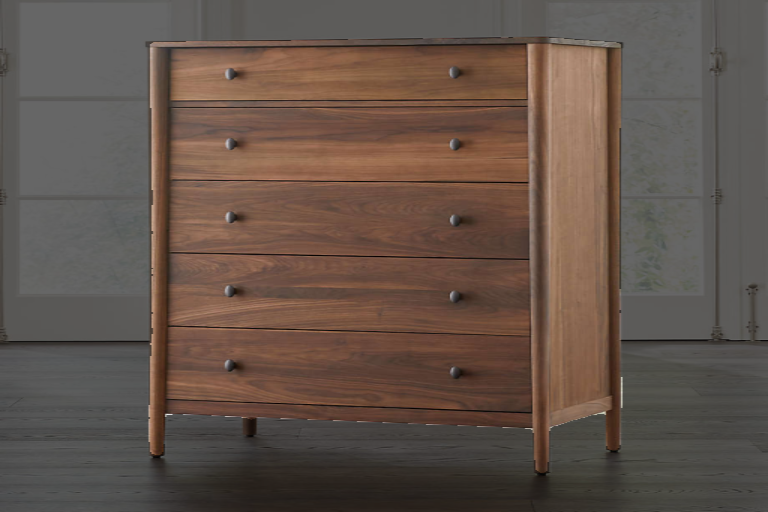} &
            \includegraphics[trim=2 0 0 2,clip,width=0.25\columnwidth]{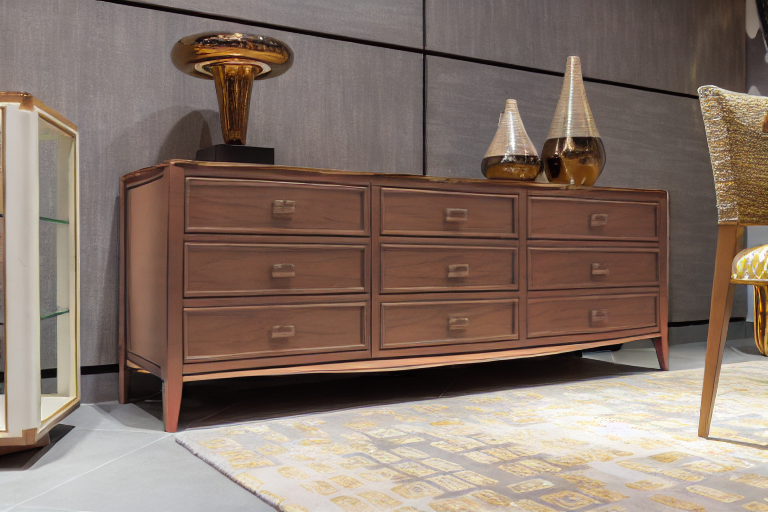} \\

            \includegraphics[trim=2 0 0 2,clip,width=0.25\columnwidth]{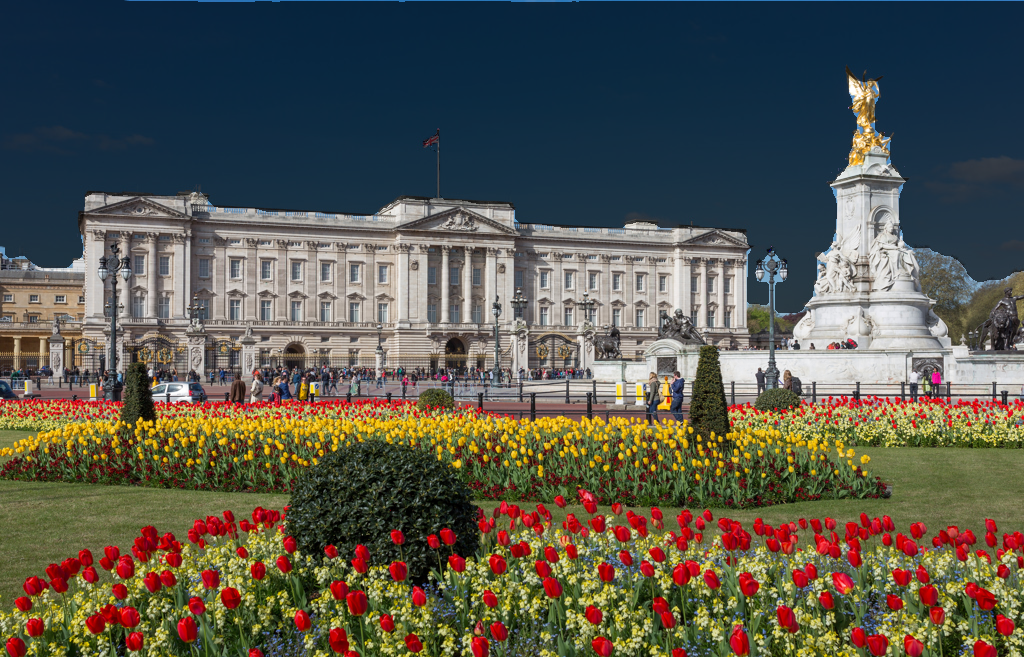} &
            \includegraphics[trim=2 0 0 2,clip,width=0.25\columnwidth]{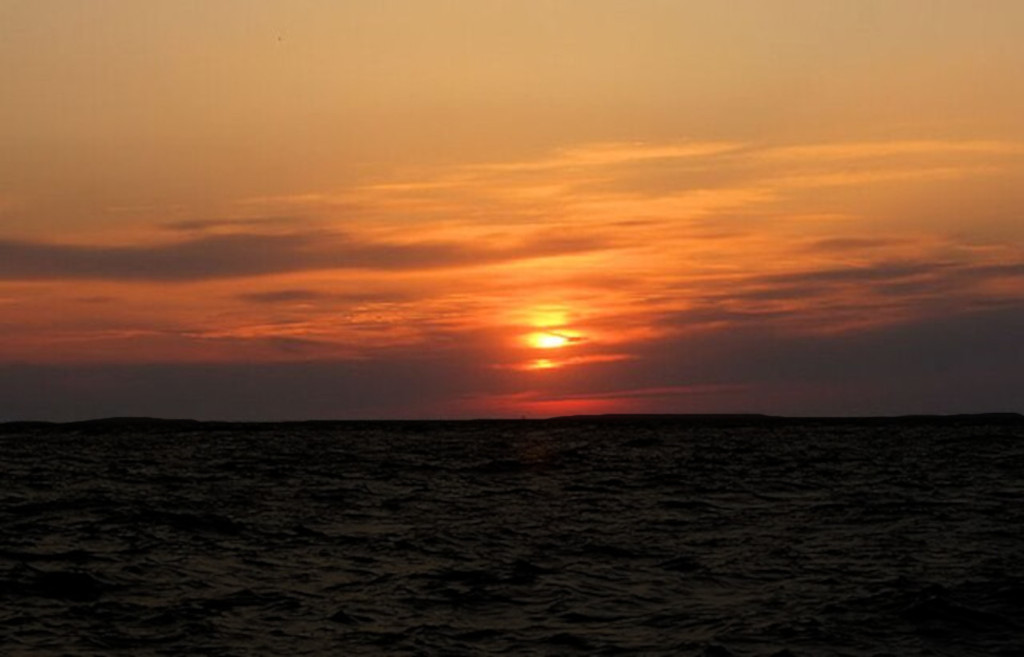} &
            \includegraphics[trim=2 0 0 2,clip,width=0.25\columnwidth]{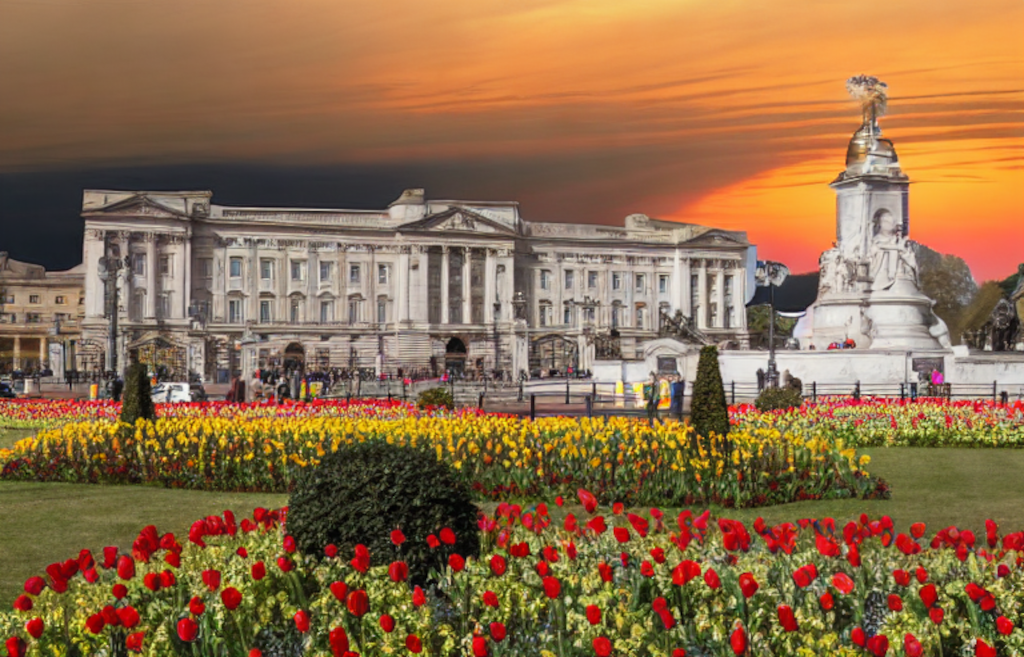} \\
            
    \end{tabular}}
    \captionof{figure}{Appearance editing in the wild.}
    \label{fig:sd_app_edits}
\end{table*}

\begin{table*}[p]
    \centering
    \def\arraystretch{0.5}
    \resizebox{0.95\linewidth}{!}{
    \setlength\tabcolsep{0.5pt}
    \footnotesize
    \begin{tabular}{ccc}
            \scalebox{.5}{Input} &  \scalebox{.5}{Reference Appearance} &  \scalebox{.5}{\textbf{PAIR Diffusion}} \\
            \includegraphics[trim=2 0 0 2,clip,width=0.25\columnwidth]{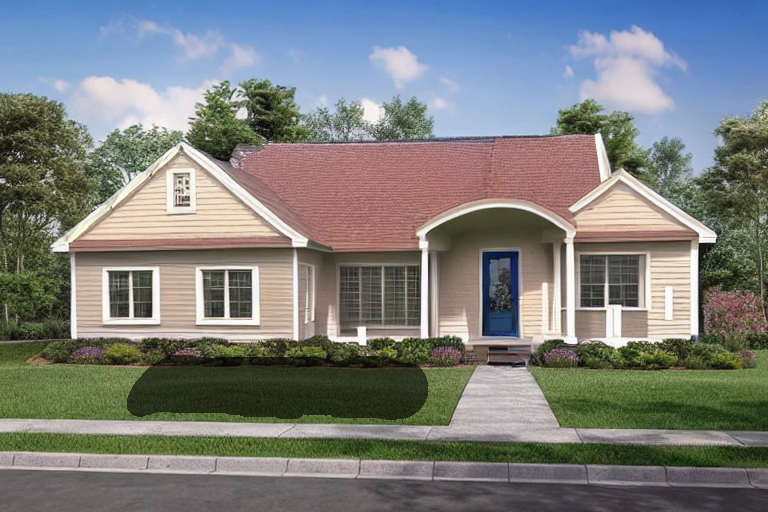} &
            \includegraphics[trim=2 0 0 2,clip,width=0.25\columnwidth]{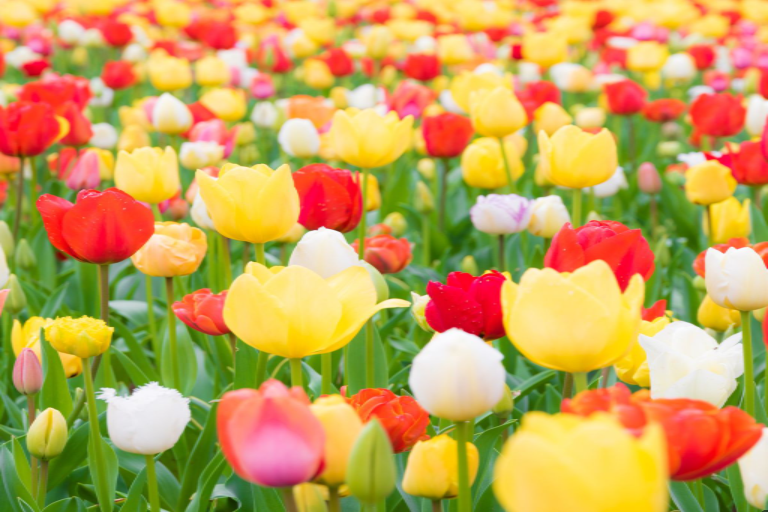} &
            \includegraphics[trim=2 0 0 2,clip,width=0.25\columnwidth]{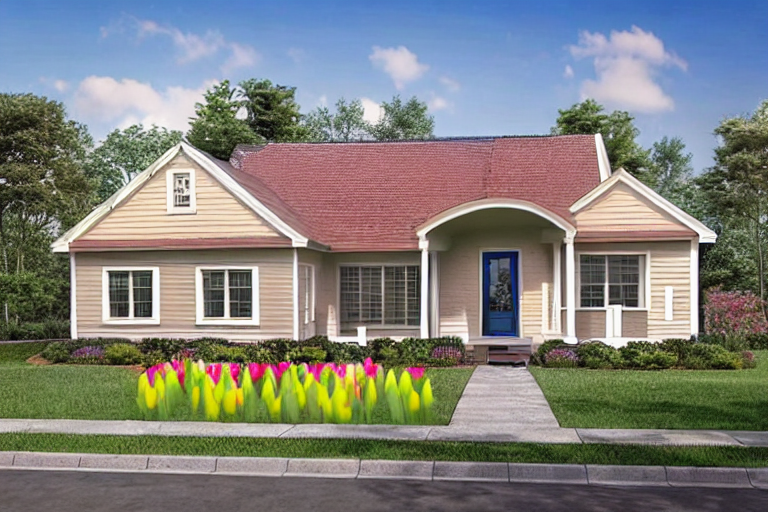}  \\
            
            \includegraphics[trim=2 0 0 2,clip,width=0.25\columnwidth]{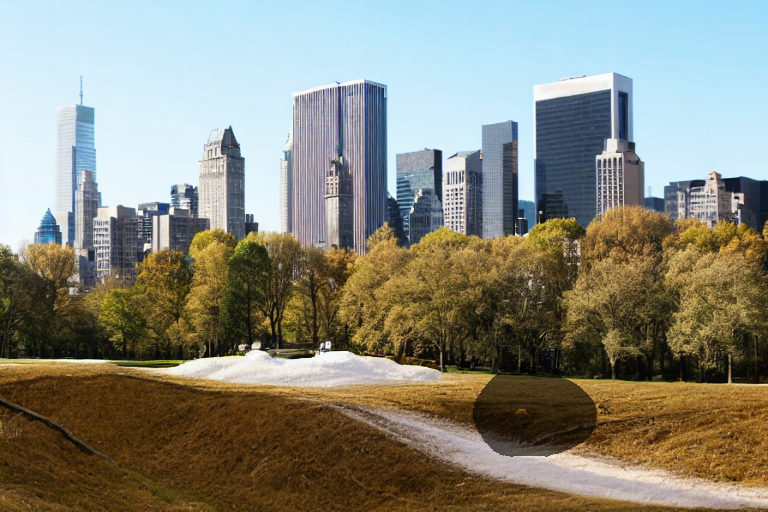} &
            \includegraphics[trim=2 0 0 2,clip,width=0.25\columnwidth]{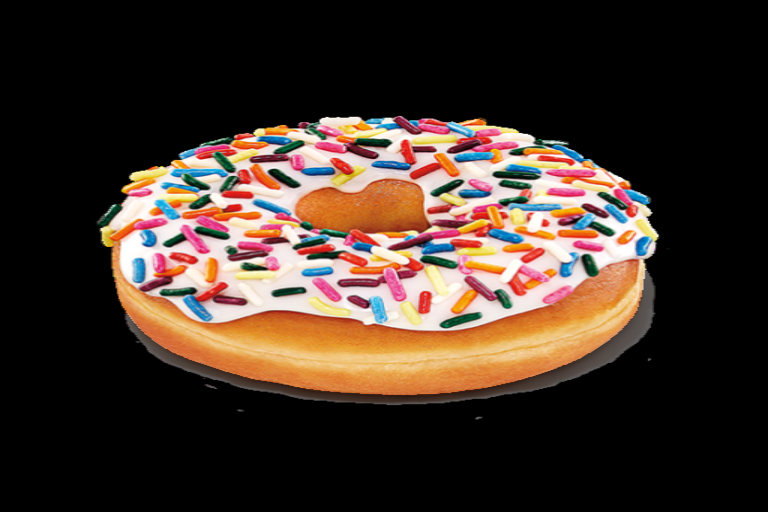} &
            \includegraphics[trim=2 0 0 2,clip,width=0.25\columnwidth]{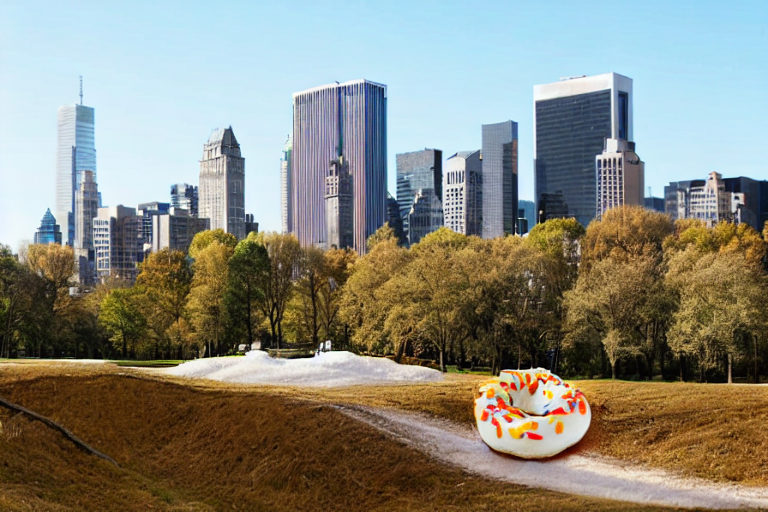}  \\
            
            \includegraphics[trim=2 0 0 2,clip,width=0.25\columnwidth]{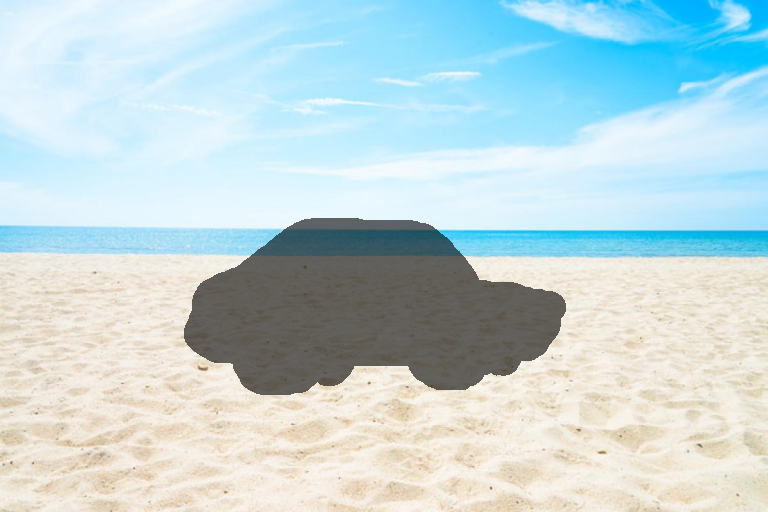} &
            \includegraphics[trim=2 0 0 2,clip,width=0.25\columnwidth]{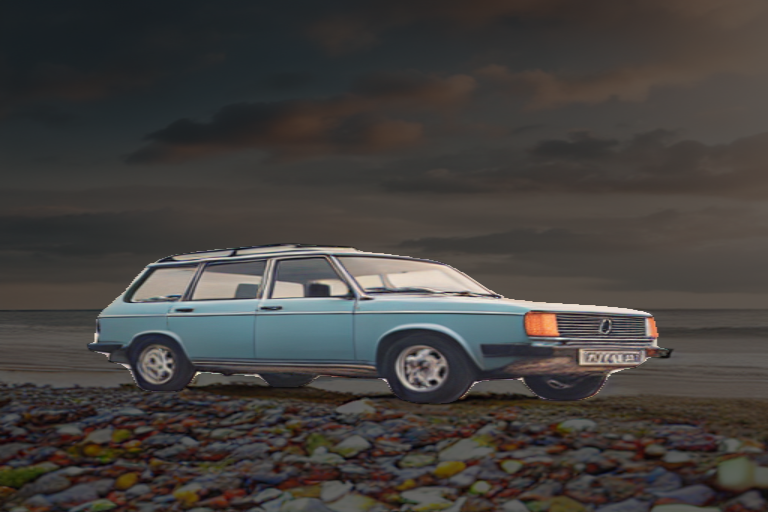} &
            \includegraphics[trim=2 0 0 2,clip,width=0.25\columnwidth]{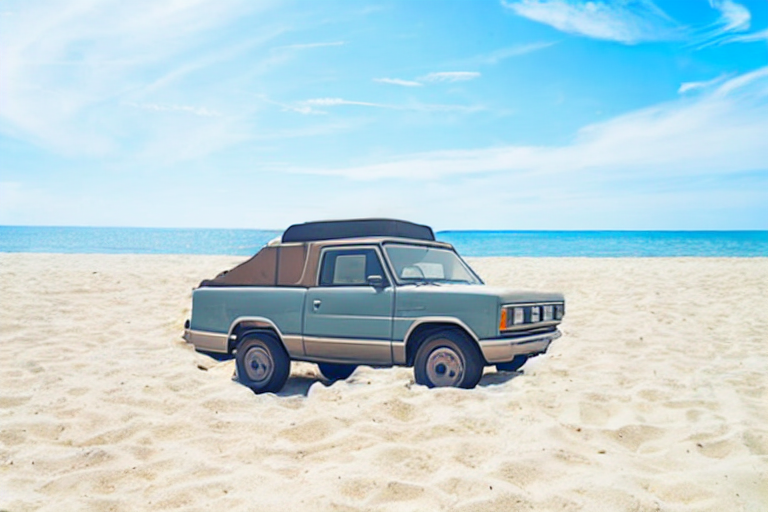}  \\
            
            \includegraphics[trim=2 0 0 2,clip,width=0.25\columnwidth]{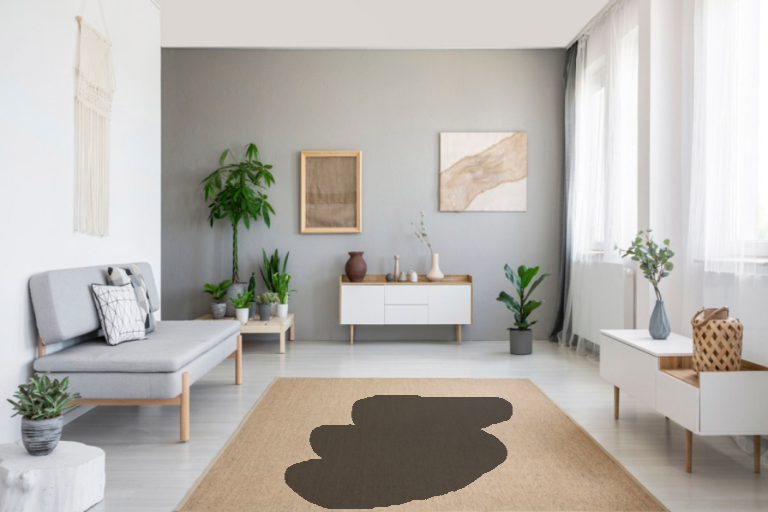} &
            \includegraphics[trim=2 0 0 2,clip,width=0.25\columnwidth]{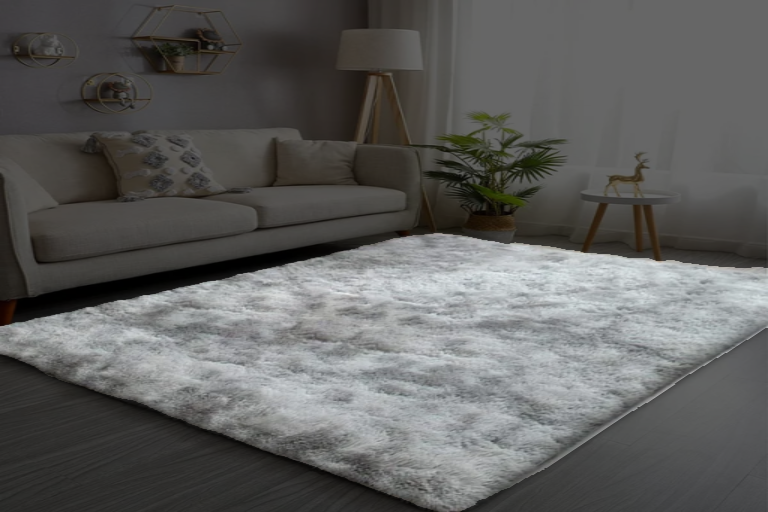} &
            \includegraphics[trim=2 0 0 2,clip,width=0.25\columnwidth]{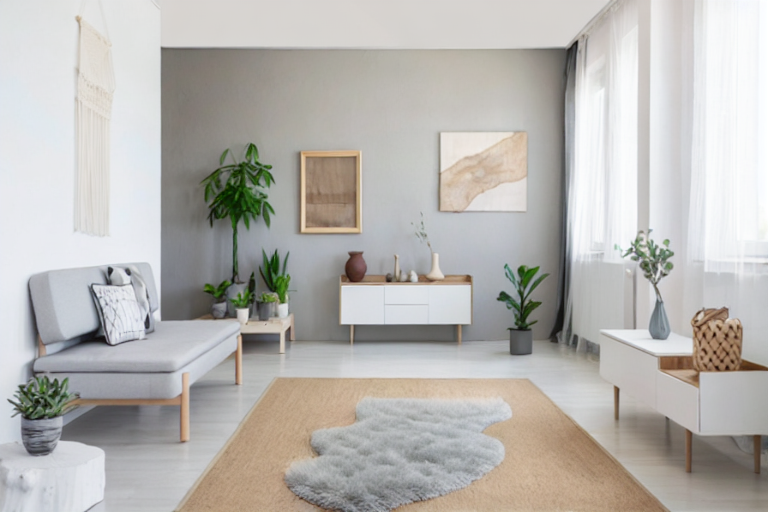}  \\

            \includegraphics[trim=2 0 0 2,clip,width=0.25\columnwidth]{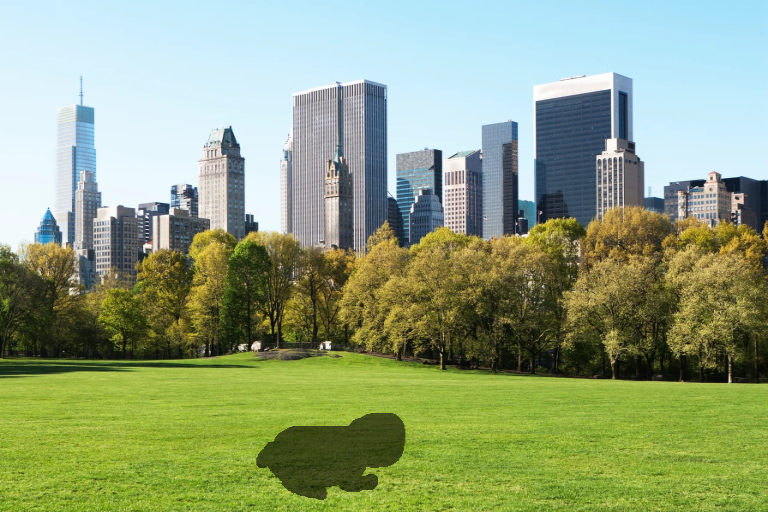} &
            \includegraphics[trim=2 0 0 2,clip,width=0.25\columnwidth]{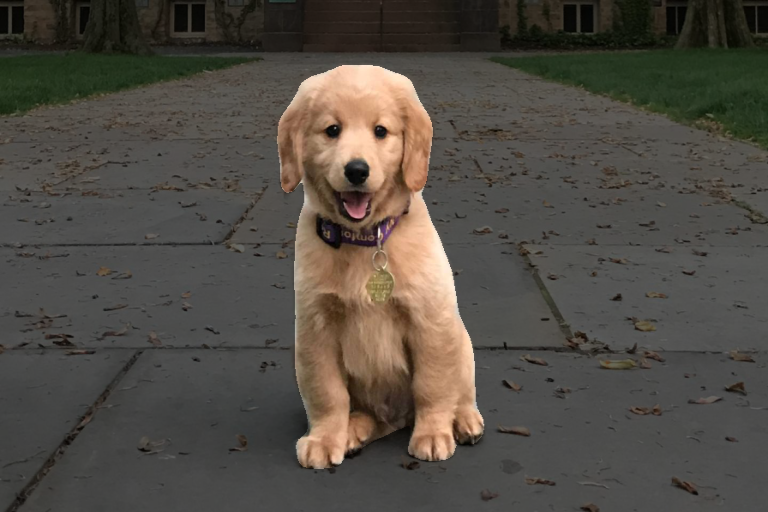} &
            \includegraphics[trim=2 0 0 2,clip,width=0.25\columnwidth]{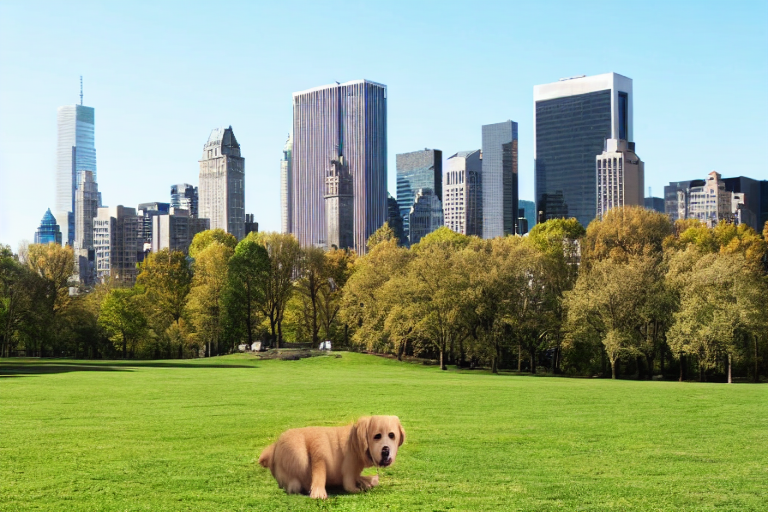}  \\

            \includegraphics[trim=2 0 0 2,clip,width=0.25\columnwidth]{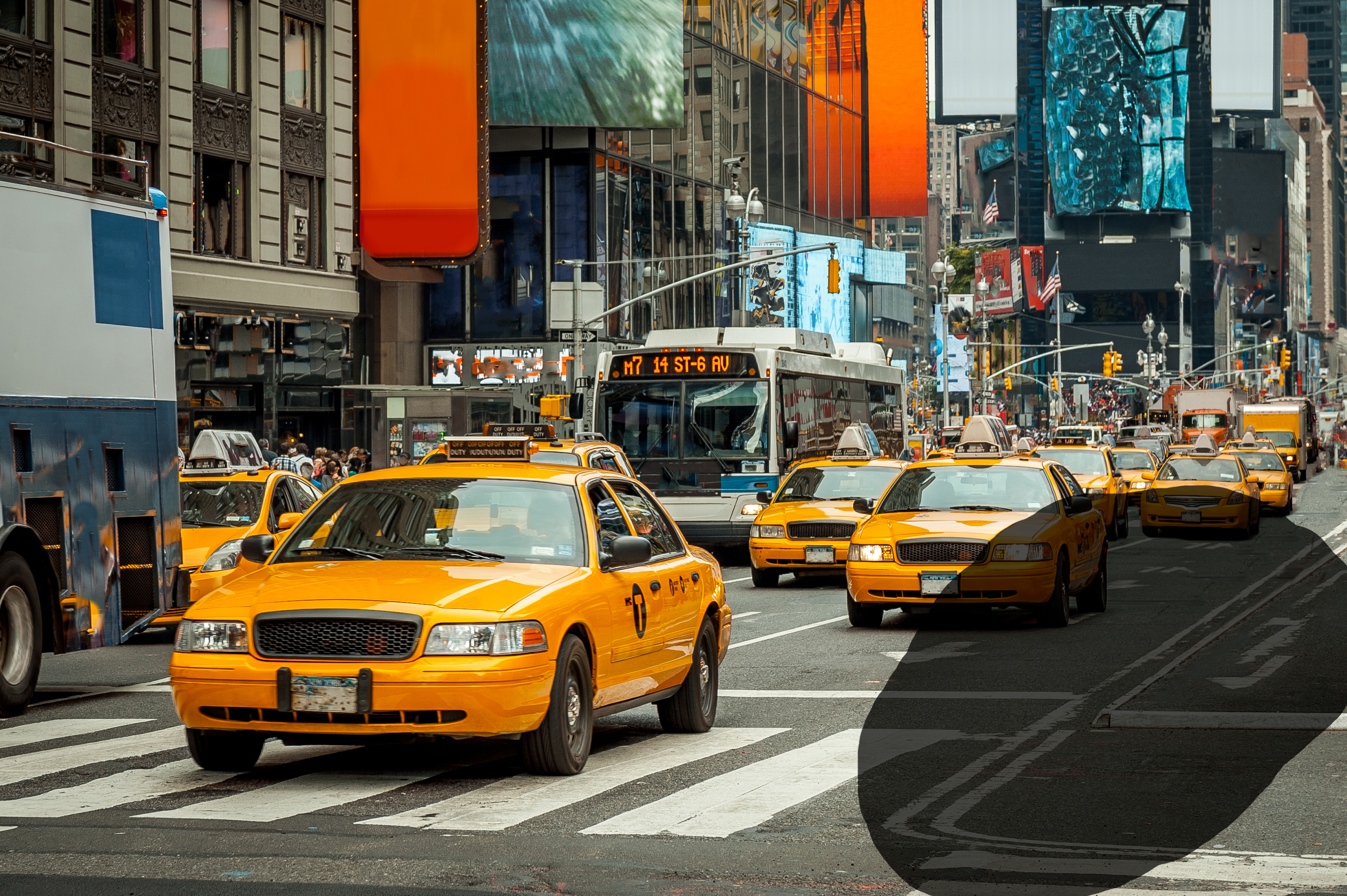} &
            \includegraphics[trim=2 0 0 2,clip,width=0.25\columnwidth]{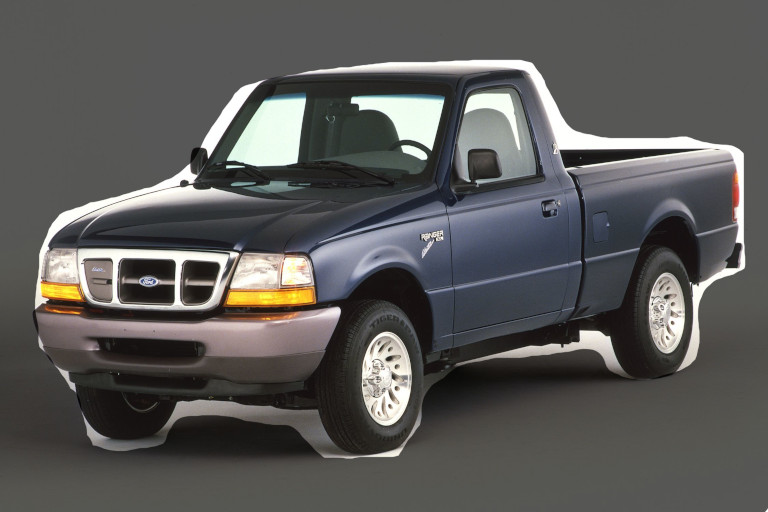} &
            \includegraphics[trim=2 0 0 2,clip,width=0.25\columnwidth]{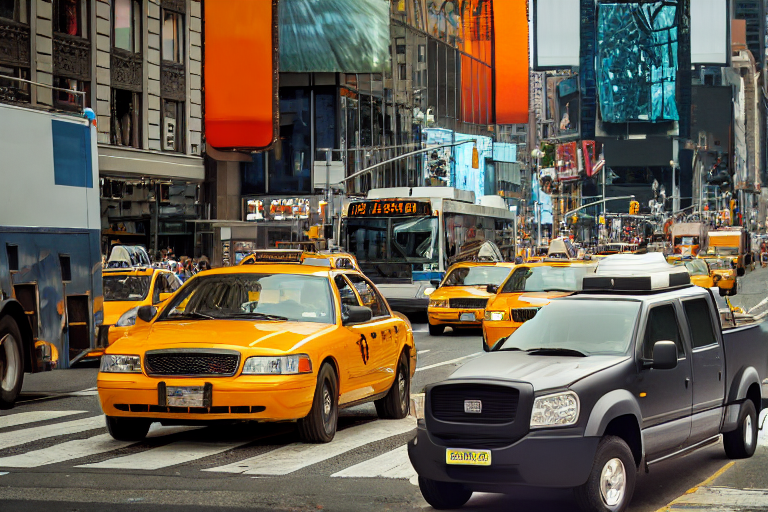}  \\
    \end{tabular}}
    \captionof{figure}{Add objects to the scene.}
    \label{fig:sd_add_objects}
\end{table*}

\begin{table*}[t]
    \centering
    \def\arraystretch{0.5}
    \resizebox{\linewidth}{!}{
    \setlength\tabcolsep{0.5pt}
    \footnotesize
    \begin{tabular}{ccccc}
            \scalebox{0.6}{Input} &  \multicolumn{4}{c}{\scalebox{0.6}{Variations}} \\
            \includegraphics[trim=2 0 0 2,clip,width=0.15\columnwidth]{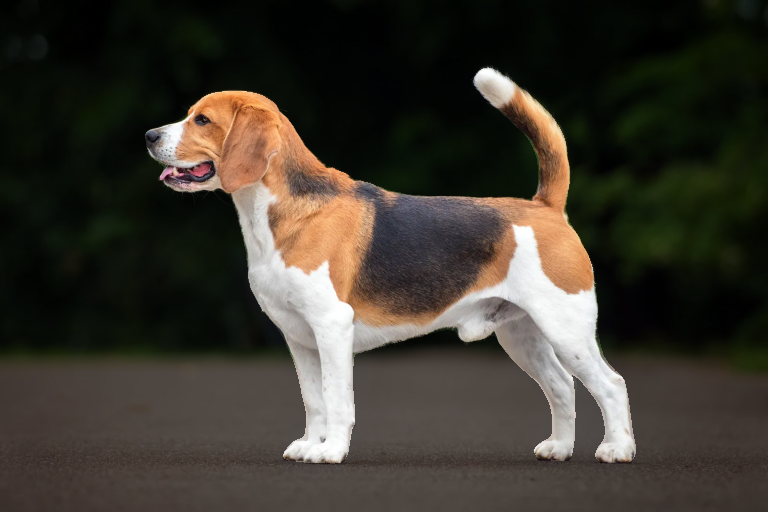} &
            \includegraphics[trim=2 0 0 2,clip,width=0.15\columnwidth]{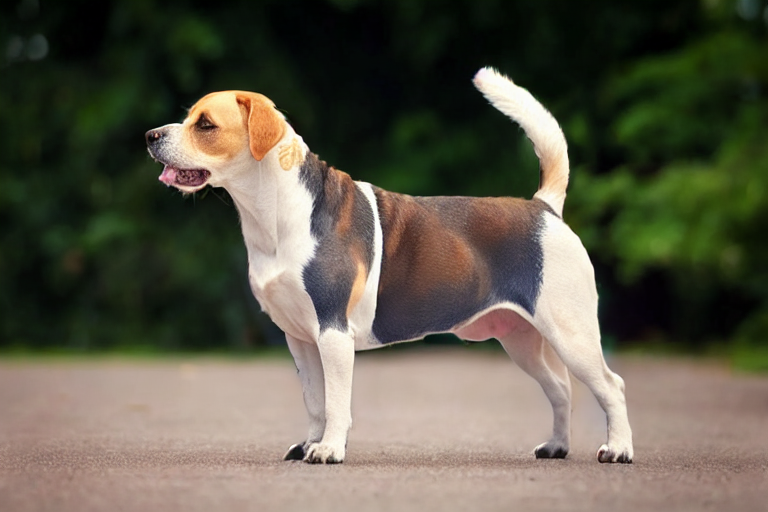} &
            \includegraphics[trim=2 0 0 2,clip,width=0.15\columnwidth]{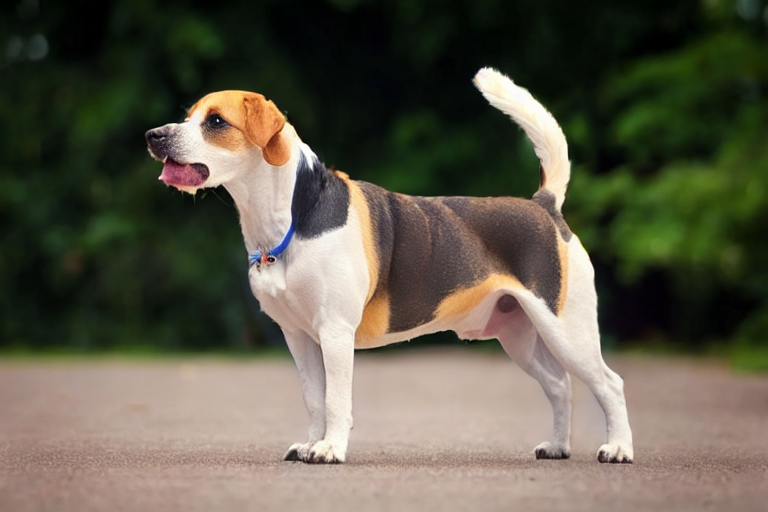} &
            \includegraphics[trim=2 0 0 2,clip,width=0.15\columnwidth]{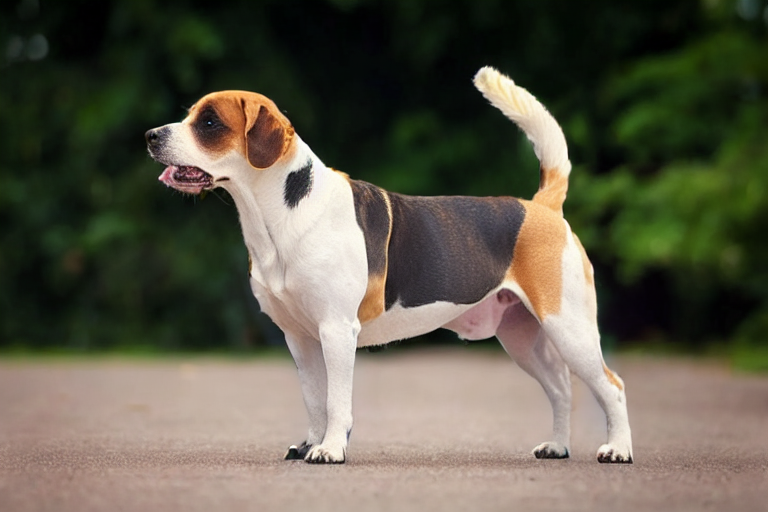} &
            \includegraphics[trim=2 0 0 2,clip,width=0.15\columnwidth]{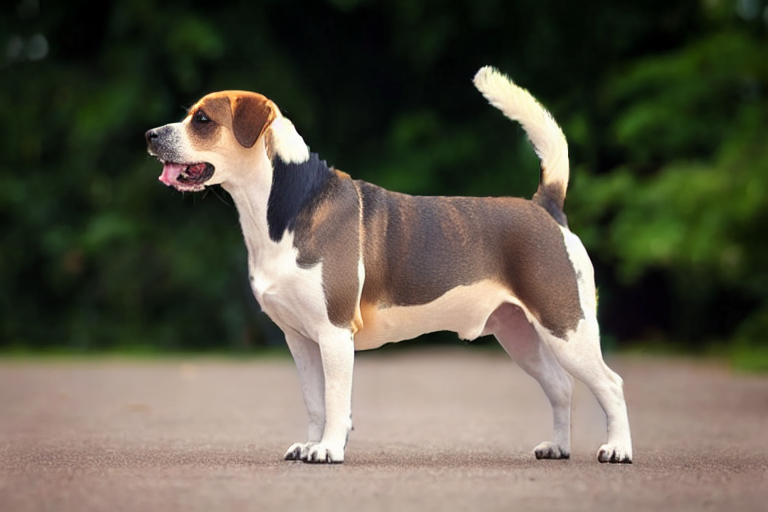} \\

            \includegraphics[trim=2 0 0 2,clip,width=0.15\columnwidth]{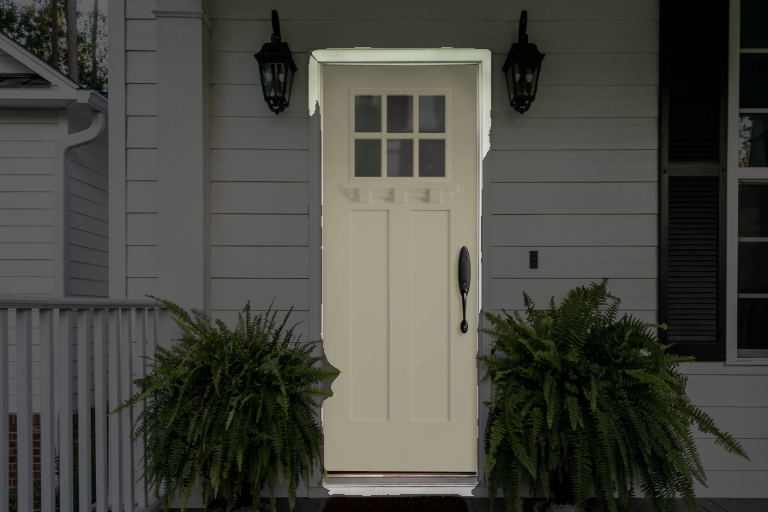} &
            \includegraphics[trim=2 0 0 2,clip,width=0.15\columnwidth]{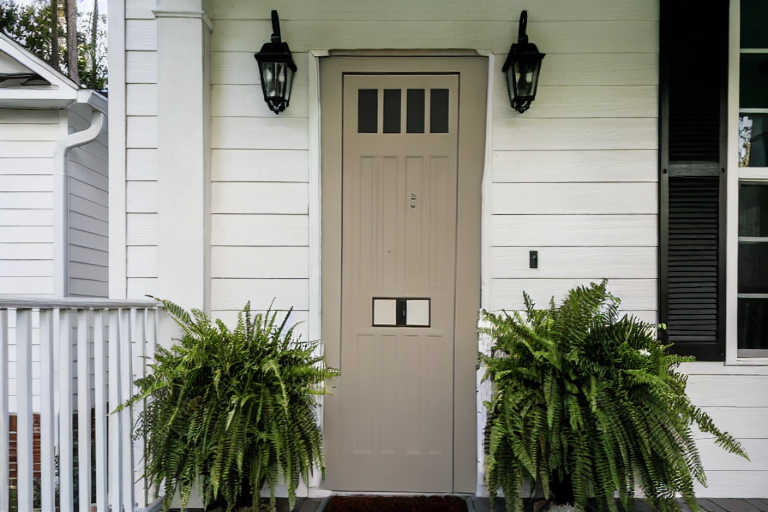} &
            \includegraphics[trim=2 0 0 2,clip,width=0.15\columnwidth]{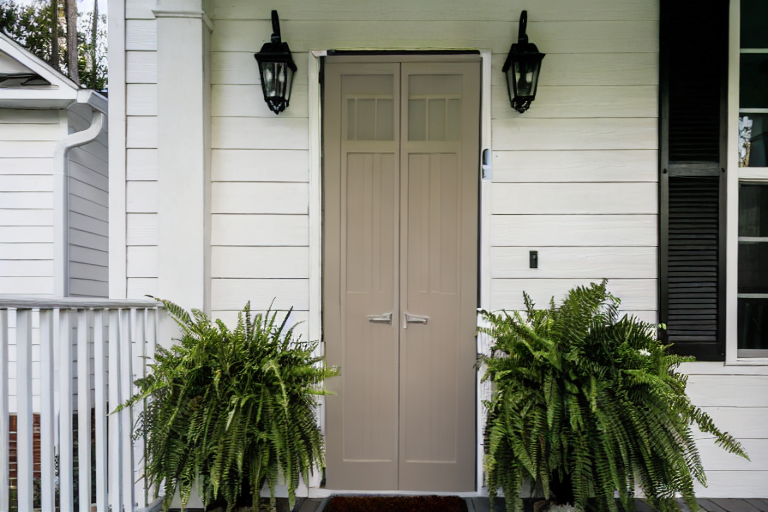} &
            \includegraphics[trim=2 0 0 2,clip,width=0.15\columnwidth]{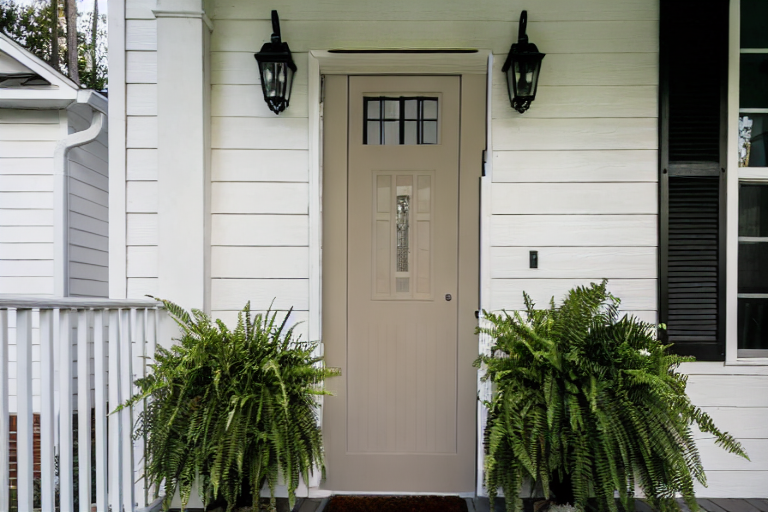} &
            \includegraphics[trim=2 0 0 2,clip,width=0.15\columnwidth]{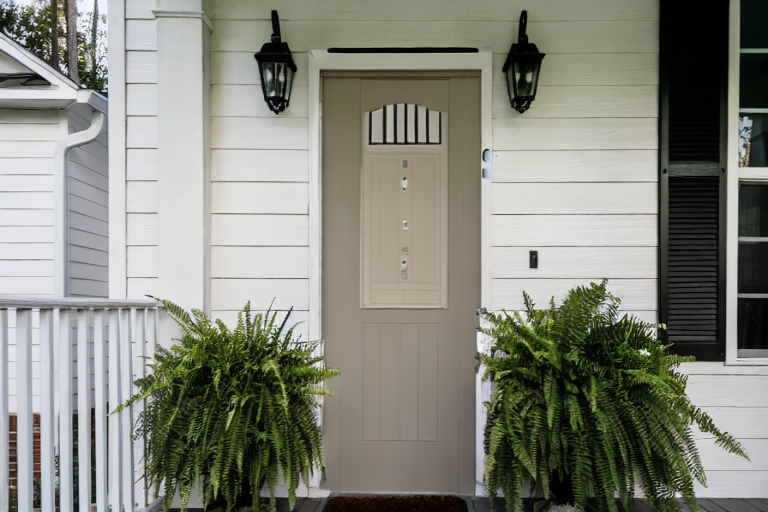} \\

            \includegraphics[trim=2 0 0 2,clip,width=0.15\columnwidth]{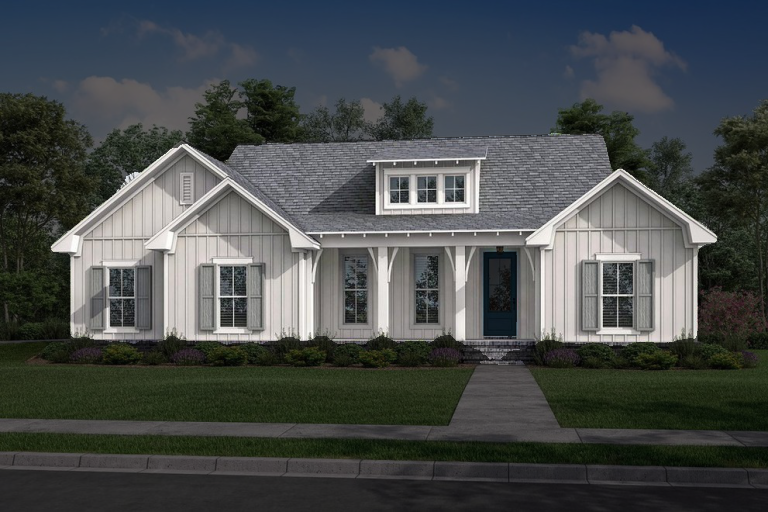} &
            \includegraphics[trim=2 0 0 2,clip,width=0.15\columnwidth]{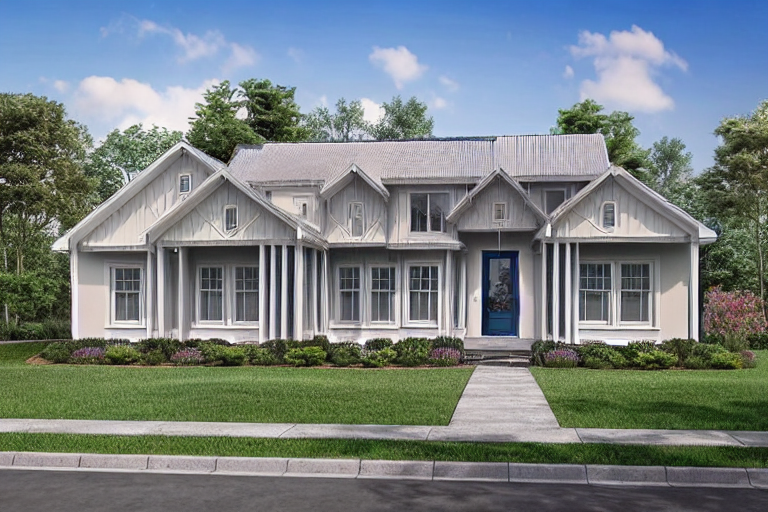} &
            \includegraphics[trim=2 0 0 2,clip,width=0.15\columnwidth]{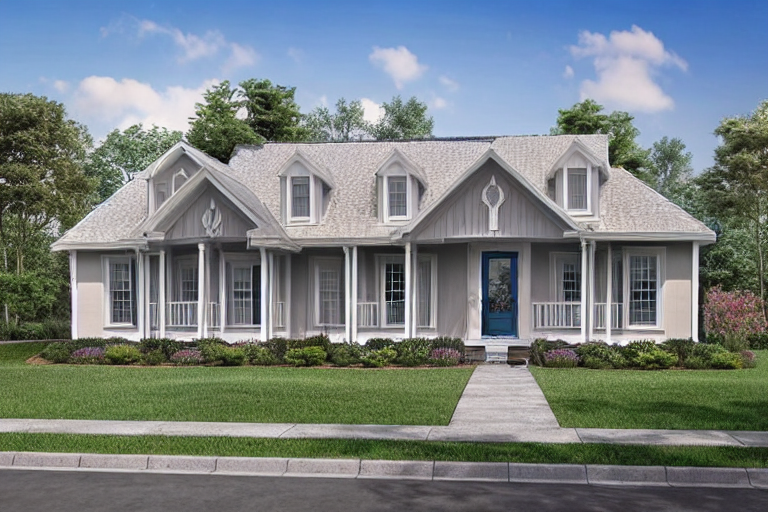} &
            \includegraphics[trim=2 0 0 2,clip,width=0.15\columnwidth]{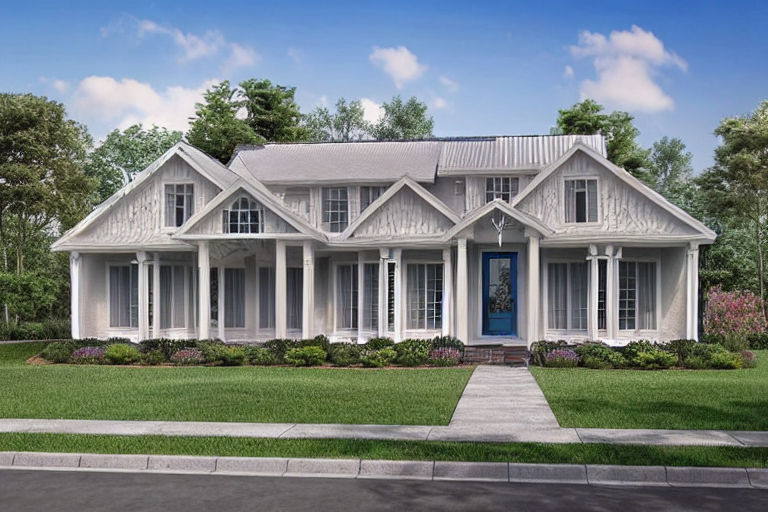} &
            \includegraphics[trim=2 0 0 2,clip,width=0.15\columnwidth]{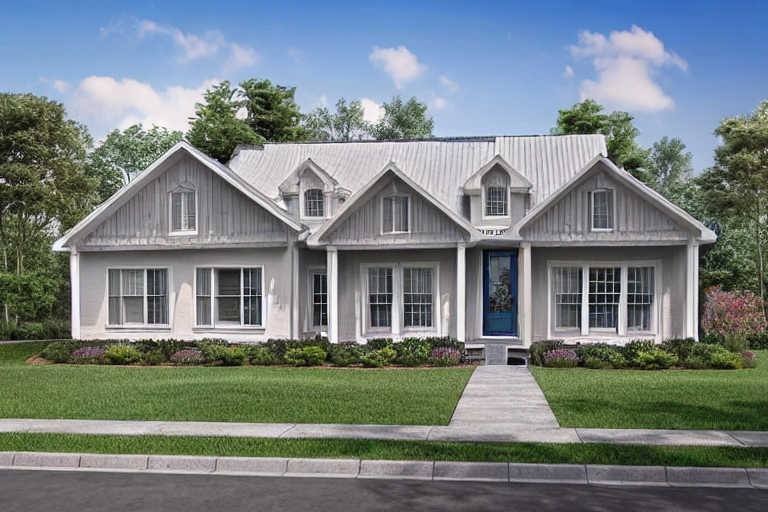} \\

            \includegraphics[trim=2 0 0 2,clip,width=0.15\columnwidth]{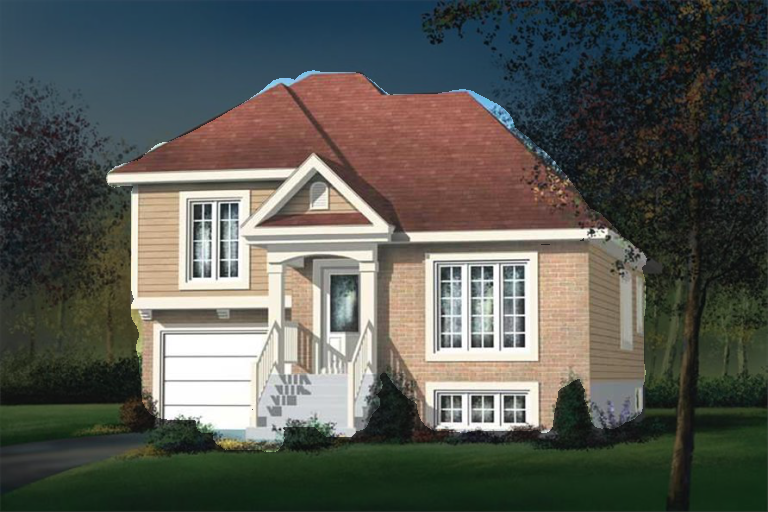} &
            \includegraphics[trim=2 0 0 2,clip,width=0.15\columnwidth]{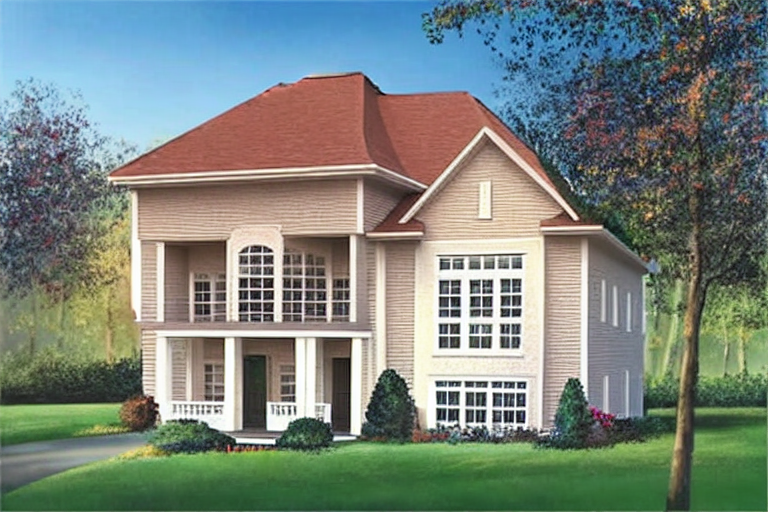} &
            \includegraphics[trim=2 0 0 2,clip,width=0.15\columnwidth]{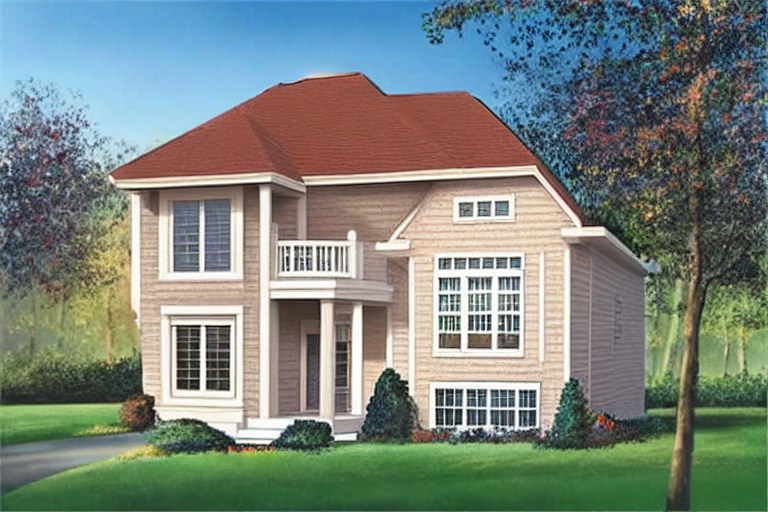} &
            \includegraphics[trim=2 0 0 2,clip,width=0.15\columnwidth]{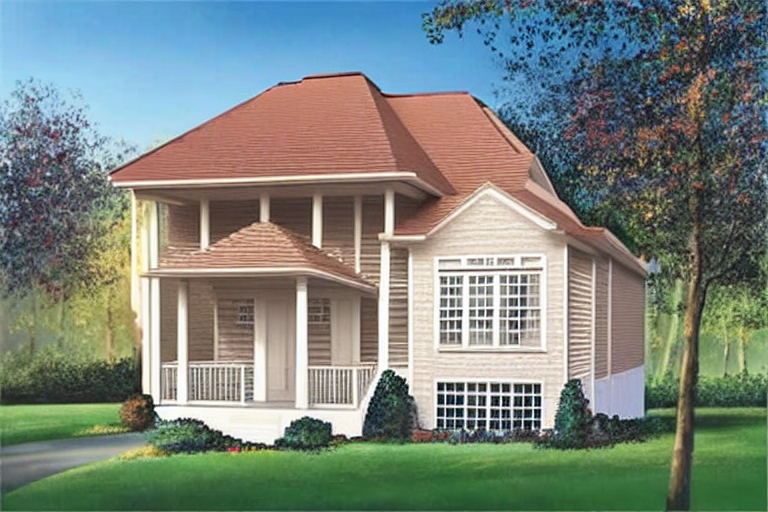} &
            \includegraphics[trim=2 0 0 2,clip,width=0.15\columnwidth]{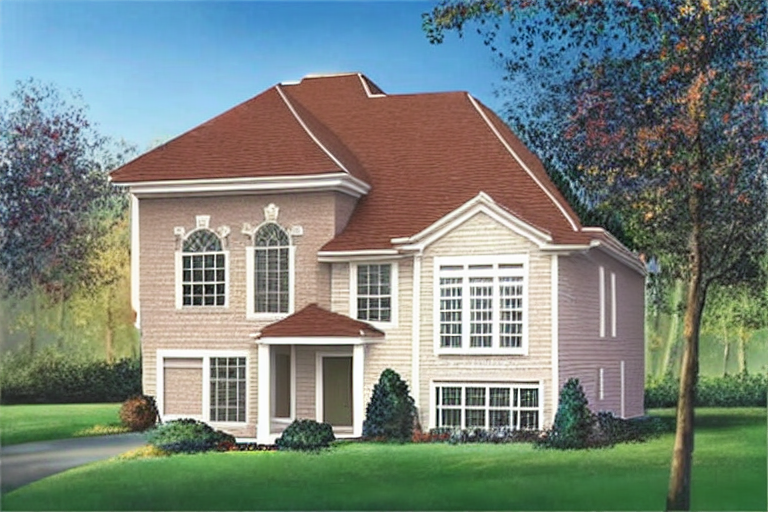} \\

            \includegraphics[trim=2 0 0 2,clip,width=0.15\columnwidth]{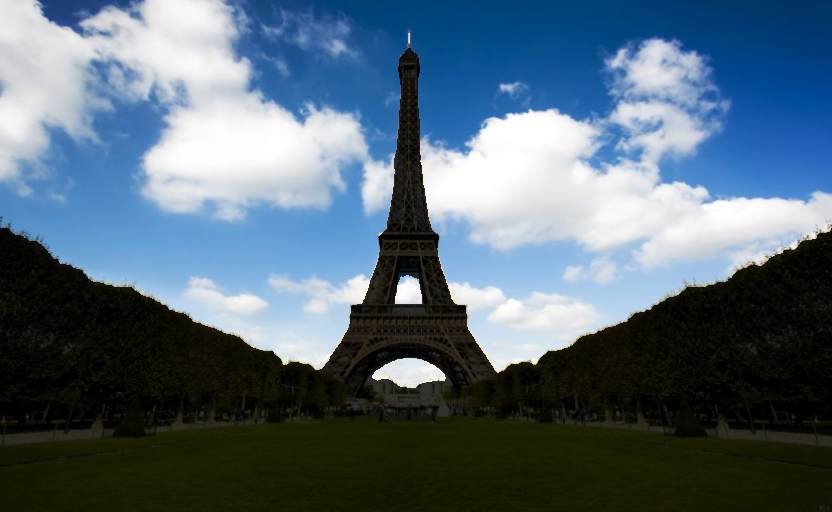} &
            \includegraphics[trim=2 0 0 2,clip,width=0.15\columnwidth]{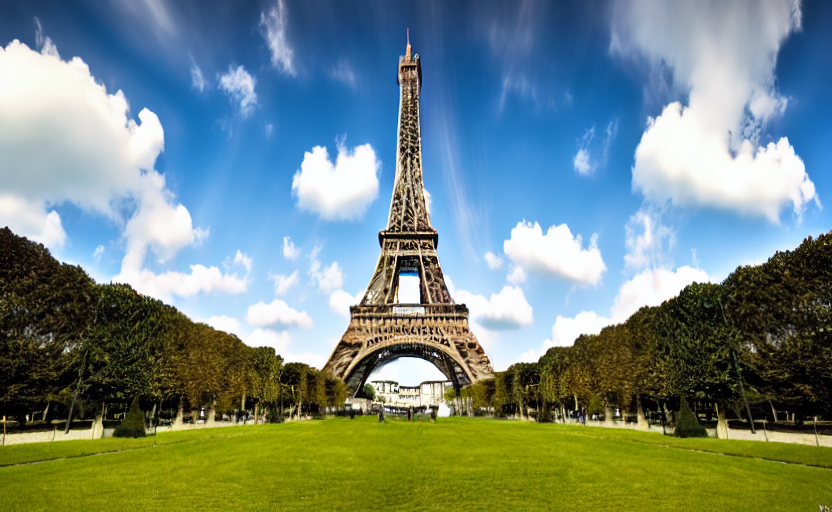} &
            \includegraphics[trim=2 0 0 2,clip,width=0.15\columnwidth]{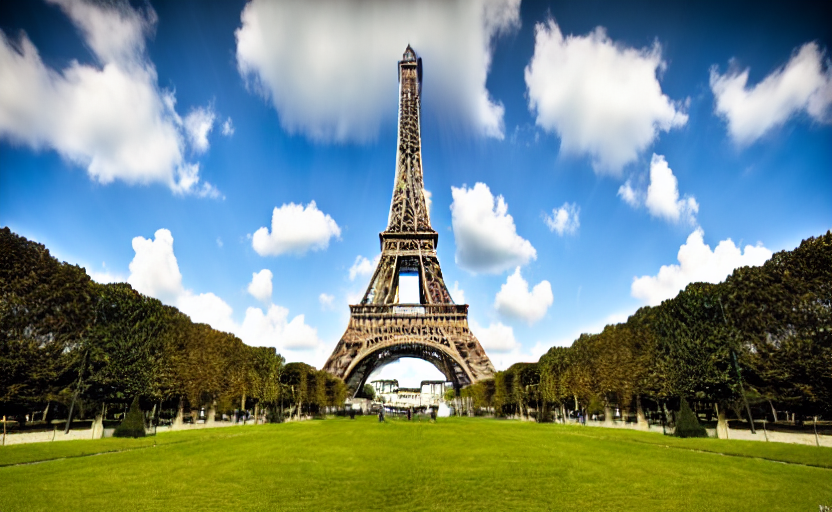} &
            \includegraphics[trim=2 0 0 2,clip,width=0.15\columnwidth]{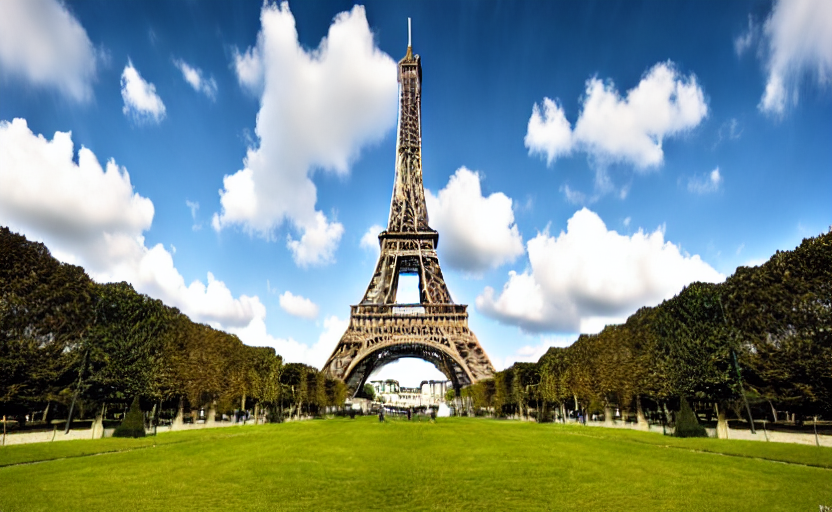} &
            \includegraphics[trim=2 0 0 2,clip,width=0.15\columnwidth]{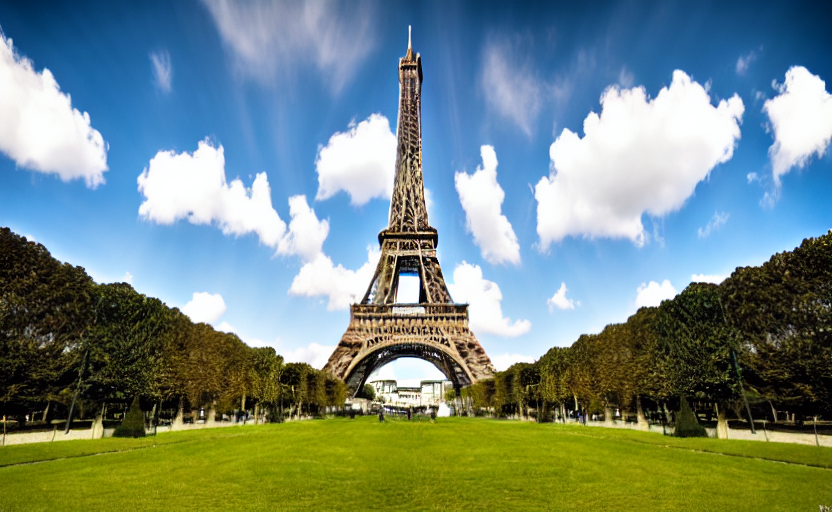} \\

            \includegraphics[trim=2 0 0 2,clip,width=0.15\columnwidth]{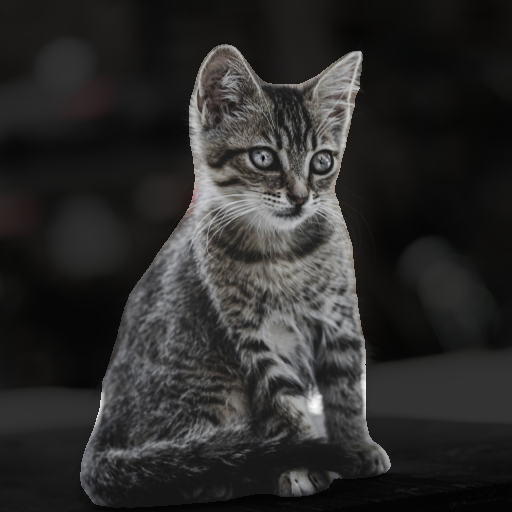} &
            \includegraphics[trim=2 0 0 2,clip,width=0.15\columnwidth]{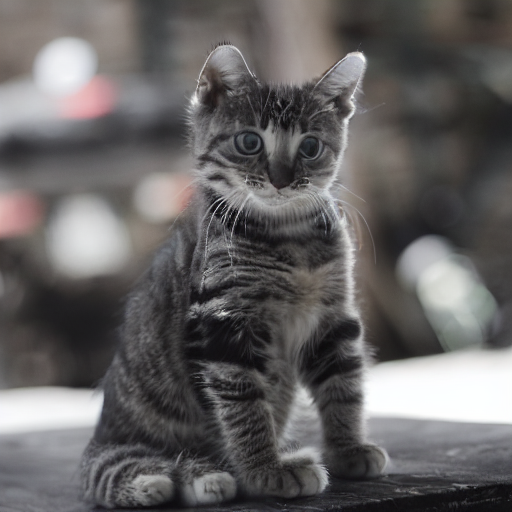} &
            \includegraphics[trim=2 0 0 2,clip,width=0.15\columnwidth]{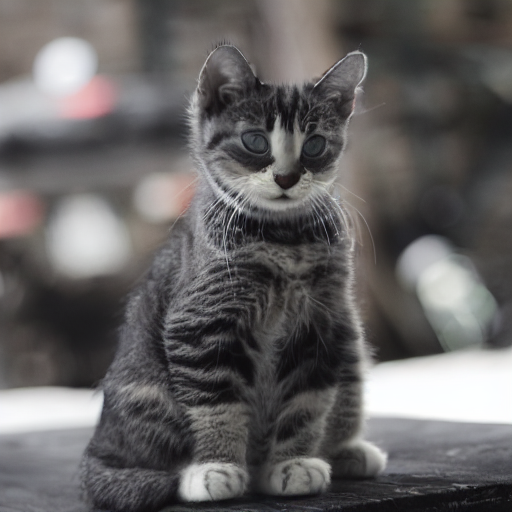} &
            \includegraphics[trim=2 0 0 2,clip,width=0.15\columnwidth]{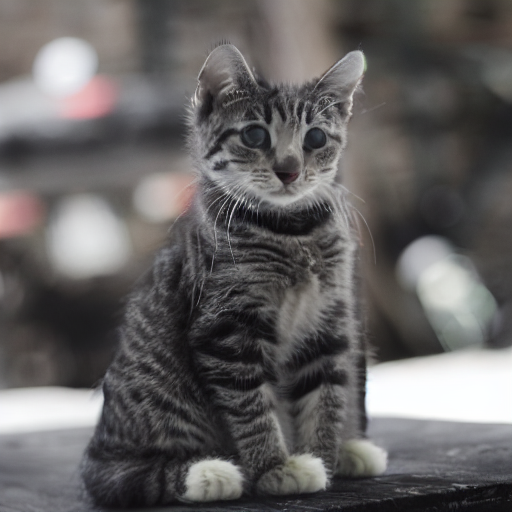} &
            \includegraphics[trim=2 0 0 2,clip,width=0.15\columnwidth]{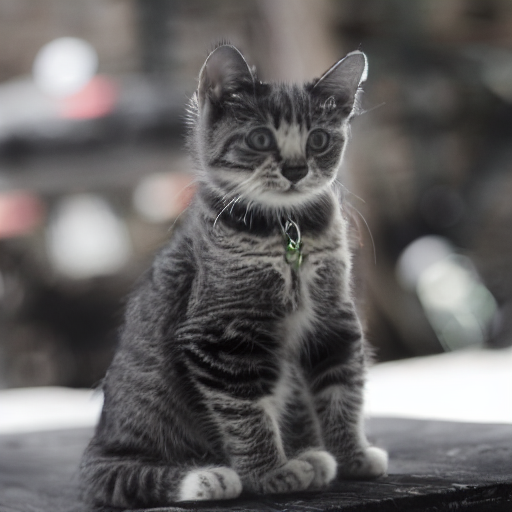} \\

            \includegraphics[trim=2 0 0 2,clip,width=0.15\columnwidth]{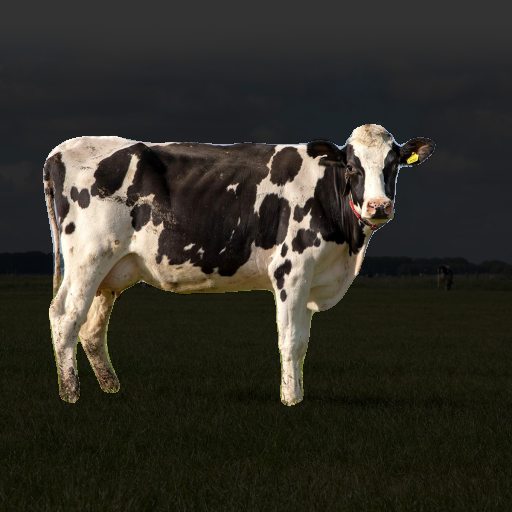} &
            \includegraphics[trim=2 0 0 2,clip,width=0.15\columnwidth]{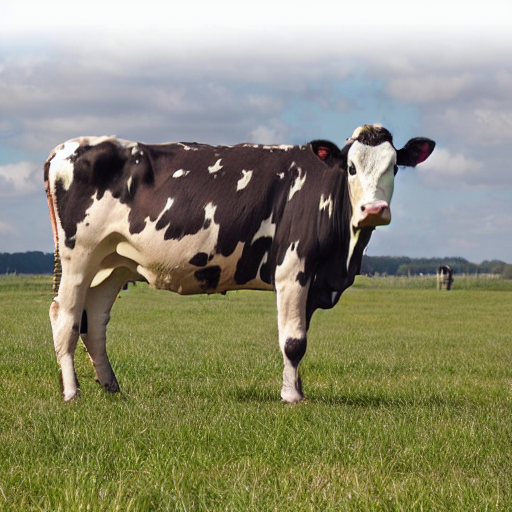} &
            \includegraphics[trim=2 0 0 2,clip,width=0.15\columnwidth]{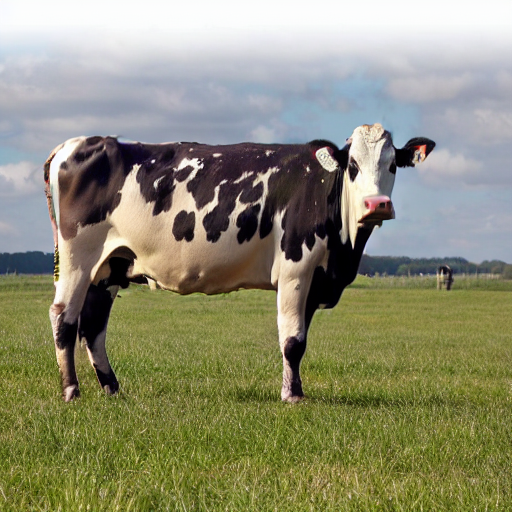} &
            \includegraphics[trim=2 0 0 2,clip,width=0.15\columnwidth]{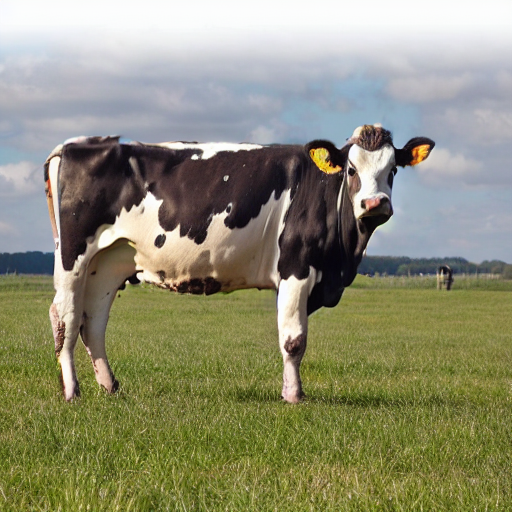} &
            \includegraphics[trim=2 0 0 2,clip,width=0.15\columnwidth]{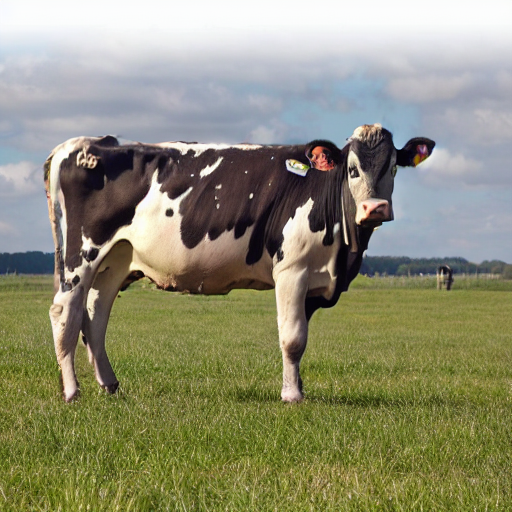} \\

            \includegraphics[trim=2 0 0 2,clip,width=0.15\columnwidth]{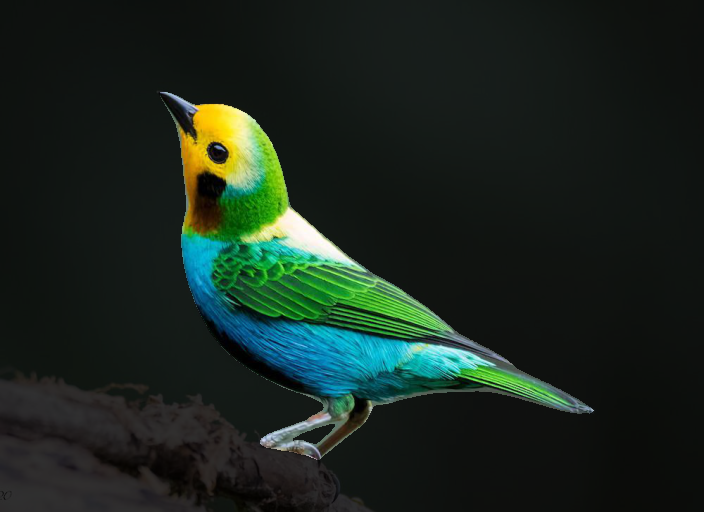} &
            \includegraphics[trim=2 0 0 2,clip,width=0.15\columnwidth]{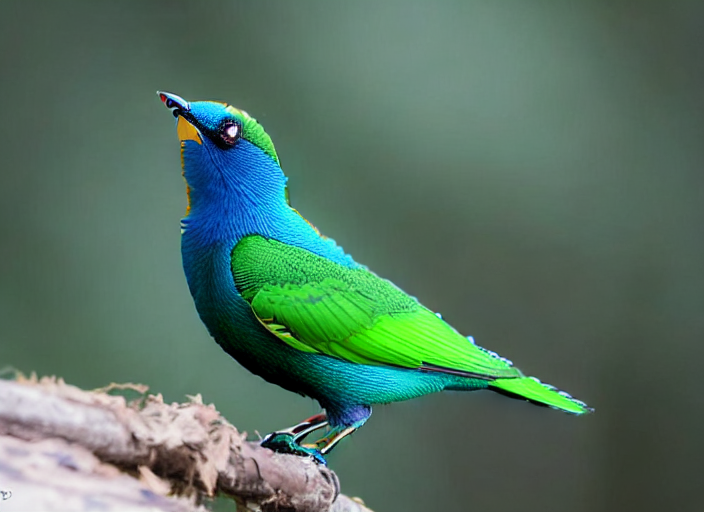} &
            \includegraphics[trim=2 0 0 2,clip,width=0.15\columnwidth]{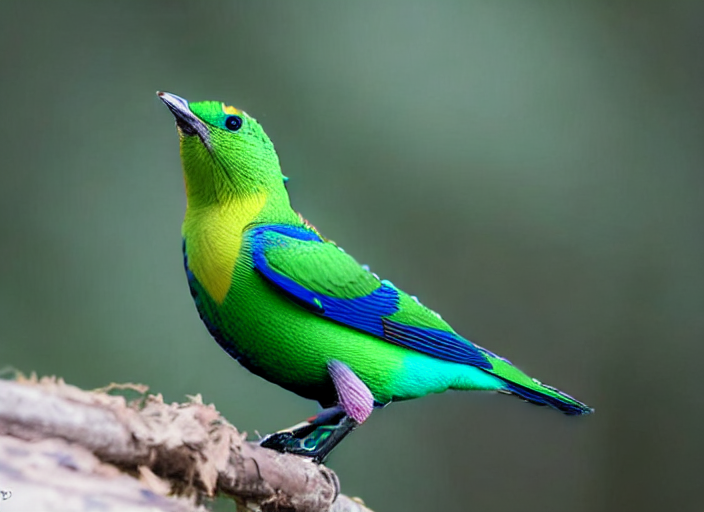} &
            \includegraphics[trim=2 0 0 2,clip,width=0.15\columnwidth]{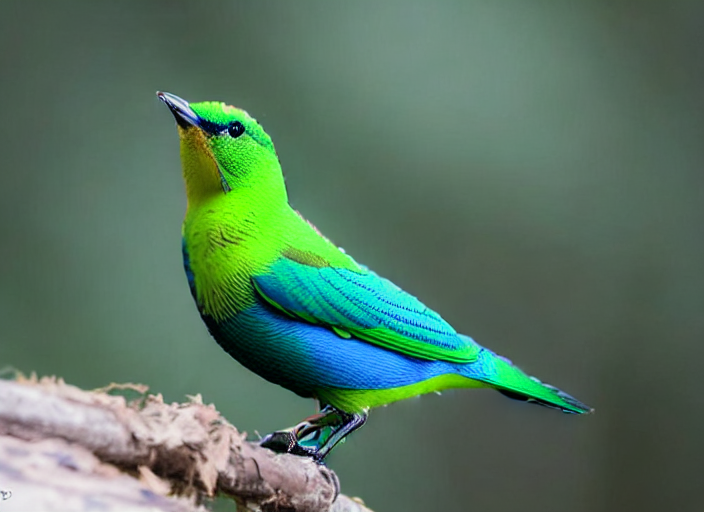} &
            \includegraphics[trim=2 0 0 2,clip,width=0.15\columnwidth]{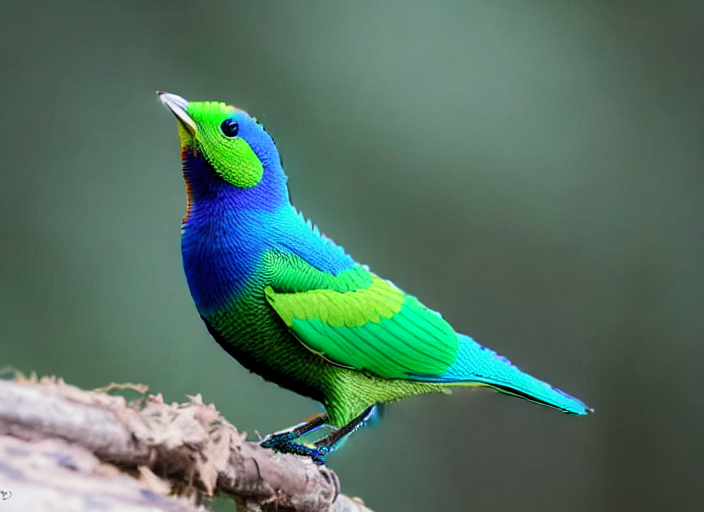} \\
            
    \end{tabular}}
    \captionof{figure}{Object variations. We obtain variations in the appearance of the objects when generating images with different seed. This can be attributed to the lossy procedure used to obtain appearance vectors.}
    \label{fig:sd_variations}
\end{table*}

\begin{table*}
    \centering
    \def\arraystretch{0.0}
    \footnotesize
    \resizebox{\linewidth}{!}{
    \setlength\tabcolsep{0.2pt}
    \begin{tabular}{ccccccc}
            %\rotatebox[origin=c]{45}{Input Image} & \rotatebox[origin=c]{45}{Reference Image}&  \rotatebox[origin=c]{45}{CP+DDIM+LDM}& E2EVE & SAP \\
            Input & Reference & CP & Inpaint & CP+DDIM& E2EVE %\citet{Brown2022E2EVE}
            &\textbf{Our} \\
            \includegraphics[trim=0 0 0 0,clip,width=0.2\columnwidth]{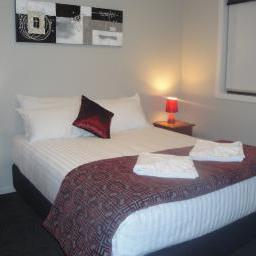} &
            \includegraphics[trim=0 0 0 0,clip,width=0.2\columnwidth]{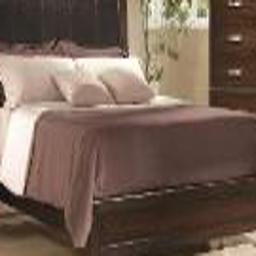} &
            \includegraphics[trim=0 0 0 0,clip,width=0.2\columnwidth]{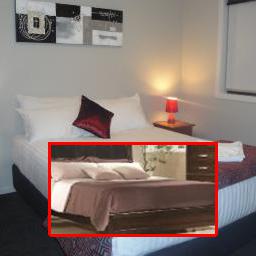} &
            \includegraphics[trim=0 0 0 0,clip,width=0.2\columnwidth]{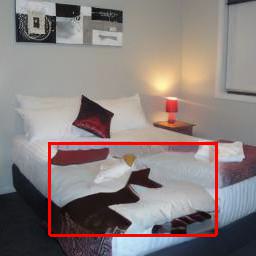} &
            \includegraphics[trim=0 0 0 0,clip,width=0.2\columnwidth]{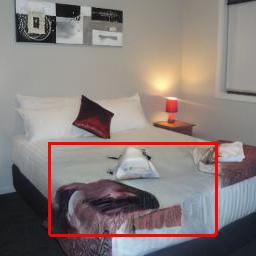} &
            \includegraphics[trim=0 0 0 0,clip,width=0.2\columnwidth]{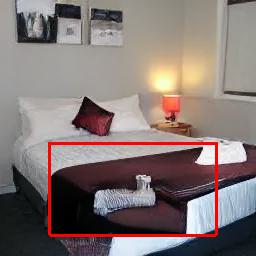} &
            \includegraphics[trim=0 0 0 0,clip,width=0.2\columnwidth]{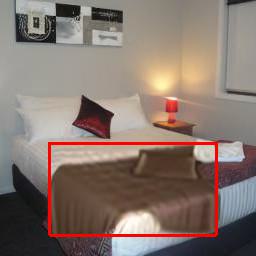} \\

            \includegraphics[trim=0 0 0 0,clip,width=0.2\columnwidth]{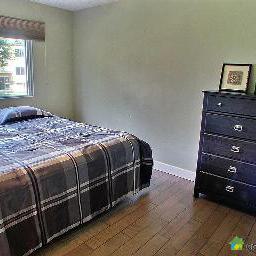} &
            \includegraphics[trim=0 0 0 0,clip,width=0.2\columnwidth]{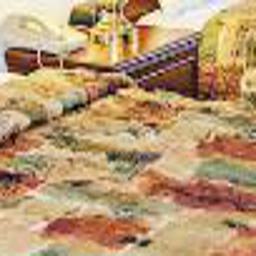} &
            \includegraphics[trim=0 0 0 0,clip,width=0.2\columnwidth]{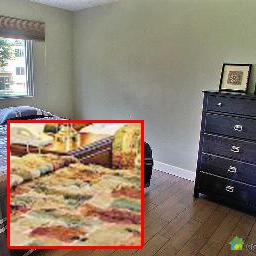} &
            \includegraphics[trim=0 0 0 0,clip,width=0.2\columnwidth]{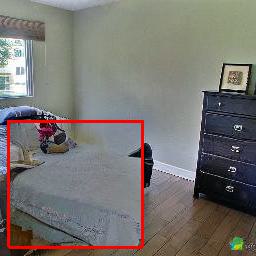} &
            \includegraphics[trim=0 0 0 0,clip,width=0.2\columnwidth]{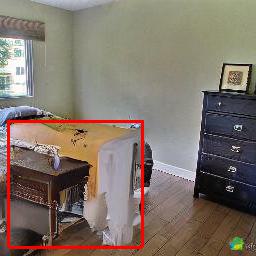} &
            \includegraphics[trim=0 0 0 0,clip,width=0.2\columnwidth]{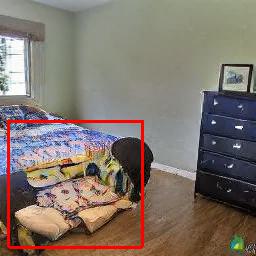} &
            \includegraphics[trim=0 0 0 0,clip,width=0.2\columnwidth]{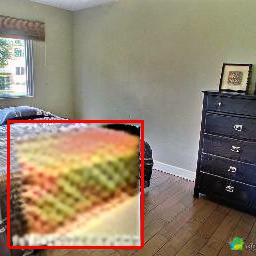} \\

                        \includegraphics[trim=0 0 0 0,clip,width=0.2\columnwidth]{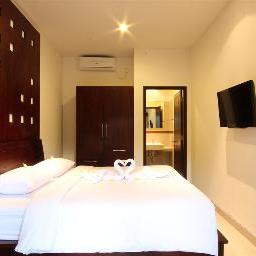} &
            \includegraphics[trim=0 0 0 0,clip,width=0.2\columnwidth]{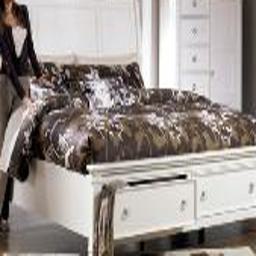} &
            \includegraphics[trim=0 0 0 0,clip,width=0.2\columnwidth]{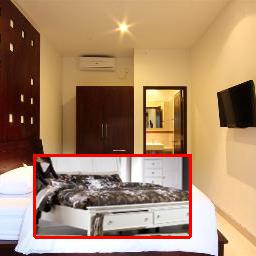} &
            \includegraphics[trim=0 0 0 0,clip,width=0.2\columnwidth]{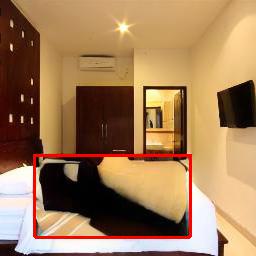} &
            \includegraphics[trim=0 0 0 0,clip,width=0.2\columnwidth]{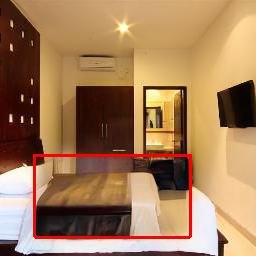} &
            \includegraphics[trim=0 0 0 0,clip,width=0.2\columnwidth]{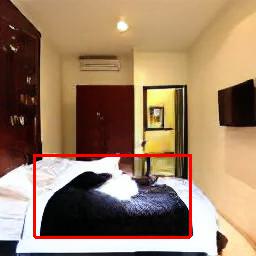} &
            \includegraphics[trim=0 0 0 0,clip,width=0.2\columnwidth]{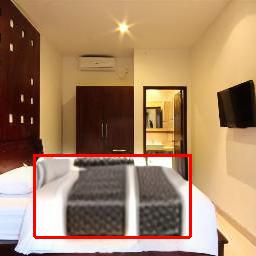} \\

                        \includegraphics[trim=0 0 0 0,clip,width=0.2\columnwidth]{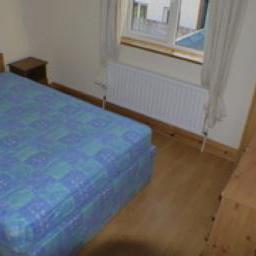} &
            \includegraphics[trim=0 0 0 0,clip,width=0.2\columnwidth]{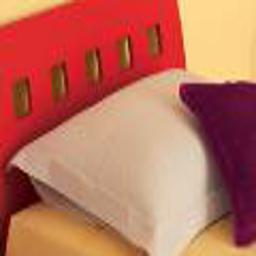} &
            \includegraphics[trim=0 0 0 0,clip,width=0.2\columnwidth]{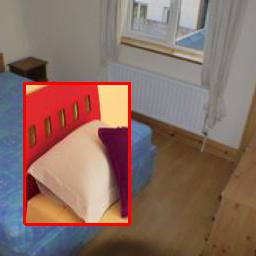} &
            \includegraphics[trim=0 0 0 0,clip,width=0.2\columnwidth]{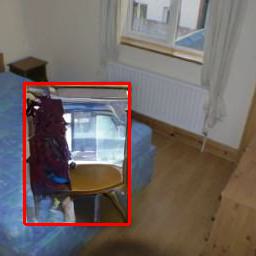} &
            \includegraphics[trim=0 0 0 0,clip,width=0.2\columnwidth]{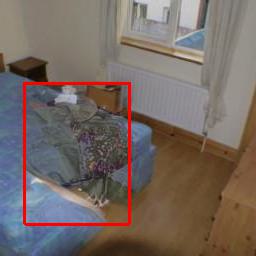} &
            \includegraphics[trim=0 0 0 0,clip,width=0.2\columnwidth]{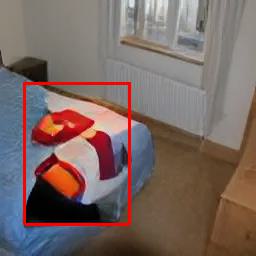} &
            \includegraphics[trim=0 0 0 0,clip,width=0.2\columnwidth]{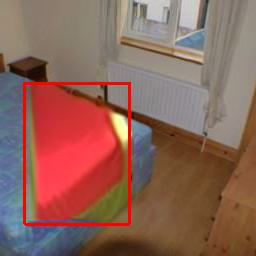} \\

            \includegraphics[trim=0 0 0 0,clip,width=0.2\columnwidth]{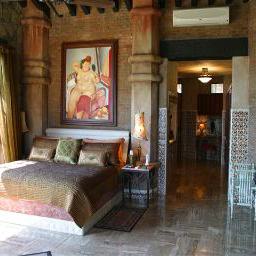} &
            \includegraphics[trim=0 0 0 0,clip,width=0.2\columnwidth]{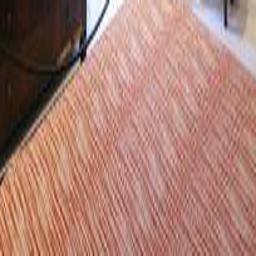} &
            \includegraphics[trim=0 0 0 0,clip,width=0.2\columnwidth]{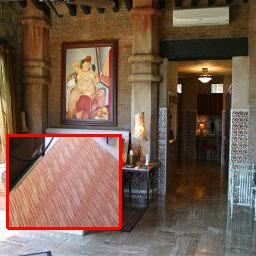} &
            \includegraphics[trim=0 0 0 0,clip,width=0.2\columnwidth]{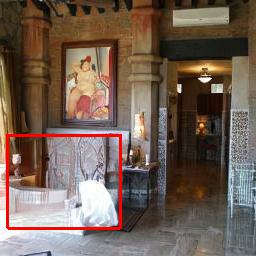} &
            \includegraphics[trim=0 0 0 0,clip,width=0.2\columnwidth]{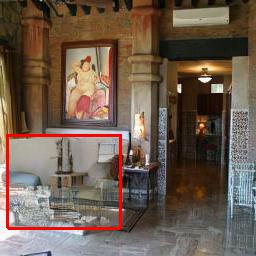} &
            \includegraphics[trim=0 0 0 0,clip,width=0.2\columnwidth]{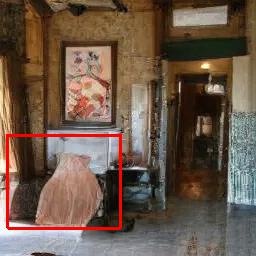} &
            \includegraphics[trim=0 0 0 0,clip,width=0.2\columnwidth]{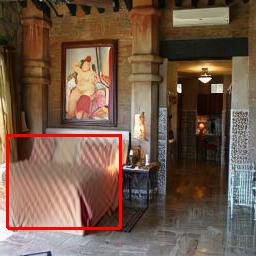} \\

                        \includegraphics[trim=0 0 0 0,clip,width=0.2\columnwidth]{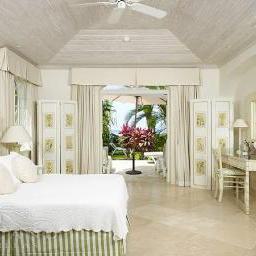} &
            \includegraphics[trim=0 0 0 0,clip,width=0.2\columnwidth]{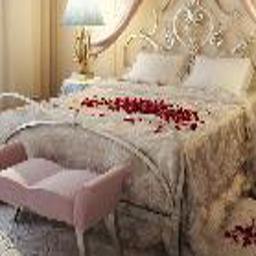} &
            \includegraphics[trim=0 0 0 0,clip,width=0.2\columnwidth]{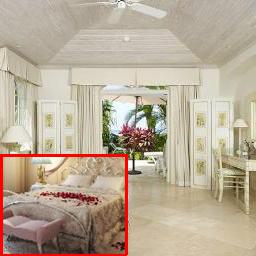} &
            \includegraphics[trim=0 0 0 0,clip,width=0.2\columnwidth]{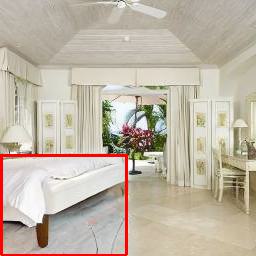} &
            \includegraphics[trim=0 0 0 0,clip,width=0.2\columnwidth]{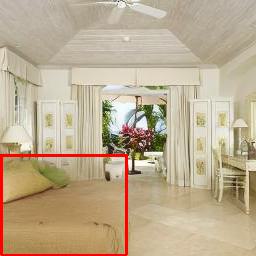} &
            \includegraphics[trim=0 0 0 0,clip,width=0.2\columnwidth]{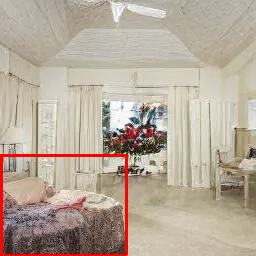} &
            \includegraphics[trim=0 0 0 0,clip,width=0.2\columnwidth]{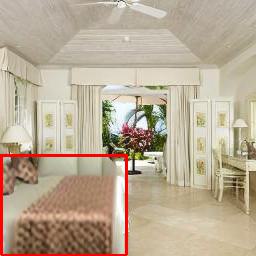} \\

                        \includegraphics[trim=0 0 0 0,clip,width=0.2\columnwidth]{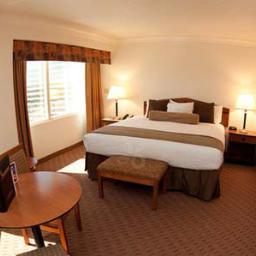} &
            \includegraphics[trim=0 0 0 0,clip,width=0.2\columnwidth]{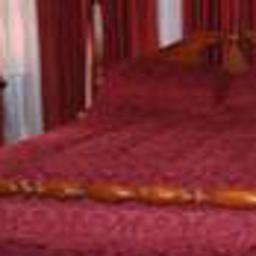} &
            \includegraphics[trim=0 0 0 0,clip,width=0.2\columnwidth]{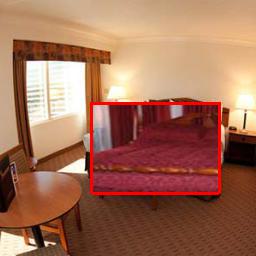} &
            \includegraphics[trim=0 0 0 0,clip,width=0.2\columnwidth]{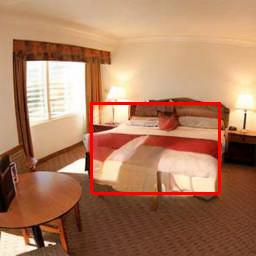} &
            \includegraphics[trim=0 0 0 0,clip,width=0.2\columnwidth]{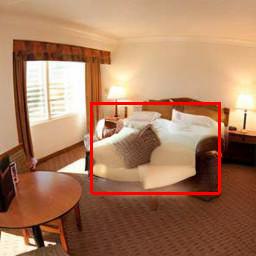} &
            \includegraphics[trim=0 0 0 0,clip,width=0.2\columnwidth]{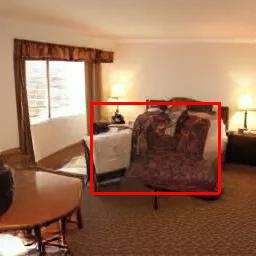} &
            \includegraphics[trim=0 0 0 0,clip,width=0.2\columnwidth]{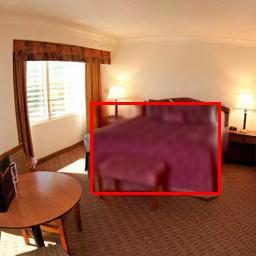} \\

    \end{tabular}
    }
    \captionof{figure}{Visual results for in-domain localized image editing. We edit the input image, using as a driver the reference image, targeting the red-boxed area. With PAIR Diffusion we can perform realistic edits in challenging scenarios. For example, in the first row, we can use the entire bed as a driver and edit only a patch of the input image. On the contrary, in the last row, we use a small patch as a driver and target the whole bed of the input image for the edit. In both cases, our method outputs realistic edited images. Moreover, due to the masked DDIM technique, we introduce almost no distortion in the area outside the red box (\ie the one that should not change). We show results for all the baselines. Note that unlike other results in the paper which uses edit regions specified by the user or generated using a segmentation model, here we define edit regions using predefined boxes.}
    \label{fig:local_edits_baselines_supp}
\end{table*}

% \begin{figure*}[b]
%   \centering
%   \includegraphics[width=\linewidth]{diagrams/Interpolationv2.pdf}
%   \caption{Interpolation results. We can interpolate between two a }
% \label{fig:interpolation_results} 
% \end{figure*}
\begin{table*}
    \centering
    \def\arraystretch{0.0}
    \footnotesize
    \resizebox{\linewidth}{!}{
    \setlength\tabcolsep{0.3pt}
    \begin{tabular}{cccccccc}
            & Input & $\lambda=0.2$ &  $\lambda=0.4$ & $\lambda=0.6$ & $\lambda=0.8$ & $\lambda=1.0$ & Reference \\
            \rotatebox{90}{\hspace{3.75mm}Floor} &
            \includegraphics[trim=0 0 0 0,clip,width=0.2\columnwidth]{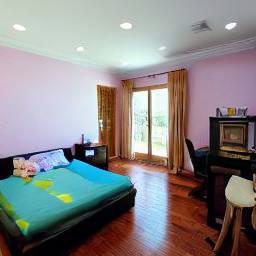} &
            \includegraphics[trim=0 0 0 0,clip,width=0.2\columnwidth]{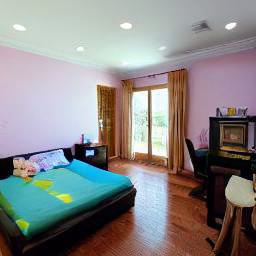} &
            \includegraphics[trim=0 0 0 0,clip,width=0.2\columnwidth]{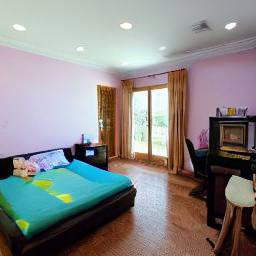} &
            \includegraphics[trim=0 0 0 0,clip,width=0.2\columnwidth]{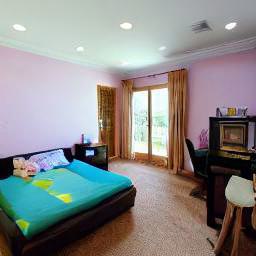} &
            \includegraphics[trim=0 0 0 0,clip,width=0.2\columnwidth]{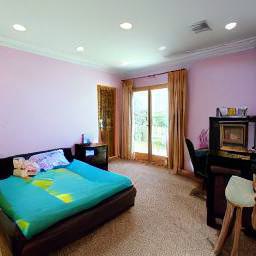} &
            \includegraphics[trim=0 0 0 0,clip,width=0.2\columnwidth]{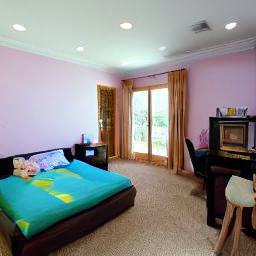} &
            \includegraphics[trim=0 0 0 0,clip,width=0.2\columnwidth]{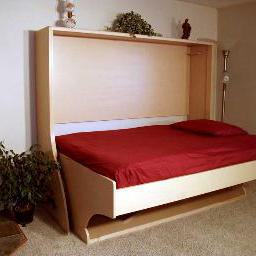} \\

            \rotatebox{90}{\hspace{3.75mm}Bed} &
            \includegraphics[trim=0 0 0 0,clip,width=0.2\columnwidth]{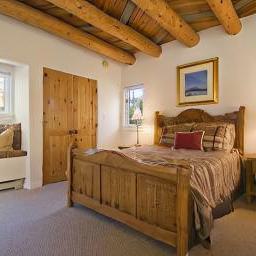} &
            \includegraphics[trim=0 0 0 0,clip,width=0.2\columnwidth]{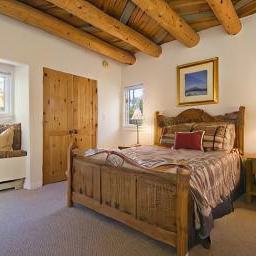} &
            \includegraphics[trim=0 0 0 0,clip,width=0.2\columnwidth]{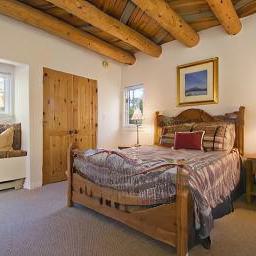} &
            \includegraphics[trim=0 0 0 0,clip,width=0.2\columnwidth]{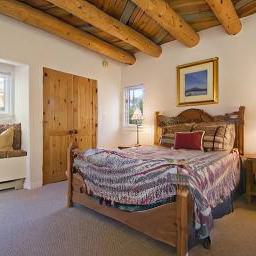} &
            \includegraphics[trim=0 0 0 0,clip,width=0.2\columnwidth]{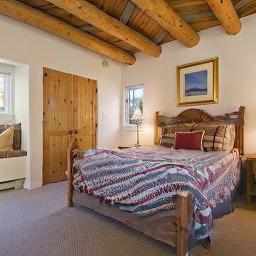} &
            \includegraphics[trim=0 0 0 0,clip,width=0.2\columnwidth]{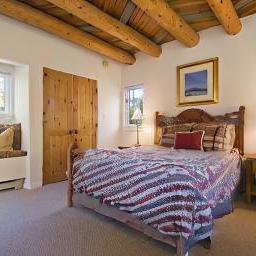} &
            \includegraphics[trim=0 0 0 0,clip,width=0.2\columnwidth]{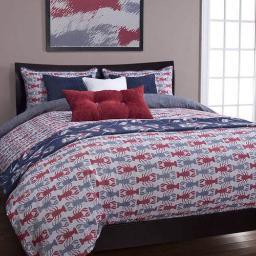} \\

            \rotatebox{90}{\hspace{3.75mm}Tree} &
            \includegraphics[trim=0 0 0 0,clip,width=0.2\columnwidth]{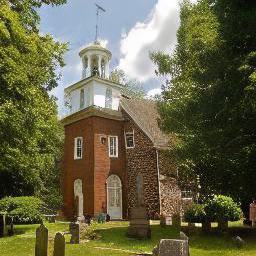} &
            \includegraphics[trim=0 0 0 0,clip,width=0.2\columnwidth]{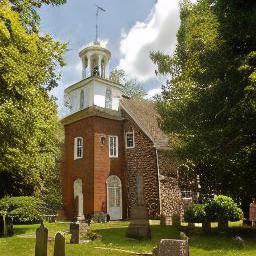} &
            \includegraphics[trim=0 0 0 0,clip,width=0.2\columnwidth]{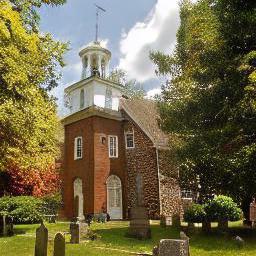} &
            \includegraphics[trim=0 0 0 0,clip,width=0.2\columnwidth]{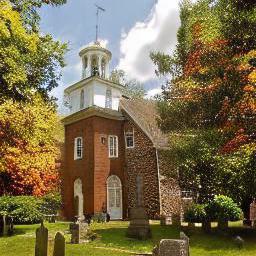} &
            \includegraphics[trim=0 0 0 0,clip,width=0.2\columnwidth]{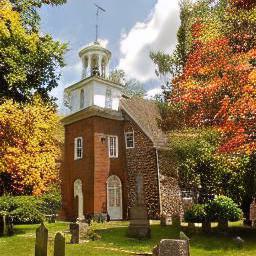} &
            \includegraphics[trim=0 0 0 0,clip,width=0.2\columnwidth]{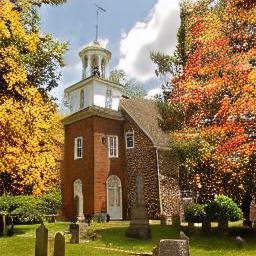} &
            \includegraphics[trim=0 0 0 0,clip,width=0.2\columnwidth]{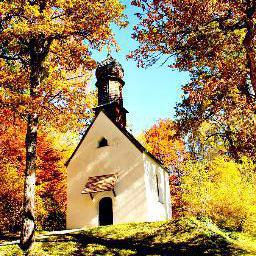} \\

            \rotatebox{90}{\hspace{3.75mm}Sky} &
            \includegraphics[trim=0 0 0 0,clip,width=0.2\columnwidth]{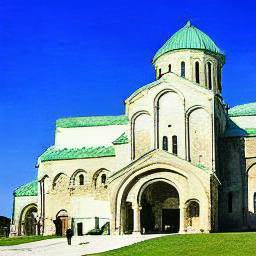} &
            \includegraphics[trim=0 0 0 0,clip,width=0.2\columnwidth]{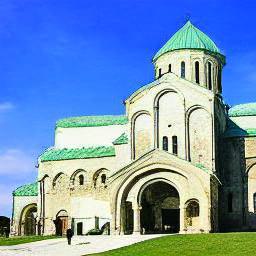} &
            \includegraphics[trim=0 0 0 0,clip,width=0.2\columnwidth]{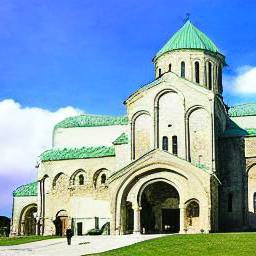} &
            \includegraphics[trim=0 0 0 0,clip,width=0.2\columnwidth]{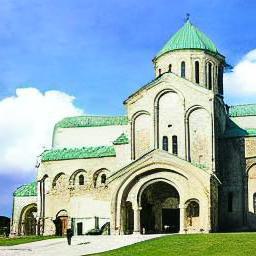} &
            \includegraphics[trim=0 0 0 0,clip,width=0.2\columnwidth]{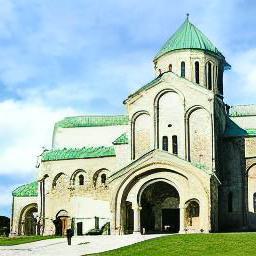} &
            \includegraphics[trim=0 0 0 0,clip,width=0.2\columnwidth]{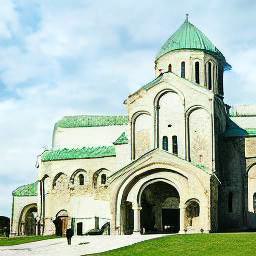} &
            \includegraphics[trim=0 0 0 0,clip,width=0.2\columnwidth]{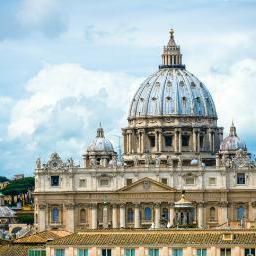} \\

                        \rotatebox{90}{\hspace{3.75mm}Skin} &
            \includegraphics[trim=0 0 0 0,clip,width=0.2\columnwidth]{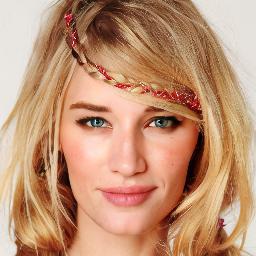} &
            \includegraphics[trim=0 0 0 0,clip,width=0.2\columnwidth]{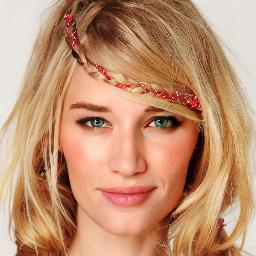} &
            \includegraphics[trim=0 0 0 0,clip,width=0.2\columnwidth]{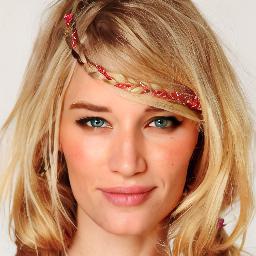} &
            \includegraphics[trim=0 0 0 0,clip,width=0.2\columnwidth]{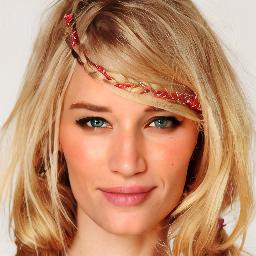} &
            \includegraphics[trim=0 0 0 0,clip,width=0.2\columnwidth]{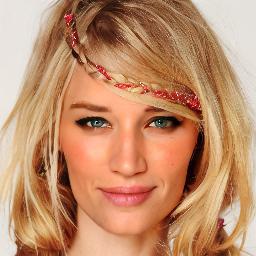} &
            \includegraphics[trim=0 0 0 0,clip,width=0.2\columnwidth]{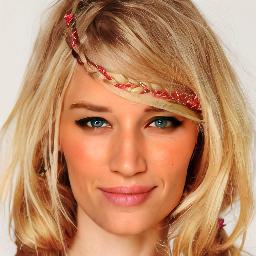} &
            \includegraphics[trim=0 0 0 0,clip,width=0.2\columnwidth]{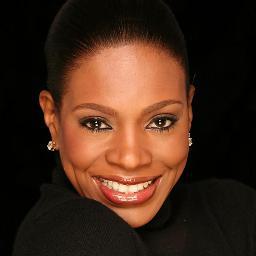} \\

                        \rotatebox{90}{\hspace{3.75mm}Hair} &
            \includegraphics[trim=0 0 0 0,clip,width=0.2\columnwidth]{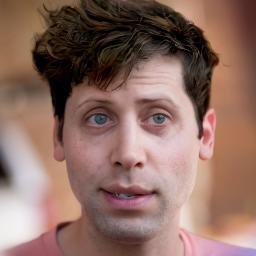} &
            \includegraphics[trim=0 0 0 0,clip,width=0.2\columnwidth]{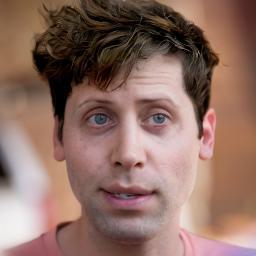} &
            \includegraphics[trim=0 0 0 0,clip,width=0.2\columnwidth]{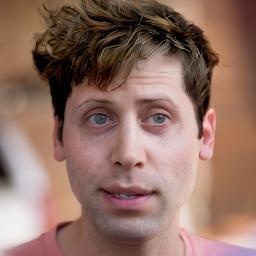} &
            \includegraphics[trim=0 0 0 0,clip,width=0.2\columnwidth]{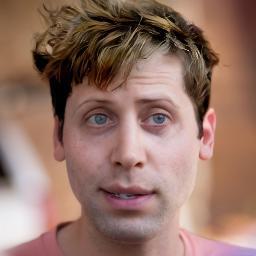} &
            \includegraphics[trim=0 0 0 0,clip,width=0.2\columnwidth]{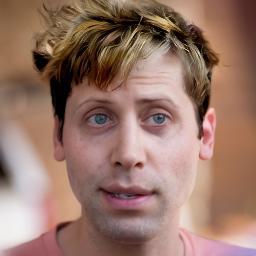} &
            \includegraphics[trim=0 0 0 0,clip,width=0.2\columnwidth]{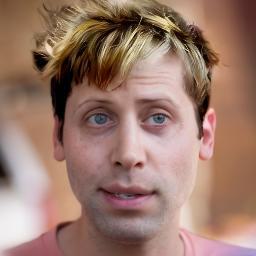} &
            \includegraphics[trim=0 0 0 0,clip,width=0.2\columnwidth]{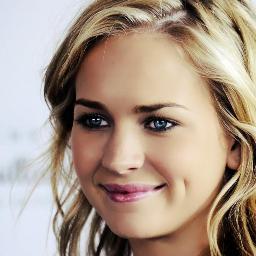} \\
    \end{tabular}
    }

    \captionof{figure}{We can control the strength of appearance and interpolate between the reference and input appearances. We set appearance as $f' = (1 - \lambda)f_i + \lambda f^{R}_j$ where $f_i$ is the input appearance and $f^{R}_j$ is the reference appearance and vary $\lambda$ from 0 to 1.}
    \label{fig:inter_supp}
\end{table*}
\begin{table*}
    \centering
    \def\arraystretch{0.0}
    \footnotesize
    \resizebox{0.75\linewidth}{!}{
    \setlength\tabcolsep{0.18pt}
    \begin{tabular}{ccccc}
            Input & \multicolumn{2}{c}{Wall} & \multicolumn{2}{c}{Bed} \\
            \includegraphics[trim=0 0 0 0,clip,width=0.18\columnwidth]{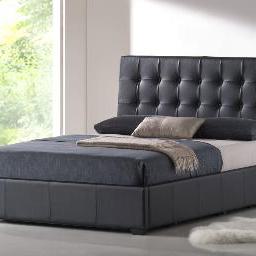} &
            \includegraphics[trim=0 0 0 0,clip,width=0.18\columnwidth]{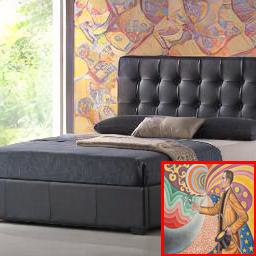} &
            \includegraphics[trim=0 0 0 0,clip,width=0.18\columnwidth]{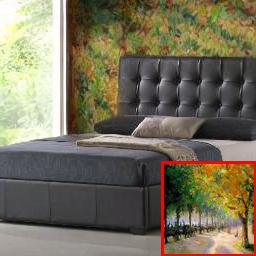} &
            \includegraphics[trim=0 0 0 0,clip,width=0.18\columnwidth]{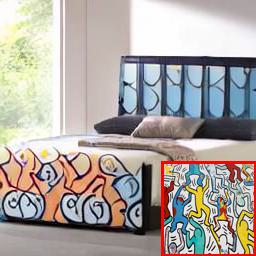} &
            \includegraphics[trim=0 0 0 0,clip,width=0.18\columnwidth]{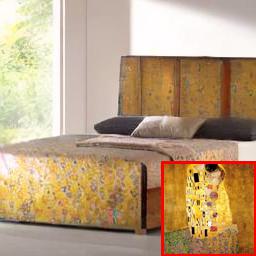} \\
            
            \includegraphics[trim=0 0 0 0,clip,width=0.18\columnwidth]{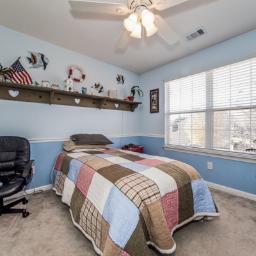} &
            \includegraphics[trim=0 0 0 0,clip,width=0.18\columnwidth]{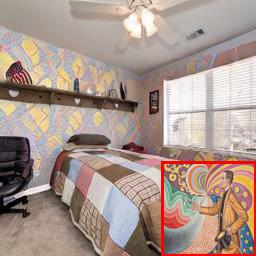} &
            \includegraphics[trim=0 0 0 0,clip,width=0.18\columnwidth]{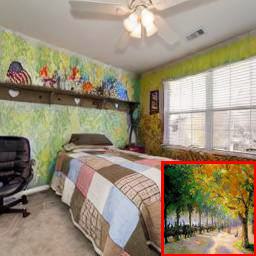} &
            \includegraphics[trim=0 0 0 0,clip,width=0.18\columnwidth]{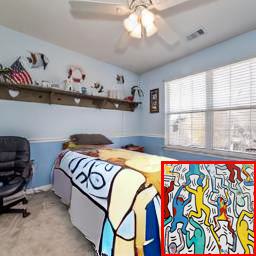} &
            \includegraphics[trim=0 0 0 0,clip,width=0.18\columnwidth]{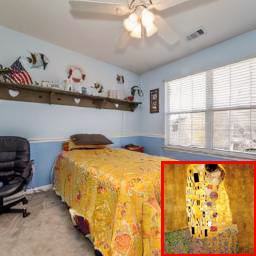} \\

            Input & \multicolumn{2}{c}{Church} & \multicolumn{2}{c}{Sky} \\
            \includegraphics[trim=0 0 0 0,clip,width=0.18\columnwidth]{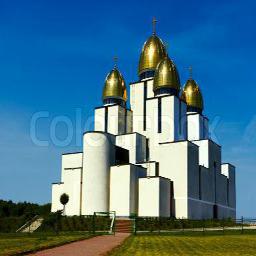} &
            \includegraphics[trim=0 0 0 0,clip,width=0.18\columnwidth]{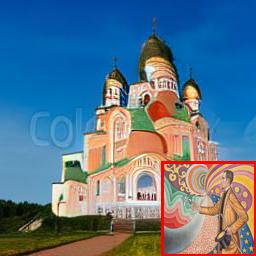} &
            \includegraphics[trim=0 0 0 0,clip,width=0.18\columnwidth]{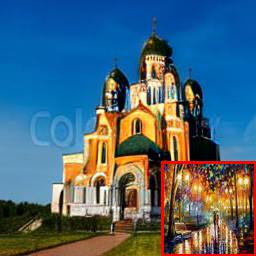} &            \includegraphics[trim=0 0 0 0,clip,width=0.18\columnwidth]{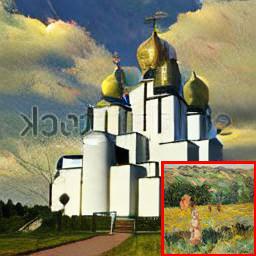} &            \includegraphics[trim=0 0 0 0,clip,width=0.18\columnwidth]{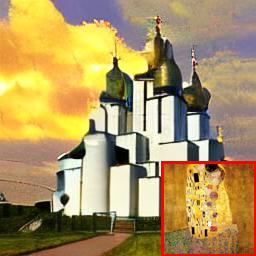} \\

            \includegraphics[trim=0 0 0 0,clip,width=0.18\columnwidth]{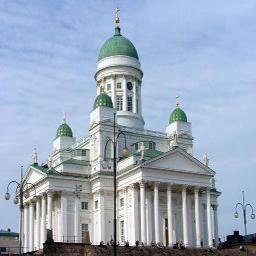} &
            \includegraphics[trim=0 0 0 0,clip,width=0.18\columnwidth]{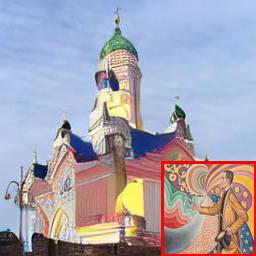} &
            \includegraphics[trim=0 0 0 0,clip,width=0.18\columnwidth]{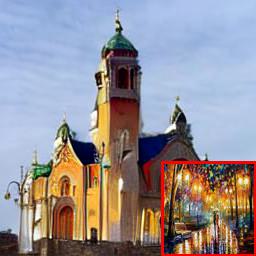} &            
            \includegraphics[trim=0 0 0 0,clip,width=0.18\columnwidth]{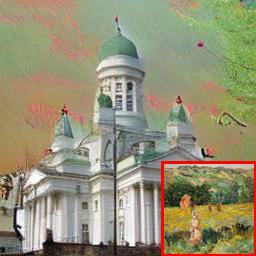} &            
            \includegraphics[trim=0 0 0 0,clip,width=0.18\columnwidth]{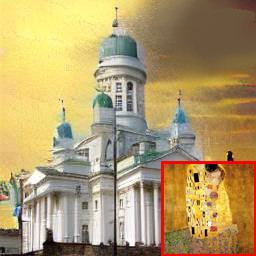} \\
            \vspace{3pt} \\
            \multicolumn{5}{c}{(a)} \\
            \vspace{3pt} \\

            % new line
            
            \includegraphics[trim=0 0 0 0,clip,width=0.18\columnwidth]{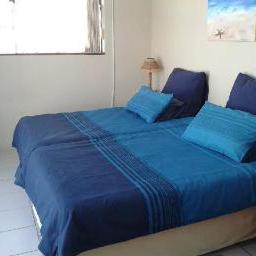} &
            \includegraphics[trim=0 0 0 0,clip,width=0.18\columnwidth]{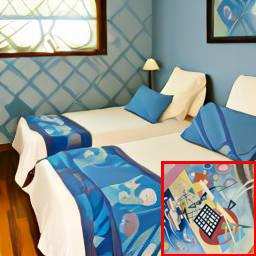} &
            \includegraphics[trim=0 0 0 0,clip,width=0.18\columnwidth]{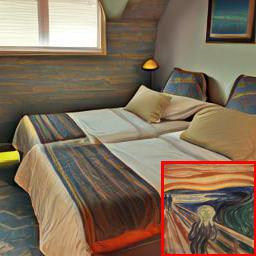} &
            \includegraphics[trim=0 0 0 0,clip,width=0.18\columnwidth]{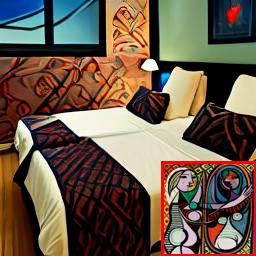} &
            \includegraphics[trim=0 0 0 0,clip,width=0.18\columnwidth]{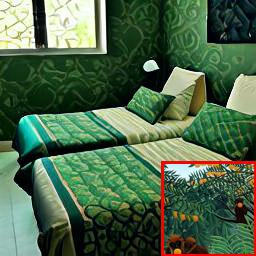} \\

            \includegraphics[trim=0 0 0 0,clip,width=0.18\columnwidth]{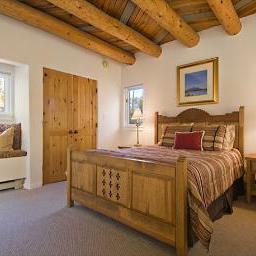} &
            \includegraphics[trim=0 0 0 0,clip,width=0.18\columnwidth]{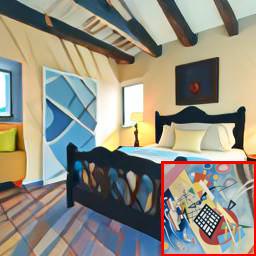} &
            \includegraphics[trim=0 0 0 0,clip,width=0.18\columnwidth]{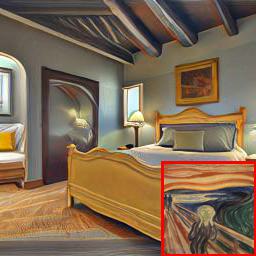} &
            \includegraphics[trim=0 0 0 0,clip,width=0.18\columnwidth]{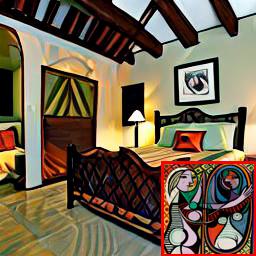} &
            \includegraphics[trim=0 0 0 0,clip,width=0.18\columnwidth]{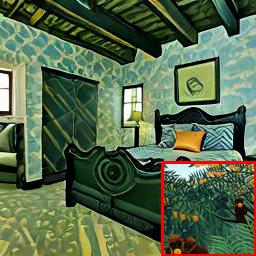}  \\

             \includegraphics[trim=0 0 0 0,clip,width=0.18\columnwidth]{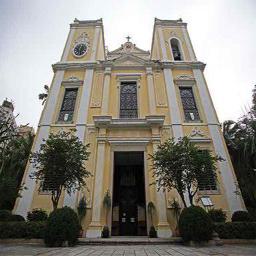} &
            \includegraphics[trim=0 0 0 0,clip,width=0.18\columnwidth]{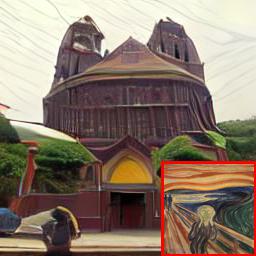} &
            \includegraphics[trim=0 0 0 0,clip,width=0.18\columnwidth]{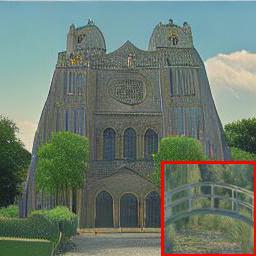} &
            \includegraphics[trim=0 0 0 0,clip,width=0.18\columnwidth]{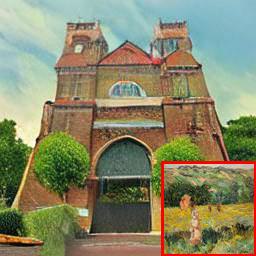} &
            \includegraphics[trim=0 0 0 0,clip,width=0.18\columnwidth]{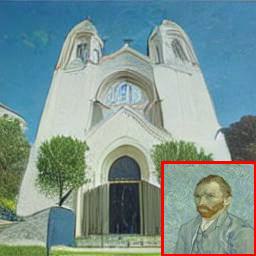} \\

            \includegraphics[trim=0 0 0 0,clip,width=0.18\columnwidth]{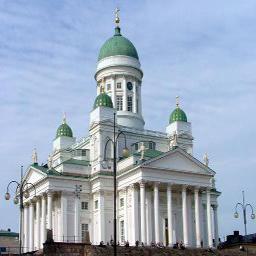} &
            \includegraphics[trim=0 0 0 0,clip,width=0.18\columnwidth]{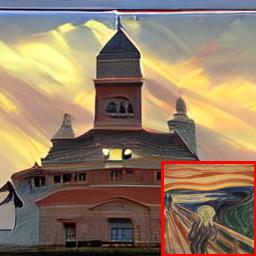} &
            \includegraphics[trim=0 0 0 0,clip,width=0.18\columnwidth]{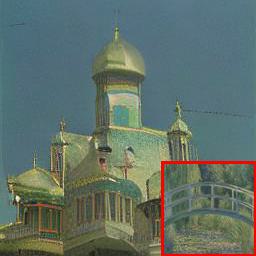} &
            \includegraphics[trim=0 0 0 0,clip,width=0.18\columnwidth]{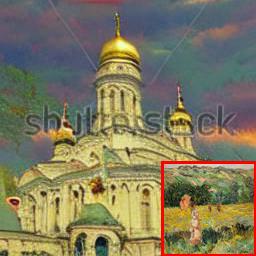} &
            \includegraphics[trim=0 0 0 0,clip,width=0.18\columnwidth]{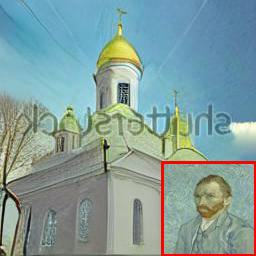} \\

            \vspace{3pt} \\
            \multicolumn{5}{c}{(b)}
            
    \end{tabular}
    }
    \captionof{figure}{Visual results for (a) local image editing, (b) global image appearance manipulation.}
    \label{fig:local_od_apperance_supp}
\end{table*}

\end{document}